\documentclass{article} 
\usepackage{iclr2027_conference,times}

\usepackage{amsmath,amsfonts,bm}

\def\eqref#1{equation~\ref{#1}}

\def\1{\bm{1}}

\DeclareMathAlphabet{\mathsfit}{\encodingdefault}{\sfdefault}{m}{sl}
\SetMathAlphabet{\mathsfit}{bold}{\encodingdefault}{\sfdefault}{bx}{n}

\usepackage{hyperref}
\usepackage{url}
\usepackage{comment}
\usepackage{booktabs}
\usepackage[table]{xcolor}
\definecolor{lightblue}{RGB}{220,235,247}
\usepackage{lipsum}
\usepackage{placeins}
\usepackage{graphicx}
\usepackage{amsmath}
\usepackage{amsthm}
\usepackage{amsfonts}
\usepackage[OT1]{fontenc}
\usepackage{bm}
\usepackage{multirow}
\usepackage{subcaption}
\usepackage{enumitem}

\usepackage{algorithm}
\usepackage{algpseudocode}

\theoremstyle{plain}

\theoremstyle{definition}

\theoremstyle{remark}

\usepackage{algorithm}
\usepackage{algpseudocode}

\definecolor{lightgreen}{HTML}{b5d9b5}
\definecolor{lightorange}{HTML}{eec8ad}

\newcommand{\methodname}{GRACE}
\newcommand{\simos}[1]{\textcolor{black}{#1}}

\title{Localisation-Aware Uncertainty for Pretrained Object Detection}

\author{
Charmaine Barker$^{1}$\thanks{Equal contribution},
Daniel Bethell$^{1}$\footnotemark[1],
Simos Gerasimou$^{1,2}$ \\
\\
$^{1}$Department of Computer Science, University of York, UK \\
$^{2}$Department of Elect. Eng., and Computer Science and Eng., Cyprus University of Technology, Cyprus \\
\texttt{\{charmaine.barker, daniel.bethell, simos.gerasimou\}@york.ac.uk}
}

\iclrfinalcopy 
\begin{document}

\maketitle

\begin{abstract}
Reliable uncertainty estimation is essential for deploying object detectors when distribution/covariate shift and adversarial attacks may occur. Existing approaches often require detector retraining, architectural modification, or repeated inference, which may be infeasible or incur significant overheads. We introduce a lightweight post-hoc evidential meta-model that learns when object localisations should be considered uncertain while keeping the base detector frozen. 
Our approach automatically identifies localisation-relevant features and uses saliency-guided modification to construct an increasingly challenging curriculum. Detection-level targets combine localisation error, modification level, and prediction instability to guide an evidential meta-model to estimate uncertainty for each predicted bounding box. Our approach requires no changes to the detector and preserves its original localisation outputs. Across adversarial attacks and evaluated strengths, GRACE improves TP-FP AUROC by 22\% relative to the strongest comparator in some cases while maintaining in-distribution detection performance.
\end{abstract}

\section{Introduction}
\label{sec:Introduction}

Object detection is a core component of modern computer vision systems, enabling models to localise objects in images~\citep{liu2020deep}. In safety-critical applications such as autonomous driving and robotics, reliable deployment requires more than strong precision on in-distribution data~\citep{hall2020probabilistic}. A detector must also indicate when its predictions are unreliable, particularly under distribution shift, image corruption, or adversarial perturbation~\citep{ovadia2019can}. This is more challenging than standard classification uncertainty because detection reliability is structured: a model may assign a high class confidence to a poorly localised box, produce spurious detections on unfamiliar inputs, or generate predictions whose locations drift substantially under small perturbations. Therefore, uncertainty in object detection must be detection-level and localisation-aware, rather than only image-level or class-confidence-based.

Existing uncertainty quantification (UQ) approaches only partially address this problem~\citep{abdar2021review}. Bayesian and sampling-based methods, such as Monte Carlo Dropout~\citep{gal2016dropout}, can provide uncertainty estimates, but often require stochastic inference, architectural assumptions, or repeated forward passes that increase deployment cost. Detection-specific approaches estimate uncertainty from detector outputs or learned auxiliary models~\citep{schubert2021metadetect}, but they generally rely on the detector's existing confidence structure, increase deployment cost, and do not explicitly \simos{guide} the uncertainty model when localisation should become unreliable. More broadly, many uncertainty methods require retraining, modify the detector, or \simos{mandate} access to training pipelines that may be unavailable for large, proprietary, or already deployed detectors~\citep{franchi2024make}. This limits their suitability for post-hoc deployment, where the base detector should remain frozen, and uncertainty should be added without changing its predictions.

We introduce \methodname\ (\textbf{G}uided evidential \textbf{R}egression for \textbf{A}dversarial and \textbf{C}ovariate uncertainty \textbf{E}stimation), a localisation-aware post-hoc evidential meta-model for object detection that provides uncertainty estimates for each predicted object. \methodname\ performs saliency calibration to identify localisation-relevant detector features and input regions. It then uses these regions to construct a monotonic noise-driven curriculum, in which progressively corrupted views expose when a retained detection becomes poorly localised, heavily affected by corruption, or unstable under perturbation. This curriculum provides detection-specific uncertainty targets that guide a lightweight meta-model to estimate uncertainty. Importantly, \methodname\ leaves the base detector unchanged and preserves its original box predictions, learning only an uncertainty signal that indicates when each localisation output should be trusted. 

\section{Related Work}
\label{sec:Related Work}

\textbf{Uncertainty Quantification.} Uncertainty in deep learning generally reflects either noise inherent in the observations or uncertainty in the model itself, both of which are important in safety-critical applications~\citep{he2023survey}. Aleatoric uncertainty captures irreducible, input-dependent variability and is commonly modelled using predictive distributions~\citep{kendall2017uncertainties}, prediction intervals~\citep{tagasovska2019single}, or generative approaches~\citep{kingma2013auto}. Epistemic uncertainty instead reflects limited model knowledge and can be estimated using Bayesian inference~\citep{goan2020bayesian, blei2017variational}, Laplace approximations~\citep{fortuin2022priors}, or scalable approximations such as Monte Carlo Dropout~\citep{gal2016dropout}, deep ensembles~\citep{lakshminarayanan2017simple}, and distance-aware methods~\citep{liu2020simple}. Model-agnostic techniques, such as test-time augmentation, estimate uncertainty from prediction variability but do not explicitly distinguish between these sources~\citep{ayhan2018test}. Since both forms may occur simultaneously, methods such as ensemble-based prediction intervals~\citep{pearce2018high} and conformal approaches~\citep{angelopoulos2023conformal,bethell2024robust} aim to capture uncertainty more broadly, although often with additional computational or calibration overhead.

\textbf{Post-hoc Uncertainty.} 
Post-hoc UQ is particularly attractive for object detection, where retraining or modifying large pretrained detectors can be computationally expensive or impractical in deployed systems. Rather than altering the detector, recent approaches estimate uncertainty from its existing representations or predictions, with recent post-hoc object detection methods deriving uncertainty from frozen internal representations~\citep{shen2023post}, detector outputs, or bounding-box characteristics~\citep{schubert2021metadetect} using lightweight auxiliary models~\citep{lee2022lightweight}. Related post-hoc approaches have also learned uncertainty from saliency-guided frozen representations, but remain classification-oriented and do not directly address detection-level localisation uncertainty~\citep{barker2026gradual}. Although these approaches avoid retraining the underlying detector, they can introduce additional inference or uncertainty-generation costs, while their robustness to covariate shift and adversarial perturbations remains limited or underexplored. This motivates the need for a lightweight post-hoc approach that preserves the frozen detector while learning localisation-aware uncertainty that remains informative as inputs become unreliable.

\section{Preliminaries}
\label{sec:Preliminaries}

We study the standard supervised object detection setting, where the objective is to train a predictive model over a finite labelled set. The input domain is $X \subseteq \mathbb{R}^{H \times W \times C}$ and the output space is a structured set of labelled bounding boxes. For an input image $x \in X$, the corresponding annotation is given by $y = \{(b_j, c_j)\}_{j=1}^{J}$, where $J$ denotes the number of objects in the image, $b_j \in B \subseteq \mathbb{R}^4$ is the bounding box associated with object $j \in \{1, \ldots J\}$, and $c_j \in Y = \{1, \ldots, K\}$ is its class label among $K$ discrete categories. The training (ID) data $\mathcal{D} = \{(x_i, y_i)\}_{i=1}^N$ comprises samples $(x_i, y_i) \in X \times \mathcal{Y}$ drawn independently and identically from the joint distribution $p(x, y) = p(x)\, p(y \mid x)$, where $\mathcal{Y}$ denotes the space of variable-sized detection annotations. The task is to estimate the conditional distribution $p(y \mid x)$ over object locations and class assignments.

Beyond detection accuracy, we also focus on assessing predictive uncertainty. A pretrained deterministic detector $f_\theta$ trained on ID data, though often accurate, is typically overconfident, especially on OOD and adversarial inputs. This occurs since its bounding-box predictions are optimised for accurate detection on the training distribution, but do not explicitly quantify epistemic uncertainty, often yielding spuriously confident detections on unfamiliar or corrupted inputs. \methodname\ reduces such overconfidence in a fully post-hoc manner, using only the base detector’s outputs and features without modifying the structure of the model $f$ or its weights $\theta$, as intrusive changes may be inaccessible (for closed-weight models), infeasible (for large models), or harmful to performance.

\begin{figure*}[tb]
    \centering
    \includegraphics[width=\linewidth]{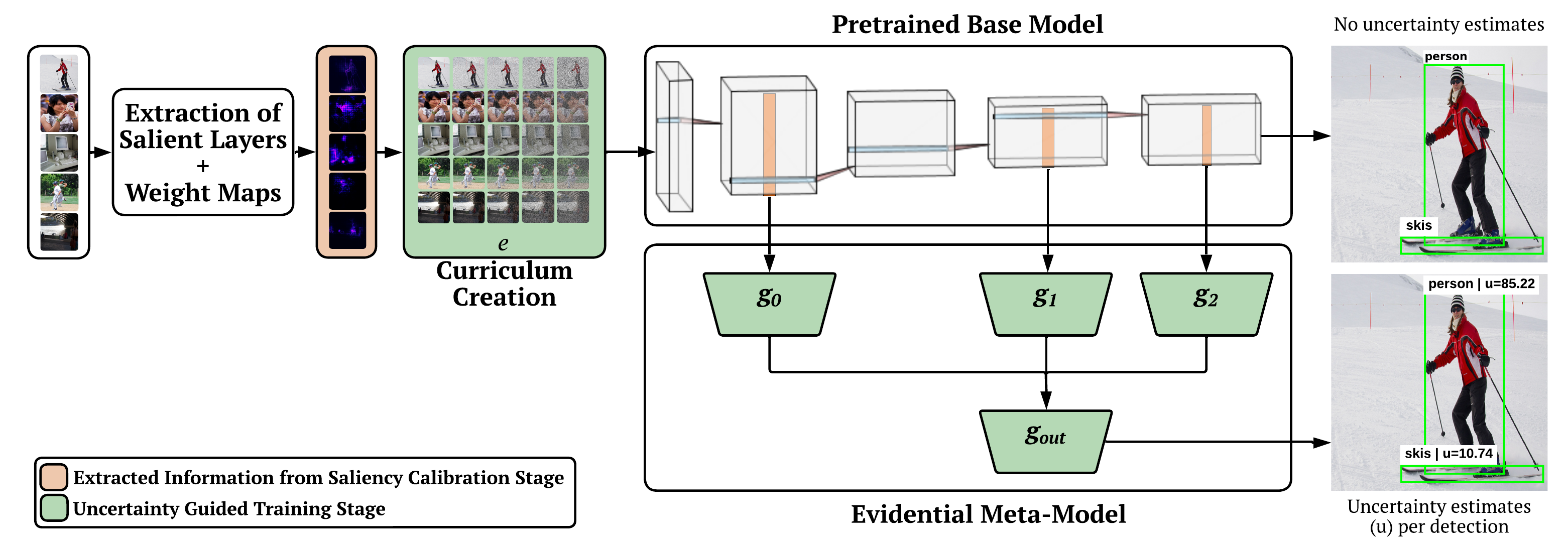}
    \caption{
    Overview of \methodname\ meta-model approach, showing the \colorbox{lightorange}{Saliency Calibration} and \colorbox{lightgreen}{Uncertainty Guided Training} stages. \methodname\ extracts salient features and weight maps from a pretrained model in a fully post-hoc manner, then generates a noise-driven curriculum to teach the meta-model when to be uncertain. In the figure, $g_0, g_1, g_2$ are evidential projection branches from salient layers, and $u$ is the predicted uncertainty from the Dirichlet evidence.}
    \label{fig:architecture}
\end{figure*}

\section{\methodname}
\label{sec:GRACE}

Our \textbf{G}uided Evidential \textbf{R}egression for \textbf{A}dversarial and \textbf{C}ovariate Uncertainty \textbf{E}stimation (\methodname) meta-model approach (Figure~\ref{fig:architecture}) enhances uncertainty calibration and robustness to OOD, covariate, and adversarial inputs. \methodname\ adaptively identifies both salient intermediate layers of the pretrained model to connect them to evidential linear layers and salient weight maps. These weight maps are then exploited to construct a monotonic noise-driven curriculum with progressively perturbed inputs. This curriculum enables \methodname\ to learn \textit{when to be uncertain} via monotonic soft-target supervision, which mimics increasing OOD shift and trains the Normal-Inverse-Gamma (NIG) to become less confident as perturbations intensify. This calibration step mitigates the propagation of overconfidence, typically inherited from the pretrained model, and improves robustness under distributional shift. \methodname\ combines relevance-based saliency, evidential meta-modelling, and curriculum noise, but does so through a novel soft-target mechanism that directly guides the NIG to become less confident as reliability decreases.

\subsection{Saliency Calibration}
\label{sec:Saliency Calibration}

The saliency calibration stage of \methodname\ identifies salient components at both the layer and instance levels of the pretrained detector. In the object detection setting, the relevance initialisation is adapted from a class-wise logit target to a localisation-wise detection target. For an input image $x$, let $\mathcal{I}(x) = \{i_{\hat{j}}\}^{\hat{J}}_{\hat{j}=1}$ denote the set of raw detector indices matched to the retained predicted detections (after removal of GT matched detections) and let $r_{i_{\hat{j}}}(x) \in \mathbb{R}^4$ denote the corresponding raw localisation output for detection $\hat{j}$, i.e., the four box-coordinate channels. We then initialise relevance at the output with respect to the scalar localisation objective:
\begin{equation}
\mathcal{R}_L(x) = \sum^{\hat{J}}_{\hat{j}=1} \sum^4_{q=1} r_{i_{\hat{j}},q}(x)
\label{eq:loc-objective}
\end{equation}
Employing the LRP-$\epsilon$ rule~\citep{letzgus2022toward}, we propagate relevance to each hidden layer of the pretrained model for $\ell = L, L-1, \ldots, 1$:
\begin{equation}
\mathcal{R}_{\ell - 1}(x) = \big( \frac{a_{\ell-1}W^\top_\ell}{\langle a_{\ell-1}, W^\top_\ell \rangle + \epsilon } \big) \; \odot \; \mathcal{R}_\ell(x)
\label{eq:lrp}
\end{equation}
where $a_{\ell-1}$ are the activations, $W_\ell$ are the weights, and $\epsilon > 0$ is a small stabilisation term that partially absorbs relevance when there is a contradiction between consecutive layers. 

This formulation yields relevance maps $\{\mathcal{R}_\ell(x)\}_{l=1}^L$ describing how each layer contributes to the final prediction.  Each $R_\ell(x)$ is a vector of size dim$(\ell)$ with each element signifying the relevance attributed to each neuron of the $\ell$-th layer. This propagation rule preserves coordinate-relevant evidence and applied to standard deep learning components (e.g., convolutional, dense, pooling). We then quantify each hidden layer’s global importance by computing the average relevance magnitude over the $N$ input samples:
\begin{equation}
M_\ell = \frac{1}{N}\sum^N_{i=1}\frac{||\mathcal{R}_\ell(x_i)||_1}{|\mathcal{R}_\ell(x_i)|}
\label{eq:layer_rel_mass}
\end{equation}
which enables selecting hidden layers that contain sufficient information and can be used to train the evidential meta-model.  Specifically, sorting layers by $M_\ell$, we select the smallest subset $L_{\text{sal}}$ that covers at least a fraction $\eta$ of the total relevance mass:
\begin{equation}
\sum_{\ell \in L_{\text{sal}}} M_\ell \geq \eta \sum_\ell M_\ell 
\label{eq:coverage-constraint}
\end{equation}
where $\eta \in (0, 1]$ is the cumulative relevance coverage threshold. For instance, $\eta=0.9$ retains the smallest subset of layers that captures 90\% of the total relevance, allowing the meta-model to use the detector representations most associated with localisation.

To construct a noise-driven curriculum aligned with salient input features, we define per-input weight maps from localisation-based attribution maps. For each retained detection $\hat{j}$ and each localisation target $q \in \{1,2,3,4\}$, we compute an input-level attribution map $\mathcal{R}^{(q)}_{0,\hat{j}}(x) \in \mathbb{R}^{H \times W \times C}$ and aggregate relevance across channels to obtain:
\begin{equation}
\mathcal{H}_{\hat{j},q}(x)[h,w] =
\frac{
\max\!\left(0,\sum_{c=1}^{C}\mathcal{R}^{(q)}_{0,\hat{j}}(x)[h,w,c]\right)
}{
\max_{h',w'}
\max\!\left(0,\sum_{c=1}^{C}\mathcal{R}^{(q)}_{0,\hat{j}}(x)[h',w',c]\right)
+\varepsilon
}
\label{eq:target_heatmap}
\end{equation}
where $(h,w)$ denotes a spatial index and $\varepsilon>0$ is a small numerical stabilisation constant. Thus, each $\mathcal{H}_{\hat{j},q}(x)$ is a normalised positive attribution map for localisation target $q$ of detection $\hat{j}$.

These target-specific maps are then reduced across localisation targets. The per-detection heatmap is defined as:
\begin{equation}
\mathcal{H}_{\hat{j}}(x)[h,w] = \frac{1}{4}\sum_{q=1}^{4}\mathcal{H}_{\hat{j},q}(x)[h,w]
\label{eq:detection_heatmap}
\end{equation}
Finally, the per-input weight map $\mathcal{W}(x)$ is obtained by merging the per-detection heatmaps:
\begin{equation}
\tilde{\mathcal{W}}(x)[h,w] = \sum_{\hat{j}=1}^{\hat{J}} \mathcal{H}_{\hat{j}}(x)[h,w].
\label{eq:weight_map_sum_raw}
\end{equation}
The final weight map is then obtained by applying non-negativity and spatial normalisation:
\begin{equation}
\mathcal{W}(x)[h,w] = \frac{ \max\!\big(0,\tilde{\mathcal{W}}(x)[h,w]\big) }{ \max_{h',w'} \max\!\big(0,\tilde{\mathcal{W}}(x)[h',w']\big) + \varepsilon }
\label{eq:weight_map}
\end{equation}
This construction yields a single spatial weight map in which larger values correspond to regions that are more relevant to the detector’s retained localisation outputs, and can therefore be used to prioritise input perturbations during curriculum generation.

\subsection{Uncertainty Guided Training}
\label{sec:Uncertainty Guided Training}

The uncertainty-guided training stage of \methodname\ constructs and trains a small evidential meta-model $g_\phi$ exclusively on features extracted from the selected layers $L_{\text{sal}}$ and guides it on when to be uncertain in the localisation of retained predicted detections. For each selected layer $\ell \in L_{\text{sal}}$ and retained predicted detection $\hat{j} \in \{1,\dots,\hat{J}\}$, the corresponding feature map is pooled over the predicted box region using ROIAlign, yielding a box-conditioned representation $\Phi_\ell(x,\hat{b}_{\hat{j}})$. 
This representation is then flattened and projected to $\mathbb{R}^{12}$ using a branch of multiple linear layers $g_\ell$, such that $g_\ell(\Phi_\ell(x,\hat{b}_{\hat{j}})) \in \mathbb{R}^{12}$. The output is partitioned into three 4-dimensional vectors, i.e., $v_{\hat{j}}(x)$, $\alpha_{\hat{j}}(x)$, and $\beta_{\hat{j}}(x)$, corresponding to the four localisation channels. The resulting meta-features from all selected layers, $\{g_\ell(\Phi_\ell(x,\hat{b}_{\hat{j}}))\}_{\ell \in L_{\text{sal}}}$, are concatenated into a single vector $\Psi(x,\hat{b}_{\hat{j}})$ and passed through a final head $g_{\mathrm{out}}$, yielding a 12-dimensional output that is partitioned into localisation uncertainty parameters: the evidential precision $\nu_{\hat{j}}(x) \in \mathbb{R}^4$, the inverse-gamma shape parameter $\alpha_{\hat{j}}(x) \in \mathbb{R}^4$, and the inverse-gamma scale parameter $\beta_{\hat{j}}(x) \in \mathbb{R}^4$. Following standard evidential regression parameterisations, these are constrained to be positive using softplus transformations, with $\alpha_{\hat{j}}(x)$ shifted by $1$ to ensure $\alpha_{\hat{j}}(x) > 1$. The predictive mean is taken directly from the retained predicted box, i.e., $\mu_{\hat{j}}(x)=\hat{b}_{\hat{j}}$; this ensures that the meta-model does not alter the detector’s box predictions and focuses on estimating their uncertainty~\citep{amini2020deep}. This strategy enables the meta-model to represent both aleatoric and epistemic uncertainty for each localisation channel. Specifically, the aleatoric and epistemic uncertainties are given by
\begin{equation}
u^{\mathrm{ale}}_{\hat{j}}(x)=\frac{\beta_{\hat{j}}(x)}{\alpha_{\hat{j}}(x)-1},
\qquad
u^{\mathrm{epi}}_{\hat{j}}(x)=\frac{\beta_{\hat{j}}(x)}{v_{\hat{j}}(x)\big(\alpha_{\hat{j}}(x)-1\big)}
\label{eq:uncertainties-meta-model}
\end{equation}
All $\phi$ parameters of the small evidential meta-model $g_\phi$ are learned from the beginning, while the pretrained detector’s model $f_\theta$ parameters $\theta$ are frozen throughout.

To induce robustness and \textit{guide} the meta-model when to be uncertain, we create a targeted noise-driven curriculum. Firstly, a monotonic exponential schedule is constructed as $s_t = 1 - e^{-\gamma t}$ for $t \in \{0, 1, \ldots, T\}$, where $s_t \in [0,1]$ is the target fraction of corrupted pixels at stage $t$, and $\gamma > 0$ is the rate of noise corruption. The first image ($t=0$) represents the clean view; then, the exponential form ensures fine-grained corruption at early stages (where decision boundaries are sensitive) and coarser granularity at higher noise levels (where decision boundaries should be more sharply defined).

To ensure salient features are targeted first, we leverage the weight maps $\mathcal{W}(x)$ from the saliency calibration stage. A global saliency budget $\tilde{\mathcal{W}} = \frac{1}{HW}\sum_{h,w}\mathcal{W}(x)[h,w]$ is defined to construct a per-pixel corruption probability $p_t(h,w)$ satisfying two conditions: the expected global corruption matches the budget $s_t$, and pixel-wise probabilities are monotonic in $s_t$. This is defined as:
\begin{equation}
p_t(h,w) =
\begin{cases}
\frac{s_t}{\tilde{\mathcal{W}}} \cdot \mathcal{W}(x)[h,w], & s_t \leq \tilde{\mathcal{W}}, \\[4pt]
\mathcal{W}(x)[h,w] + \frac{s_t - \tilde{\mathcal{W}}}{1-\tilde{\mathcal{W}}} \cdot (1 - \mathcal{W}(x)[h,w]), & s_t > \tilde{\mathcal{W}}
\end{cases}
\label{eq:s&p_noise_prob}
\end{equation}
This ensures that low-noise perturbations are focused on high-saliency regions, while high-noise settings affect the entire image. To actually apply the noise, we sample a single base mask $m \sim \emph{U}[0,1]^{H\times W}$, reused across all $t$ for a given data point. For each stage $t$, we modify $x$ and obtain $\tilde{x}_t$:
\begin{equation}
\tilde{x}_t[h,w] =
\begin{cases}
0, & m[h,w] < p_t(h,w)/2, \\
1, & m[h,w] > 1 - p_t(h,w)/2, \\
x[h,w], & \text{otherwise.}
\end{cases}
\label{eq:corrupted_image}
\end{equation}
This procedure creates stochastic binary corruption, preserving the expected noise budget per image and ensuring that $\tilde{x}_{t+1}$ is never less noisy than $\tilde{x}_t$.

To guide the meta-model on when to express uncertainty, we construct scalar uncertainty targets for retained predicted detections that respond to both input corruption and localisation degradation. This part is important to steer the model to be confident for clean inputs ($t=0$) and increasingly uncertain for strongly corrupted inputs. For each corrupted view and each retained predicted detection $\hat{j} \in \{1,\dots,\hat{J}\}$, we define three complementary terms. First, we measure the localisation uncertainty of the detection with respect to the ground truth:
\begin{equation}
q^{\mathrm{gt}}_{\hat{j},t}
=
1-\max_{j \in \{1,\dots,J\}}
\mathrm{IoU}\!\left(\bar{b}_{\hat{j},t}, b_j\right)
\label{eq:qgt_det}
\end{equation}
Here, $\bar{b}_{\hat{j},t}$ denotes the pooled box associated with retained prediction $\hat{j}$ at stage $t$. This term serves as a proxy for localisation error: it is small when the pooled box aligns well with a ground-truth object and large when the prediction is poorly localised. 
By using $q^{\mathrm{gt}}_{\hat{j},t}$, \methodname\ ensures that uncertainty increases not only by corruption, but also by intrinsically inaccurate detections. Second, we measure how strongly the detection region is affected by the applied corruption:
\begin{equation}
u^{\mathrm{corr}}_{\hat{j},t}
=
\frac{1}{|\Omega(\bar{b}_{\hat{j},t})|}
\sum_{(h,w)\in\Omega(\bar{b}_{\hat{j},t})} p_t(h,w),
\label{eq:ucorr_det}
\end{equation}
where $\Omega(\bar{b}_{\hat{j},t})$ denotes the set of pixels inside the pooled box region. This term is the average corruption probability inside the region used by the meta-model for that detection, and therefore quantifies its expected corruption exposure. Including $u^{\mathrm{corr}}_{\hat{j},t}$ encourages higher uncertainty when the relevant image region has been heavily perturbed, even before substantial localisation failure is observed. Third, we measure the stability of the detection under corruption:
\begin{equation}
u^{\mathrm{stab}}_{\hat{j},t}
=
1-\mathrm{IoU}\!\left(\hat{b}_{\hat{j}}, \hat{b}^{\,\mathrm{corr}}_{\hat{j},t}\right),
\label{eq:ustab_det}
\end{equation}
where $\hat{b}_{\hat{j}}$ is the clean retained predicted box and $\hat{b}^{\,\mathrm{corr}}_{\hat{j},t}$ is its matched corrupted-view prediction; if no valid match is found, we set $u^{\mathrm{stab}}_{\hat{j},t}=1$. This term captures prediction sensitivity: even if a detection is initially accurate, it should be treated as uncertain if small or moderate corruption causes its location to drift substantially or disappear altogether. These terms are combined into a single uncertainty target:
\begin{equation}
u^{\star}_{\hat{j},t}
=
1-\big(1-q^{\mathrm{gt}}_{\hat{j},t}\big)\big(1-u^{\mathrm{corr}}_{\hat{j},t}\big)\big(1-u^{\mathrm{stab}}_{\hat{j},t}\big),
\label{eq:utgt_det}
\end{equation}
which behaves as a union-style aggregation. Consequently, $u^{\star}_{\hat{j},t}$ becomes large whenever any one of the detections is poorly localised, strongly corrupted, or unstable under corruption, while remaining small only when the detection is simultaneously accurate, lightly corrupted, and stable. This provides a single scalar curriculum target that reflects multiple sources of localisation uncertainty in a monotonic and interpretable way.

Next, we generate a curriculum over corruption strengths. Instead of training the meta-model on all strengths uniformly, we define an epoch-dependent difficulty index $\rho_e = \big(\frac{e}{E-1}\big)^2$ for every epoch $e \in \{0, 1, \ldots, E-1\}$. 
At each epoch $e$, the sampling distribution over the discrete noise levels \simos{$s \in \{ s_0, \ldots, s_T \}$} is given by $\kappa_e(s) \propto (1-\rho_e)(1-s) + \rho_e \cdot s$ and determines which corruption levels populate the training batches. In early epochs ($\rho_e \approx 0$), the sampling distribution concentrates mass on small $s$, so a higher proportion of clean or mildly corrupted views are trained upon. In late epochs ($\rho_e \to 1$), the distribution shifts towards large $s$; evidently, strongly corrupted views dominate the training data. Thus, the curriculum does not require each individual image to pass through every corruption level in sequence. Instead, it gradually shifts the sampling distribution from clean and mildly corrupted views in early epochs toward more strongly corrupted views in later epochs. This exposes the meta-model to increasingly difficult inputs while preserving diversity across training.

Finally, the meta-model is trained using a curriculum-weighted evidential regression objective for all ground-truth matched detections. For each matched retained predicted detection $\hat{j}$, let $\mu_{\hat{j}} \in \mathbb{R}^4$ denote the clean retained predicted box, $y_{\hat{j}} \in \mathbb{R}^4$ the matched ground-truth box, and $v_{\hat{j}}, \alpha_{\hat{j}}, \beta_{\hat{j}} \in \mathbb{R}^4$ the evidential parameters predicted by the meta-model. To account for box scale, both $\mu_{\hat{j}}$ and $y_{\hat{j}}$ are normalised coordinate-wise by the width and height of $\mu_{\hat{j}}$, yielding $\bar{\mu}_{\hat{j}}$ and $\bar{y}_{\hat{j}}$. For a mini-batch $\mathcal{B}$ of matched retained detections, the loss is defined as:
\begin{equation}
\mathcal{L}
=
\frac{\sum_{\hat{j}\in\mathcal{B}} (1-u^{\star}_{\hat{j},t})\,\mathcal{L}^{\mathrm{NLL}}_{\hat{j}}}{\sum_{\hat{j}\in\mathcal{B}} (1-u^{\star}_{\hat{j},t}) + \varepsilon}
+
\lambda_{\mathrm{reg}}
\frac{\sum_{\hat{j}\in\mathcal{B}} (1+u^{\star}_{\hat{j},t})\,\mathcal{L}^{\mathrm{REG}}_{\hat{j}}}{\sum_{\hat{j}\in\mathcal{B}} (1+u^{\star}_{\hat{j},t}) + \varepsilon}
\label{eq:guide_det_loss}
\end{equation}
where $\lambda_{\mathrm{reg}}$ trades off uncertainty inflation with model fit~\citep{amini2020deep} and $\mathcal{L}^{\mathrm{NLL}}_{\hat{j}}$ is the summed NIG negative log-likelihood over the four localisation channels and $\mathcal{L}^{\mathrm{REG}}_{\hat{j}}$ is the corresponding evidential regulariser. Specifically:
\begin{equation}
\begin{aligned}
\mathcal{L}^{\mathrm{NLL}}_{\hat{j}}
&=
\sum_{q=1}^{4}
\bigg[
\frac{1}{2}\log\!\Big(\frac{\pi}{v_{\hat{j},q}}\Big)
-
\alpha_{\hat{j},q}\log\!\big(2\beta_{\hat{j},q}(1+v_{\hat{j},q})\big)
+
\Big(\alpha_{\hat{j},q}+\frac{1}{2}\Big) \\
&\cdot\log\!\Big(v_{\hat{j},q}(\bar{y}_{\hat{j},q}-\bar{\mu}_{\hat{j},q})^2 + 2\beta_{\hat{j},q}(1+v_{\hat{j},q})\Big)
+
\log\Gamma(\alpha_{\hat{j},q})
-
\log\Gamma\!\Big(\alpha_{\hat{j},q}+\frac{1}{2}\Big)
\bigg]
\label{eq:nig_nll_det}
\end{aligned}
\end{equation}
and
\begin{equation}
\mathcal{L}^{\mathrm{REG}}_{\hat{j}}
=
\sum_{q=1}^{4}
\left|
\frac{\bar{y}_{\hat{j},q}-\bar{\mu}_{\hat{j},q}}
{\sqrt{\beta_{\hat{j},q}(1+v_{\hat{j},q})/(\alpha_{\hat{j},q}v_{\hat{j},q})}}
\right|
\bigl(2v_{\hat{j},q}+\alpha_{\hat{j},q}\bigr)
\label{eq:nig_reg_det}
\end{equation}
The curriculum target $u^{\star}_{\hat{j},t}$ modulates these two terms differently. Detections with small $u^{\star}_{\hat{j},t}$, corresponding to accurate, stable, and lightly corrupted examples, receive larger weight in the regression fit term, encouraging precise localisation where the detector is reliable. Conversely, detections with large $u^{\star}_{\hat{j},t}$, corresponding to poorly localised, strongly corrupted, or unstable examples, receive larger weight in the evidential regulariser, discouraging unjustified confidence in uncertain regions of the input space. These weights are normalised over the matched detections, so that the loss remains on a comparable scale while the curriculum rebalances which detections contribute most strongly. Combined, these two terms guide the meta-model to remain confident only when localisation is well supported and to become increasingly uncertain as corruption and localisation degradation increase.

\section{Experimental Evaluation}
\label{sec:Experimental Evaluation}
We evaluate \methodname\ in a comprehensive set of experiments comparing against state-of-the-art approaches across multiple tasks and models. 
Implementation details, theoretical analysis, experimental setup, additional experiments, and extended discussion are provided in Appendices~\ref{sec:appendix-Implementation Details},~\ref{sec:appendix-Theoretical Analysis},~\ref{sec:appendix-Additional Experiments},~\ref{sec:appendix-Additional Discussion}, and~\ref{sec:appendix-Experimental Setup}.

\textbf{Comparative Approaches.} We compare \methodname\ against the base model, EMM~\citep{shen2023post}, ModelNet~\citep{lee2022lightweight}, and MetaDetect~\citep{schubert2021metadetect} to represent a range of approaches in post-hoc UQ for object detection that allow for a fair comparison.

\textbf{Datasets.} 
Adopting a similar procedure from recent research~\citep{schubert2021metadetect}, we evaluate all approaches on the COCO~\citep{lin2014microsoft}, CelebA~\citep{liu2015faceattributes}, VisDrone~\citep{zhu2021detection}, and TT100k~\citep{Zhe_2016_CVPR} datasets to cover a diverse set of domains, tasks, and challenges.

\subsection{Core Results}
\label{sec:Core Results}

For our core experiments, we evaluate the ability of each uncertainty approach to identify unreliable detections and rank them accordingly. We consider TP-FP AUROC and FP AUPRC for incorrect-detection identification, risk-coverage and IoU-coverage for selective prediction, and the Spearman correlation between uncertainty and detection IoU. These results are summarised in Table~\ref{tab:main_results}.

\begin{figure*}[tb]
    \centering
    \includegraphics[width=\linewidth]{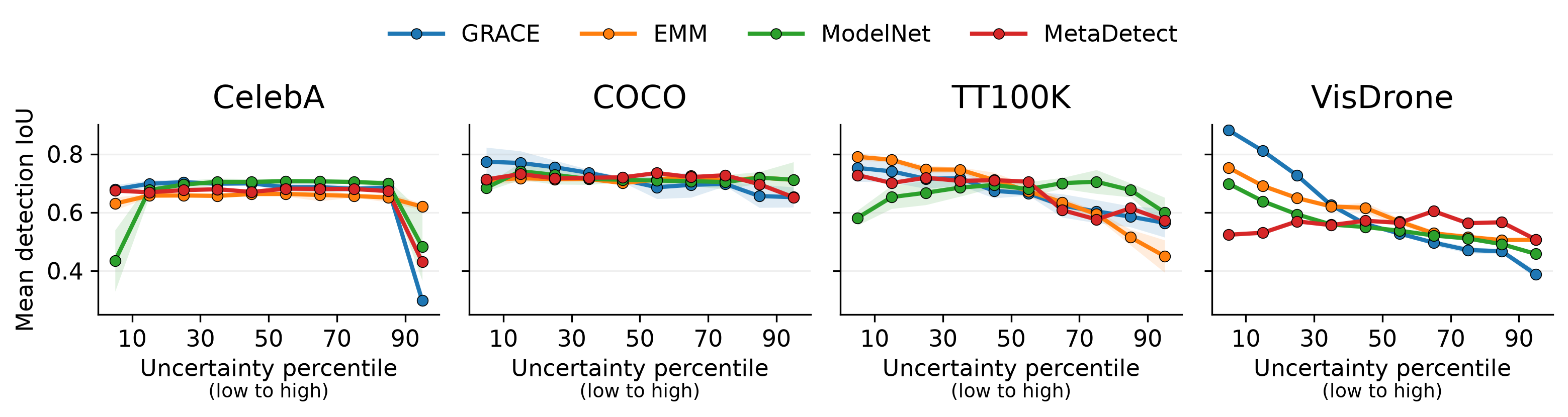}
    \caption{Mean detection IoU across equal-count increasing uncertainty percentile bins within each run for all datasets. Stronger downward trends and more negative mean Spearman uncertainty–IoU correlations ($r_s$) indicate better uncertainty–localisation alignment.}
    \label{fig:all_datasets_binned_uncertainty_iou_spearman}
\end{figure*}

\begin{table*}[tb]
    \centering
    \caption{Uncertainty performance across four datasets: TP-FP AUROC and FP AUPRC measure incorrect-detection identification; AURC and IoU-coverage AUC, selective prediction; uncertainty-IoU correlation, Spearman alignment with localisation quality; and added inference time, overhead relative to the frozen detector. Best results are bold and \colorbox{lightblue}{highlighted}.}
    \label{tab:main_results}
    \small
    \setlength{\tabcolsep}{4pt}
    \resizebox{\textwidth}{!}{%
    \begin{tabular}{lcccccc}
        \toprule
        Method
        & TP-FP AUROC $\uparrow$
        & FP AUPRC $\uparrow$
        & \shortstack{Risk-coverage\\AUC$\downarrow$}
        & \shortstack{IoU-coverage\\AUC $\uparrow$}
        & \shortstack{Uncertainty-IoU\\correlation $\downarrow$}
        & \shortstack{Added inference\\time (ms) $\downarrow$} \\
        \midrule
        \multicolumn{7}{c}{\textbf{CelebA}} \\
        \midrule
        \methodname
        & \cellcolor{lightblue}$\mathbf{0.83 \pm 0.02}$
        & \cellcolor{lightblue}$\mathbf{0.72 \pm 0.05}$
        & \cellcolor{lightblue}$\mathbf{0.04 \pm 0.01}$
        & \cellcolor{lightblue}$\mathbf{0.69 \pm 0.01}$
        & \cellcolor{lightblue}$\mathbf{-0.36 \pm 0.07}$
        & \cellcolor{lightblue}$\mathbf{2.66 \pm 0.48}$ \\
        EMM
        & $0.50 \pm 0.03$
        & $0.12 \pm 0.02$
        & $0.10 \pm 0.01$
        & $0.64 \pm 0.01$
        & $0.00 \pm 0.05$
        & $11.76 \pm 0.75$ \\
        ModelNet
        & $0.42 \pm 0.18$
        & $0.39 \pm 0.18$
        & $0.20 \pm 0.05$
        & $0.59 \pm 0.04$
        & $0.12 \pm 0.20$
        & $42.93 \pm 3.44$ \\
        MetaDetect
        & $0.74 \pm 0.00$
        & $0.47 \pm 0.00$
        & $0.05 \pm 0.00$
        & $0.67 \pm 0.00$
        & $-0.35 \pm 0.00$
        & $27.00 \pm 15.77$ \\
        \midrule
        \addlinespace
        \multicolumn{7}{c}{\textbf{COCO}} \\
        \midrule
        \methodname
        & \cellcolor{lightblue}$\mathbf{0.55 \pm 0.03}$
        & \cellcolor{lightblue}$\mathbf{0.29 \pm 0.02}$
        & \cellcolor{lightblue}$\mathbf{0.23 \pm 0.02}$
        & \cellcolor{lightblue}$\mathbf{0.75 \pm 0.03}$
        & \cellcolor{lightblue}$\mathbf{-0.17 \pm 0.11}$
        & $8.01 \pm 4.76$ \\
        EMM
        & $0.50 \pm 0.00$
        & $0.25 \pm 0.00$
        & $0.26 \pm 0.00$
        & $0.71 \pm 0.01$
        & $-0.02 \pm 0.04$
        & \cellcolor{lightblue}$\mathbf{2.79 \pm 0.61}$ \\
        ModelNet
        & $0.50 \pm 0.02$
        & $0.26 \pm 0.02$
        & $0.27 \pm 0.01$
        & $0.70 \pm 0.01$
        & $-0.00 \pm 0.09$
        & $49.37 \pm 1.09$ \\
        MetaDetect
        & $0.52 \pm 0.00$
        & $0.28 \pm 0.00$
        & $0.25 \pm 0.00$
        & $0.72 \pm 0.00$
        & $-0.07 \pm 0.00$
        & $20.39 \pm 0.95$ \\
        \midrule
        \addlinespace
        \multicolumn{7}{c}{\textbf{TT100K}} \\
        \midrule
        \methodname
        & $0.53 \pm 0.02$
        & $0.42 \pm 0.01$
        & $0.37 \pm 0.02$
        & $0.72 \pm 0.03$
        & $-0.44 \pm 0.02$
        & $4.30 \pm 0.21$ \\
        EMM
        & \cellcolor{lightblue}$\mathbf{0.57 \pm 0.01}$
        & \cellcolor{lightblue}$\mathbf{0.47 \pm 0.02}$
        & \cellcolor{lightblue}$\mathbf{0.35 \pm 0.00}$
        & \cellcolor{lightblue}$\mathbf{0.75 \pm 0.00}$
        & \cellcolor{lightblue}$\mathbf{-0.48 \pm 0.01}$
        & \cellcolor{lightblue}$\mathbf{4.20 \pm 0.22}$ \\
        ModelNet
        & $0.49 \pm 0.01$
        & $0.41 \pm 0.02$
        & $0.42 \pm 0.01$
        & $0.64 \pm 0.01$
        & $0.02 \pm 0.11$
        & $45.82 \pm 2.71$ \\
        MetaDetect
        & $0.55 \pm 0.00$
        & $0.43 \pm 0.00$
        & $0.37 \pm 0.00$
        & $0.71 \pm 0.00$
        & $-0.17 \pm 0.00$
        & $80.75 \pm 2.00$ \\
        \midrule
        \addlinespace
        \multicolumn{7}{c}{\textbf{VisDrone}} \\
        \midrule
        \methodname
        & \cellcolor{lightblue}$\mathbf{0.66 \pm 0.00}$
        & $0.61 \pm 0.00$
        & \cellcolor{lightblue}$\mathbf{0.35 \pm 0.00}$
        & \cellcolor{lightblue}$\mathbf{0.74 \pm 0.00}$
        & \cellcolor{lightblue}$\mathbf{-0.60 \pm 0.00}$
        & $6.24 \pm 1.14$ \\
        EMM
        & $0.58 \pm 0.00$
        & $0.54 \pm 0.00$
        & $0.41 \pm 0.00$
        & $0.67 \pm 0.00$
        & $-0.36 \pm 0.01$
        & \cellcolor{lightblue}$\mathbf{6.16 \pm 0.76}$ \\
        ModelNet
        & $0.63 \pm 0.00$
        & \cellcolor{lightblue}$\mathbf{0.63 \pm 0.00}$
        & $0.43 \pm 0.01$
        & $0.62 \pm 0.00$
        & $-0.27 \pm 0.01$
        & $80.18 \pm 0.94$ \\
        MetaDetect
        & $0.51 \pm 0.00$
        & $0.53 \pm 0.00$
        & $0.54 \pm 0.00$
        & $0.55 \pm 0.00$
        & $-0.07 \pm 0.00$
        & $84.58 \pm 1.66$ \\
        \bottomrule
    \end{tabular}%
}
\end{table*}

\methodname\ performs best overall on CelebA, COCO and VisDrone. On CelebA, it achieves TP-FP AUROC of $0.83 \pm 0.02$ with the lowest inference overhead of $2.66 \pm 0.48\mathrm{ms}$. On COCO, it leads every uncertainty-quality metric, including TP-FP AUROC of $0.55 \pm 0.03$, although EMM is faster with substantially weaker uncertainty-IoU alignment. On VisDrone, it leads four uncertainty metrics, including uncertainty-IoU correlation of $-0.60 \pm 0.00$; ModelNet has slightly higher FP AUPRC and EMM marginally lower overhead. TT100K is the main exception: EMM leads the uncertainty metrics, achieving uncertainty-IoU correlation of $-0.48 \pm 0.01$ compared with $-0.44 \pm 0.02$ for \methodname, which remains competitive across all metrics with similar overhead. Figure~\ref{fig:all_datasets_binned_uncertainty_iou_spearman} shows mean IoU decreasing with increasing uncertainty for \methodname\ across all four datasets, particularly VisDrone, indicating that higher uncertainty generally corresponds to poorer localisation. EMM also shows strong ordering on TT100K. ModelNet frequently exhibits flatter trends, indicating weaker uncertainty-localisation alignment.

\begin{figure*}[tb]
    \centering
    \includegraphics[width=\linewidth]{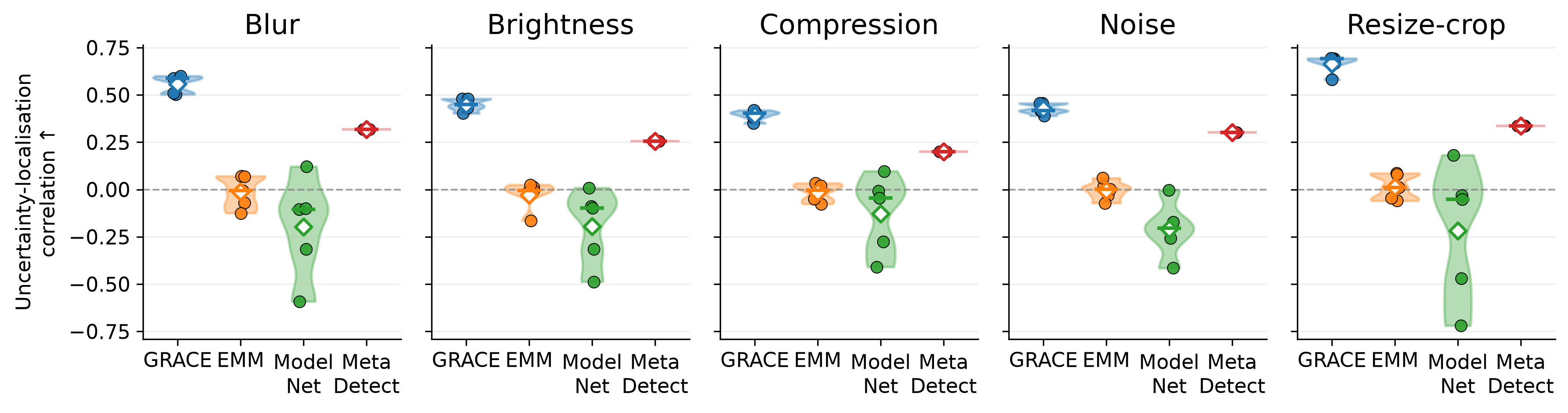}
    \caption{CelebA between uncertainty assigned to clean detections and their localisation error under five perturbations at severity $t=8$. Violins show the distribution across five runs, dots individual runs, diamonds means, solid horizontal lines medians, and the dashed line zero correlation.}
    \label{fig:celeba-pertu-violins}
\end{figure*}

\subsection{Covariate Shift Results}
\label{sec:Covariate Shift Results}

We next evaluate whether uncertainty assigned to clean detections can anticipate localisation degradation under covariate shift. Figure~\ref{fig:celeba-pertu-violins} visualises this relationship under five perturbations at severity $t=8$. \methodname\ achieves consistently positive uncertainty-localisation error correlations, with mean Pearson $r$ ranging from approximately $0.37$ under compression to $0.65$ under resize-crop, and around $0.56$ under blur. In contrast, EMM remains close to zero across most perturbations, while ModelNet is frequently negatively correlated.

\begin{figure*}[tb]
    \centering
    \includegraphics[width=\linewidth]{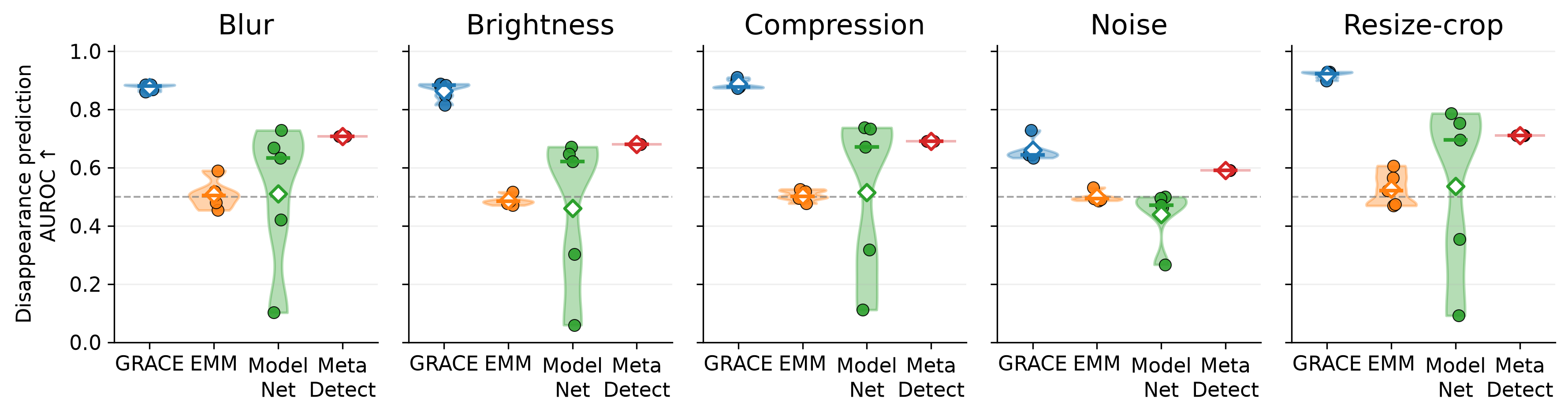}
    \caption{CelebA disappearance-prediction AUROC using uncertainty assigned to clean detections under five perturbations at severity $t=8$. Violins show the distribution across five runs, dots individual runs, diamonds means, solid horizontal lines medians, and the dashed line zero correlation.}
    \label{fig:celeba-disp-violins}
\end{figure*}

Figure~\ref{fig:celeba-disp-violins} considers the stronger failure case of whether a clean detection subsequently disappears under perturbation. \methodname\ achieves the highest disappearance-prediction AUROC across all five shifts, reaching approximately $0.87$-$0.91$ for blur, brightness, compression, and resize-crop, and around $0.65$ under noise. The comparative approaches generally remain substantially lower, with MetaDetect typically around $0.59$-$0.72$ and EMM close to $0.50$. These results show that \methodname\ uncertainty anticipates both localisation degradation and detection failure under unseen covariates.

\subsection{Adversarial Attack Results}
\label{sec:Adversarial Attack Results}

\begin{figure*}[tb]
    \centering
    \includegraphics[width=\linewidth]{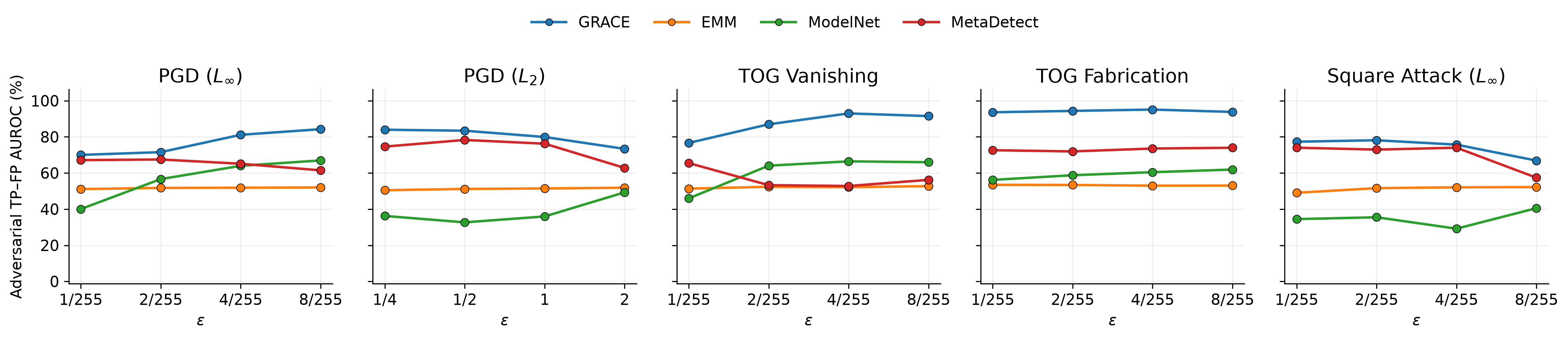}
    \caption{CelebA TP–FP AUROC under five adversarial attacks across increasing attack strength $\epsilon$. Lines show mean performance across five runs for all methods; higher AUROC indicates better discrimination between TP and FP detections under adversarial perturbation.}
    \label{fig:celeba-adv-attacks}
\end{figure*}

We further evaluate uncertainty robustness under five adversarial attacks across increasing perturbation strengths. As shown in Figure~\ref{fig:celeba-adv-attacks}, \methodname\ provides the strongest and most consistent TP-FP separation on CelebA across black-box and white-box attacks. Performance remains particularly strong under the TOG attacks, with TP-FP AUROC remaining around $0.90$ or above across the evaluated strengths, while the comparative approaches generally remain substantially lower.

Across attacks and strengths, \methodname\ achieves adversarial TP-FP AUROC of $0.825 \pm 0.020$, outperforming MetaDetect ($0.676 \pm 0.000$), EMM ($0.519 \pm 0.028$), and ModelNet ($0.501 \pm 0.241$), alongside the highest adversarial FP AUPRC ($0.850 \pm 0.025$). This suggests its uncertainty remains informative under attacks substantially different from the corruption used in curriculum training.

\section{Conclusions}
\label{sec:Conclusions}

We introduced \methodname, a lightweight post-hoc uncertainty method for object detection that learns when predicted localisations are unreliable while keeping the base detector frozen. Across four datasets, \methodname\ generally provides stronger localisation-aware uncertainty than comparative approaches and remains effective under covariate shift and adversarial attacks. Future work will look into video object detection by incorporating temporal information into uncertainty estimation.


\bibliography{iclr2027_conference}

@inproceedings{lin2014microsoft,
  title={Microsoft coco: Common objects in context},
  author={Lin, Tsung-Yi and Maire, Michael and Belongie, Serge and Hays, James and Perona, Pietro and Ramanan, Deva and Doll{\'a}r, Piotr and Zitnick, C Lawrence},
  booktitle={European conference on computer vision},
  pages={740--755},
  year={2014},
  organization={Springer}
}

@inproceedings{liu2015faceattributes,
  title = {Deep Learning Face Attributes in the Wild},
  author = {Liu, Ziwei and Luo, Ping and Wang, Xiaogang and Tang, Xiaoou},
  booktitle = {Proceedings of International Conference on Computer Vision (ICCV)},
  month = {December},
  year = {2015} 
}

@article{zhu2021detection,
  title={Detection and tracking meet drones challenge},
  author={Zhu, Pengfei and Wen, Longyin and Du, Dawei and Bian, Xiao and Fan, Heng and Hu, Qinghua and Ling, Haibin},
  journal={IEEE Transactions on Pattern Analysis and Machine Intelligence},
  volume={44},
  number={11},
  pages={7380--7399},
  year={2021},
  publisher={IEEE}
}

@InProceedings{Zhe_2016_CVPR,
    author = {Zhu, Zhe and Liang, Dun and Zhang, Songhai and Huang, Xiaolei and Li, Baoli and Hu, Shimin},
    title = {Traffic-Sign Detection and Classification in the Wild},
    booktitle = {The IEEE Conference on Computer Vision and Pattern Recognition (CVPR)},
    year = {2016}
}

@inproceedings{franchi2024make,
  title={Make me a bnn: A simple strategy for estimating bayesian uncertainty from pre-trained models},
  author={Franchi, Gianni and Laurent, Olivier and Legu{\'e}ry, Maxence and Bursuc, Andrei and Pilzer, Andrea and Yao, Angela},
  booktitle={2024 IEEE/CVF Conference on Computer Vision and Pattern Recognition (CVPR)},
  pages={12194--12204},
  year={2024},
  organization={IEEE}
}

@article{ovadia2019can,
  title={Can you trust your model's uncertainty? evaluating predictive uncertainty under dataset shift},
  author={Ovadia, Yaniv and Fertig, Emily and Ren, Jie and Nado, Zachary and Sculley, David and Nowozin, Sebastian and Dillon, Joshua and Lakshminarayanan, Balaji and Snoek, Jasper},
  journal={Advances in neural information processing systems},
  volume={32},
  year={2019}
}

@inproceedings{hall2020probabilistic,
  title={Probabilistic object detection: Definition and evaluation},
  author={Hall, David and Dayoub, Feras and Skinner, John and Zhang, Haoyang and Miller, Dimity and Corke, Peter and Carneiro, Gustavo and Angelova, Anelia and S{\"u}nderhauf, Niko},
  booktitle={2020 IEEE Winter Conference on Applications of Computer Vision (WACV)},
  pages={1020--1029},
  year={2020},
  organization={IEEE}
}

@article{liu2020deep,
  title={Deep learning for generic object detection: A survey},
  author={Liu, Li and Ouyang, Wanli and Wang, Xiaogang and Fieguth, Paul and Chen, Jie and Liu, Xinwang and Pietik{\"a}inen, Matti},
  journal={International journal of computer vision},
  volume={128},
  number={2},
  pages={261--318},
  year={2020},
  publisher={Springer}
}

@article{abdar2021review,
  title={A review of uncertainty quantification in deep learning: Techniques, applications and challenges},
  author={Abdar, Moloud and Pourpanah, Farhad and Hussain, Sadiq and Rezazadegan, Dana and Liu, Li and Ghavamzadeh, Mohammad and Fieguth, Paul and Cao, Xiaochun and Khosravi, Abbas and Acharya, U Rajendra and others},
  journal={Information fusion},
  volume={76},
  pages={243--297},
  year={2021},
  publisher={Elsevier}
}

@inproceedings{shen2023post,
  title={Post-hoc uncertainty learning using a dirichlet meta-model},
  author={Shen, Maohao and Bu, Yuheng and Sattigeri, Prasanna and Ghosh, Soumya and Das, Subhro and Wornell, Gregory},
  booktitle={Proceedings of the AAAI Conference on Artificial Intelligence},
  volume={37},
  pages={9772--9781},
  year={2023}
}

@inproceedings{schubert2021metadetect,
  title={Metadetect: Uncertainty quantification and prediction quality estimates for object detection},
  author={Schubert, Marius and Kahl, Karsten and Rottmann, Matthias},
  booktitle={2021 international joint conference on neural networks (IJCNN)},
  pages={1--10},
  year={2021},
  organization={IEEE}
}

@inproceedings{lee2022lightweight,
  title={Lightweight model uncertainty estimation for deep neural object detection},
  author={Lee, Minah and Mudassar, Burhan and Mukhopadhyay, Saibal},
  booktitle={2022 International Joint Conference on Neural Networks (IJCNN)},
  pages={1--8},
  year={2022},
  organization={IEEE}
}

@article{khanam2024yolov11,
  title={Yolov11: An overview of the key architectural enhancements},
  author={Khanam, Rahima and Hussain, Muhammad},
  journal={arXiv preprint arXiv:2410.17725},
  year={2024}
}

@article{madry2018towards,
  title={Towards deep learning models resistant to adversarial attacks},
  author={Madry, Aleksander and Makelov, Aleksandar and Schmidt, Ludwig and Tsipras, Dimitris and Vladu, Adrian},
  journal={arXiv preprint arXiv:1706.06083},
  year={2017}
}

@article{chow2020adversarial,
  title={TOG: targeted adversarial objectness gradient attacks on real-time object detection systems},
  author={Chow, Ka-Ho and Liu, Ling and Gursoy, Mehmet Emre and Truex, Stacey and Wei, Wenqi and Wu, Yanzhao},
  journal={arXiv preprint arXiv:2004.04320},
  year={2020}
}

@inproceedings{andriushchenko2020square,
  title={Square attack: a query-efficient black-box adversarial attack via random search},
  author={Andriushchenko, Maksym and Croce, Francesco and Flammarion, Nicolas and Hein, Matthias},
  booktitle={European conference on computer vision},
  pages={484--501},
  year={2020},
  organization={Springer}
}

@article{he2023survey,
  title={A survey on uncertainty quantification methods for deep neural networks: An uncertainty source perspective},
  author={He, Wenchong and Jiang, Z},
  journal={Perspective},
  volume={1},
  pages={88},
  year={2023}
}

@article{fortuin2022priors,
  title={Priors in bayesian deep learning: A review},
  author={Fortuin, Vincent},
  journal={International Statistical Review},
  volume={90},
  number={3},
  pages={563--591},
  year={2022},
  publisher={Wiley Online Library}
}

@article{blei2017variational,
  title={Variational inference: A review for statisticians},
  author={Blei, David M and Kucukelbir, Alp and McAuliffe, Jon D},
  journal={Journal of the American Statistical Association},
  volume={112},
  number={518},
  pages={859--877},
  year={2017},
  publisher={Taylor \& Francis}
}

@article{lakshminarayanan2017simple,
  title={Simple and scalable predictive uncertainty estimation using deep ensembles},
  author={Lakshminarayanan, Balaji and Pritzel, Alexander and Blundell, Charles},
  journal={Advances in Neural Information Processing Systems},
  volume={30},
  year={2017}
}

@article{liu2020simple,
  title={Simple and principled uncertainty estimation with deterministic deep learning via distance awareness},
  author={Liu, Jeremiah and Lin, Zi and Padhy, Shreyas and Tran, Dustin and Bedrax Weiss, Tania and Lakshminarayanan, Balaji},
  journal={Advances in Neural Information Processing Systems},
  volume={33},
  pages={7498--7512},
  year={2020}
}

@article{kendall2017uncertainties,
  title={What uncertainties do we need in bayesian deep learning for computer vision?},
  author={Kendall, Alex and Gal, Yarin},
  journal={Advances in Neural Information Processing Systems},
  volume={30},
  year={2017}
}

@article{letzgus2022toward,
  title={Toward explainable artificial intelligence for regression models: A methodological perspective},
  author={Letzgus, Simon and Wagner, Patrick and Lederer, Jonas and Samek, Wojciech and M{\"u}ller, Klaus-Robert and Montavon, Gregoire},
  journal={IEEE Signal Processing Magazine},
  volume={39},
  number={4},
  pages={40--58},
  year={2022},
  publisher={IEEE}
}

@article{tagasovska2019single,
  title={Single-model uncertainties for deep learning},
  author={Tagasovska, Natasa and Lopez-Paz, David},
  journal={Advances in Neural Information Processing Systems},
  volume={32},
  year={2019}
}

@article{amini2020deep,
  title={Deep evidential regression},
  author={Amini, Alexander and Schwarting, Wilko and Soleimany, Ava and Rus, Daniela},
  journal={Advances in neural information processing systems},
  volume={33},
  pages={14927--14937},
  year={2020}
}

@misc{kingma2013auto,
  title={Auto-encoding variational bayes},
  author={Kingma, Diederik P and Welling, Max and others},
  year={2013},
  publisher={Banff, Canada}
}

@inproceedings{gal2016dropout,
  title={Dropout as a bayesian approximation: Representing model uncertainty in deep learning},
  author={Gal, Yarin and Ghahramani, Zoubin},
  booktitle={international cCnference on Machine Learning},
  pages={1050--1059},
  year={2016},
  organization={PMLR}
}

@inproceedings{kuppers2020multivariate,
  title={Multivariate Confidence Calibration for Object Detection},
  author={Kuppers, Fabian and Kronenberger, Jan and Shantia, Amirhossein and Haselhoff, Anselm},
  booktitle={Proceedings of the IEEE/CVF Conference on Computer Vision and Pattern Recognition Workshops},
  year={2020}
}

@inproceedings{barker2026gradual,
  title={Gradual Uncertainty Refinement via Noise-Driven Curriculum: A Post-Hoc Meta-Model for Robust Uncertainty Quantification},
  author={Barker, Charmaine and Bethell, Daniel and Gerasimou, Simos},
  booktitle={Forty-Second Annual Conference on Uncertainty in Artificial Intelligence},
  year={2026}
}

@article{angelopoulos2023conformal,
  title={Conformal prediction: A gentle introduction},
  author={Angelopoulos, Anastasios N and Bates, Stephen and others},
  journal={Foundations and Trends in Machine Learning},
  volume={16},
  number={4},
  pages={494--591},
  year={2023},
  publisher={Now Publishers, Inc.}
}

@inproceedings{ayhan2018test,
  title={Test-time data augmentation for estimation of heteroscedastic aleatoric uncertainty in deep neural networks},
  author={Ayhan, Murat Seckin and Berens, Philipp},
  booktitle={Medical Imaging with Deep Learning},
  year={2018}
}

@inproceedings{pearce2018high,
  title={High-quality prediction intervals for deep learning: A distribution-free, ensembled approach},
  author={Pearce, Tim and Brintrup, Alexandra and Zaki, Mohamed and Neely, Andy},
  booktitle={International Conference on Machine Learning},
  pages={4075--4084},
  year={2018},
  organization={PMLR}
}

@article{goan2020bayesian,
  title={Bayesian neural networks: An introduction and survey},
  author={Goan, Ethan and Fookes, Clinton},
  journal={Case Studies in Applied Bayesian Data Science: CIRM Jean-Morlet Chair, Fall 2018},
  pages={45--87},
  year={2020},
  publisher={Springer}
}

@inproceedings{bethell2024robust,
  title={Robust uncertainty quantification using conformalised Monte Carlo prediction},
  author={Bethell, Daniel and Gerasimou, Simos and Calinescu, Radu},
  booktitle={Proceedings of the AAAI conference on artificial intelligence},
  volume={38},
  pages={20939--20948},
  year={2024}
}
\bibliographystyle{iclr2027_conference}


\setlength{\parindent}{0pt}
\setlength{\parskip}{0.5em}
\newpage
\onecolumn
\appendix

\section*{Appendix Contents}
\begin{center}
\begin{minipage}{0.5\textwidth}
Implementation Details \dotfill \pageref{sec:appendix-Implementation Details}\\
Theoretical Analysis \dotfill \pageref{sec:appendix-Theoretical Analysis}\\
Additional Experiments \dotfill \pageref{sec:appendix-Additional Experiments}\\
Additional Discussion \dotfill \pageref{sec:appendix-Additional Discussion}\\
Experimental Setup \dotfill \pageref{sec:appendix-Experimental Setup}\\
\end{minipage}
\end{center}
\setcounter{secnumdepth}{1}


\section{Implementation Details}
\label{sec:appendix-Implementation Details}
Algorithm~\ref{alg:1} summarises the full \methodname\ procedure. The pretrained detector remains frozen throughout; only the parameters of the evidential meta-model are updated.

\begin{algorithm}[htb]
\caption{\textsc{\methodname}: Guided evidential Regression for Adversarial and Covariate uncertainty Estimation}
\label{alg:1}
\begin{algorithmic}[1]
\Require Frozen pretrained detector $f_\theta$; calibration data $\mathcal{D}_{\mathrm{cal}}$; relevance coverage threshold $\eta$; curriculum steps $T$; corruption rate $\gamma$; training epochs $E$
\Ensure Trained evidential meta-model $g_\phi$ and selected salient layers $L_{\mathrm{sal}}$
\State Freeze $f_\theta$ and set it to inference mode
\For{each $(x_i,y_i) \in \mathcal{D}_{\mathrm{cal}}$}
\State Compute retained detections $\{\hat{b}_{\hat{j}}\}_{\hat{j}=1}^{\hat{J}}$ and initialise output relevance $R_L(x_i)=\sum_{\hat{j}=1}^{\hat{J}}\sum_{q=1}^{4} r_{i_{\hat{j}},q}(x_i)$
\State Propagate relevance through $f_\theta$ using the LRP-$\epsilon$ rule
\State Accumulate layer relevance scores $M_\ell$ using Eq.~\ref{eq:layer_rel_mass}
\State Store the localisation-based input weight map $\mathcal{W}(x_i)$ using Eq.~\ref{eq:weight_map}
\EndFor
\State Select the smallest salient layer set $L_{\mathrm{sal}}$ satisfying Eq.~\ref{eq:coverage-constraint}
\State Construct $g_\phi$ using projection branches from layers in $L_{\mathrm{sal}}$ and a Normal-Inverse-Gamma output head
\For{epoch $e=0,\ldots,E-1$}
\State Set curriculum difficulty $\rho_e = \big(e/(E-1)\big)^2$
\For{each mini-batch $\mathcal{B} \subset \mathcal{D}_{\mathrm{cal}}$}
\For{each $(x,y) \in \mathcal{B}$}
\State Sample a curriculum stage $t \in \{0,\ldots,T\}$ using $\kappa_e(s) \propto (1-\rho_e)(1-s)+\rho_e s$
\State Compute the corruption level $s_t = 1-\exp(-\gamma t)$
\State Compute saliency-weighted corruption probabilities $p_t(h,w)$ using Eq.~\ref{eq:s&p_noise_prob}
\State Corrupt $x$ with the fixed base mask rule in Eq.~\ref{eq:corrupted_image} to obtain $\tilde{x}_t$
\State Match retained clean detections $\hat{b}_{\hat{j}}$ to detections $\hat{b}^{\mathrm{corr}}_{\hat{j},t}$ from $f_\theta(\tilde{x}_t)$
\State Form detection-level targets $u^\star_{\hat{j},t}
=1-(1-q^{\mathrm{gt}}_{\hat{j},t})(1-u^{\mathrm{corr}}_{\hat{j},t})(1-u^{\mathrm{stab}}_{\hat{j},t})$
\EndFor
\State Extract box-conditioned frozen features $\{\Phi_\ell(\tilde{x}_t,\bar{b}_{\hat{j},t})\}_{\ell \in L_{\mathrm{sal}}}$ using ROIAlign
\State Predict NIG parameters $(v_{\hat{j}},\alpha_{\hat{j}},\beta_{\hat{j}})
=g_\phi(\{\Phi_\ell(\tilde{x}_t,\bar{b}_{\hat{j},t})\}_{\ell \in L_{\mathrm{sal}}})$, with $\mu_{\hat{j}}=\hat{b}_{\hat{j}}$
\State Update $\phi$ by minimising the \methodname\ loss in Eq.~\ref{eq:guide_det_loss}
\EndFor
\EndFor
\State \Return $g_\phi$, $L_{\mathrm{sal}}$
\end{algorithmic}
\end{algorithm}

For each predicted bounding box $\hat{b}_{j}$, the evidential head outputs the NIG parameters $(v_{j,q},\alpha_{j,q},\beta_{j,q})$ for each coordinate $q\in{x_1,y_1,x_2,y_2}$, while the predictive mean is fixed to the original detector coordinate, $\mu_{j,q}=\hat{b}_{j,q}$. We compute coordinate-wise aleatoric and epistemic uncertainty as:

\begin{equation}
u^{\mathrm{ale}}_{j,q}=\frac{\beta_{j,q}}{\alpha_{j,q}-1},
\qquad
u^{\mathrm{epi}}_{j,q}=\frac{\beta_{j,q}}{v_{j,q}(\alpha_{j,q}-1)}
\end{equation}

The final localisation uncertainty for a detection is the mean total uncertainty across its four coordinates:

\begin{equation}
u_j=\frac{1}{4}\sum_q\left(u^{\mathrm{ale}}_{j,q}+u^{\mathrm{epi}}_{j,q}\right)
\end{equation}

Thus, \methodname\ estimates uncertainty around the detector's original localisation without modifying the predicted bounding box.


\section{Theoretical Analysis}
\label{sec:appendix-Theoretical Analysis}
This appendix provides theoretical analysis to complement \methodname.

\subsection{Lemma 1}

\textbf{Lemma 1.} 
For two sets of uncertainty terms for retained detection $\hat{j}$ at curriculum stage $t$:

\begin{equation}
\left(
q^{\mathrm{gt}}_{\hat{j},t},
u^{\mathrm{corr}}_{\hat{j},t},
u^{\mathrm{stab}}_{\hat{j},t}
\right),
\qquad
\left(
q^{\mathrm{gt}\prime}_{\hat{j},t},
u^{\mathrm{corr}\prime}_{\hat{j},t},
u^{\mathrm{stab}\prime}_{\hat{j},t}
\right)
\in [0,1]^3
\end{equation}

the corresponding targets from Equation~\ref{eq:utgt_det} satisfy:

\begin{equation}
\left|u^{*}_{\hat{j},t} - u^{*\prime}_{\hat{j},t}\right|
\leq
\left|q^{\mathrm{gt}}_{\hat{j},t} - q^{\mathrm{gt}\prime}_{\hat{j},t}\right|
+
\left|u^{\mathrm{corr}}_{\hat{j},t} - u^{\mathrm{corr}\prime}_{\hat{j},t}\right|
+
\left|u^{\mathrm{stab}}_{\hat{j},t} - u^{\mathrm{stab}\prime}_{\hat{j},t}\right|
\end{equation}

Hence, the uncertainty target in~\ref{eq:utgt_det} is 1-Lipschitz with respect to the $\ell_1$ distance between its uncertainty terms.

\begin{proof}
Let $u^\star_{\hat j,t}$ and $u^{\star\prime}_{\hat j,t}$ be the uncertainty targets corresponding to the two sets of constituent uncertainty terms. From Equation~\ref{eq:utgt_det}:

\begin{equation}
 u^{*}_{\hat{j},t} = 1-
\left(1-q^{\mathrm{gt}}_{\hat{j},t}\right)
\left(1-u^{\mathrm{corr}}_{\hat{j},t}\right)
\left(1-u^{\mathrm{stab}}_{\hat{j},t}\right)
\end{equation}

and similarly, for the second set of uncertainty terms:

\begin{equation}
u^{*\prime}_{\hat{j},t} = 1-
\left(1-q^{\mathrm{gt}\prime}_{\hat{j},t}\right)
\left(1-u^{\mathrm{corr}\prime}_{\hat{j},t}\right)
\left(1-u^{\mathrm{stab}\prime}_{\hat{j},t}\right)
\end{equation}

Subtracting the two targets and rearranging the resulting product difference gives:

\begin{equation}
\begin{aligned}
u^{*}_{\hat{j},t} - u^{*\prime}_{\hat{j},t}
={}&
\left(q^{\mathrm{gt}}_{\hat{j},t} - q^{\mathrm{gt}\prime}_{\hat{j},t}\right)
\left(1-u^{\mathrm{corr}}_{\hat{j},t}\right)
\left(1-u^{\mathrm{stab}}_{\hat{j},t}\right)
\\
&+
\left(1-q^{\mathrm{gt}\prime}_{\hat{j},t}\right)
\left(u^{\mathrm{corr}}_{\hat{j},t} - u^{\mathrm{corr}\prime}_{\hat{j},t}\right)
\left(1-u^{\mathrm{stab}}_{\hat{j},t}\right)
\\
&+
\left(1-q^{\mathrm{gt}\prime}_{\hat{j},t}\right)
\left(1-u^{\mathrm{corr}\prime}_{\hat{j},t}\right)
\left(u^{\mathrm{stab}}_{\hat{j},t} - u^{\mathrm{stab}\prime}_{\hat{j},t}\right)
\end{aligned}
\end{equation}

Applying the triangle inequality therefore yields:

\begin{equation}
\begin{aligned}
\left|u^{*}_{\hat{j},t} - u^{*\prime}_{\hat{j},t}\right|
\leq{}&
\left|q^{\mathrm{gt}}_{\hat{j},t} - q^{\mathrm{gt}\prime}_{\hat{j},t}\right|
\left(1-u^{\mathrm{corr}}_{\hat{j},t}\right)
\left(1-u^{\mathrm{stab}}_{\hat{j},t}\right)
\\
&+
\left|u^{\mathrm{corr}}_{\hat{j},t} - u^{\mathrm{corr}\prime}_{\hat{j},t}\right|
\left(1-q^{\mathrm{gt}\prime}_{\hat{j},t}\right)
\left(1-u^{\mathrm{stab}}_{\hat{j},t}\right)
\\
&+
\left|u^{\mathrm{stab}}_{\hat{j},t} - u^{\mathrm{stab}\prime}_{\hat{j},t}\right|
\left(1-q^{\mathrm{gt}\prime}_{\hat{j},t}\right)
\left(1-u^{\mathrm{corr}\prime}_{\hat{j},t}\right)
\end{aligned}
\end{equation}

Since all constituent uncertainty terms lie in $[0,1]$, each complementary factor also lies in $[0,1]$. Consequently:

\begin{equation}
\left|u^{*}_{\hat{j},t} - u^{*\prime}_{\hat{j},t}\right|
\leq
\left|q^{\mathrm{gt}}_{\hat{j},t} - q^{\mathrm{gt}\prime}_{\hat{j},t}\right|
+
\left|u^{\mathrm{corr}}_{\hat{j},t} - u^{\mathrm{corr}\prime}_{\hat{j},t}\right|
+
\left|u^{\mathrm{stab}}_{\hat{j},t} - u^{\mathrm{stab}\prime}_{\hat{j},t}\right|
\end{equation}

Thus, perturbations in the uncertainty terms cannot be amplified beyond their $\ell_1$ distance by the aggregation in Equation~\ref{eq:utgt_det}. Hence, $u^{*}_{\hat{j},t}$ is 1-Lipschitz with respect to the $\ell_1$ distance between its constituent uncertainty terms.

This property is useful to \methodname\ because the individual uncertainty terms may vary due to changes in localisation, corruption exposure, or prediction stability; the bound guarantees that the aggregation itself does not amplify such variation when forming the final uncertainty target $u^{*}_{\hat{j},t}$.
\end{proof}


\section{Additional Experiments}
\label{sec:appendix-Additional Experiments}
This appendix provides additional experimental insights and analyses to complement the core results presented in Section~\ref{sec:Experimental Evaluation}. Specifically, we include extended analysis of metrics, adversarial attacks, thresholds, and ablations of hyperparameters introduced in \methodname.

\subsection{Full Results}
\label{sec:appendix-Full Results}

Table~\ref{tab:detector_results} reports the performance of the frozen YOLO11n detectors prior to uncertainty estimation. Detection performance varies across datasets, reflecting their different task characteristics. TT100K achieves the strongest overall detection performance, with mAP$_{50:95}$ of $0.52 \pm 0.00$, AP$_{75}$ of $0.62 \pm 0.00$, and median best IoU of $0.85 \pm 0.00$. COCO achieves a lower mAP$_{50:95}$ of $0.29 \pm 0.00$, but retains strong localisation among detected objects, with a median best IoU of $0.84 \pm 0.00$. VisDrone is more challenging, achieving mAP$_{50:95}$ of $0.29 \pm 0.00$ and median best IoU of $0.67 \pm 0.00$. CelebA exhibits near-perfect AP$_{50}$ and recall$_{50}$ of $0.99 \pm 0.00$, while its much lower AP$_{75}$ of $0.04 \pm 0.00$ indicates that precise localisation remains difficult despite reliable face detection. These results establish the fixed detection performance upon which all uncertainty approaches are evaluated.

\begin{table*}[tb]
    \centering
    \caption{Frozen detector performance across the four evaluation datasets. mAP$_{50:95}$ reports average precision across IoU thresholds from 0.50 to 0.95, while AP$_{50}$ and AP$_{75}$ report performance at fixed IoU thresholds. Precision$_{50}$ and recall$_{50}$ are calculated at an IoU threshold of 0.50. Median best IoU describe the localisation overlap between predicted detections and their best-matching ground-truth boxes.}
    \label{tab:detector_results}
    \small
    \setlength{\tabcolsep}{5pt}
    \begin{tabular}{lcccccc}
        \toprule
        Dataset
        & mAP$_{50:95}$ $\uparrow$
        & AP$_{50}$ $\uparrow$
        & AP$_{75}$ $\uparrow$
        & \shortstack{Precision$_{50}$ $\uparrow$}
        & \shortstack{Recall$_{50}$ $\uparrow$}
        & \shortstack{Median best IoU $\uparrow$} \\
        \midrule
        CelebA
        & $0.40 \pm 0.00$
        & $0.99 \pm 0.00$
        & $0.04 \pm 0.00$
        & $0.91 \pm 0.00$
        & $0.99 \pm 0.00$
        & $0.70 \pm 0.00$ \\
        COCO
        & $0.29 \pm 0.00$
        & $0.41 \pm 0.00$
        & $0.31 \pm 0.00$
        & $0.74 \pm 0.00$
        & $0.44 \pm 0.00$
        & $0.84 \pm 0.00$ \\
        TT100K
        & $0.52 \pm 0.00$
        & $0.70 \pm 0.00$
        & $0.62 \pm 0.00$
        & $0.60 \pm 0.00$
        & $0.84 \pm 0.00$
        & $0.85 \pm 0.00$ \\
        VisDrone
        & $0.29 \pm 0.00$
        & $0.52 \pm 0.00$
        & $0.27 \pm 0.00$
        & $0.51 \pm 0.00$
        & $0.60 \pm 0.00$
        & $0.67 \pm 0.00$ \\
        \bottomrule
    \end{tabular}
\end{table*}

Table~\ref{tab:calibration_results} provides additional analysis of detection-level uncertainty calibration and its alignment with localisation quality. These metrics complement the selective-prediction results in the main paper by assessing whether uncertainty separates correct from incorrect detections, whether the resulting detection-success probabilities are calibrated, and whether uncertainty increases as localisation quality deteriorates.

EMM exhibits strongly dataset-dependent behaviour. On CelebA and COCO, it provides little alignment between uncertainty and localisation quality, with Pearson correlations of $-0.003 \pm 0.075$ and $-0.007 \pm 0.020$, respectively, alongside comparatively high Brier scores of $0.797 \pm 0.035$ and $0.669 \pm 0.004$. Its strongest results occur on TT100K, where it achieves the lowest FPR at 95\% TPR of $0.907 \pm 0.023$ and the strongest Pearson correlation of $-0.292 \pm 0.019$. EMM is also competitive on VisDrone, obtaining the lowest SSCE of $0.188 \pm 0.007$ and a correlation of $-0.262 \pm 0.001$. These results suggest that learning uncertainty from frozen detector representations alone can be effective in some domains, but does not consistently produce uncertainty that tracks localisation quality.

\begin{table}[tb]
    \centering
    \caption{Detection uncertainty calibration and alignment across four datasets. FPR at 95\% TPR measures the false-positive rate required to retain 95\% of correct detections. Brier score and success-score calibration error (SSC) measure the calibration of predicted detection success probabilities. Pearson uncertainty-IoU correlation measures linear alignment between uncertainty and localisation quality, where a more negative correlation is better. Results are mean $\pm$ standard deviation over up to five runs. Best results are bold and \colorbox{lightblue}{highlighted}.}
    \label{tab:calibration_results}
    \small
    \setlength{\tabcolsep}{5pt}
    \begin{tabular}{lcccc}
        \toprule
        Method
        & \shortstack{FPR at 95\% TPR $\downarrow$}
        & \shortstack{Brier score $\downarrow$}
        & \shortstack{SSC $\downarrow$}
        & \shortstack{Pearson uncertainty-IoU\\correlation $\downarrow$} \\
        \midrule
        \multicolumn{5}{c}{\textbf{CelebA}} \\
        \midrule
        \methodname
        & \cellcolor{lightblue}$\mathbf{0.281 \pm 0.062}$
        & $0.284 \pm 0.177$
        & $0.421 \pm 0.235$
        & \cellcolor{lightblue}$\mathbf{-0.773 \pm 0.031}$ \\
        EMM
        & $0.907 \pm 0.026$
        & $0.797 \pm 0.035$
        & $0.844 \pm 0.021$
        & $-0.003 \pm 0.075$ \\
        ModelNet
        & $0.647 \pm 0.185$
        & $0.783 \pm 0.033$
        & $0.846 \pm 0.023$
        & $0.324 \pm 0.300$ \\
        MetaDetect
        & $0.483 \pm 0.000$
        & \cellcolor{lightblue}$\mathbf{0.084 \pm 0.000}$
        & \cellcolor{lightblue}$\mathbf{0.028 \pm 0.000}$
        & $-0.524 \pm 0.000$ \\
        \midrule
        \addlinespace
        \multicolumn{5}{c}{\textbf{COCO}} \\
        \midrule
        \methodname
        & $0.934 \pm 0.009$
        & $0.456 \pm 0.133$
        & $0.487 \pm 0.145$
        & \cellcolor{lightblue}$\mathbf{-0.146 \pm 0.080}$ \\
        EMM
        & $0.951 \pm 0.000$
        & $0.669 \pm 0.004$
        & $0.692 \pm 0.003$
        & $-0.007 \pm 0.020$ \\
        ModelNet
        & $0.945 \pm 0.020$
        & $0.672 \pm 0.000$
        & $0.693 \pm 0.000$
        & $0.032 \pm 0.061$ \\
        MetaDetect
        & \cellcolor{lightblue}$\mathbf{0.929 \pm 0.000}$
        & \cellcolor{lightblue}$\mathbf{0.192 \pm 0.000}$
        & \cellcolor{lightblue}$\mathbf{0.029 \pm 0.000}$
        & $-0.084 \pm 0.000$ \\
        \midrule
        \addlinespace
        \multicolumn{5}{c}{\textbf{TT100K}} \\
        \midrule
        \methodname
        & $0.943 \pm 0.008$
        & $0.409 \pm 0.090$
        & $0.370 \pm 0.134$
        & $-0.209 \pm 0.018$ \\
        EMM
        & \cellcolor{lightblue}$\mathbf{0.907 \pm 0.023}$
        & $0.299 \pm 0.019$
        & $0.206 \pm 0.022$
        & \cellcolor{lightblue}$\mathbf{-0.292 \pm 0.019}$ \\
        ModelNet
        & $0.927 \pm 0.012$
        & $0.545 \pm 0.000$
        & $0.553 \pm 0.000$
        & $0.037 \pm 0.044$ \\
        MetaDetect
        & $0.940 \pm 0.000$
        & \cellcolor{lightblue}$\mathbf{0.266 \pm 0.000}$
        & \cellcolor{lightblue}$\mathbf{0.164 \pm 0.000}$
        & $-0.166 \pm 0.000$ \\
        \midrule
        \addlinespace
        \multicolumn{5}{c}{\textbf{VisDrone}} \\
        \midrule
        \methodname
        & \cellcolor{lightblue}$\mathbf{0.905 \pm 0.008}$
        & \cellcolor{lightblue}$\mathbf{0.281 \pm 0.028}$
        & $0.192 \pm 0.067$
        & \cellcolor{lightblue}$\mathbf{-0.494 \pm 0.043}$ \\
        EMM
        & $0.937 \pm 0.004$
        & $0.293 \pm 0.003$
        & \cellcolor{lightblue}$\mathbf{0.188 \pm 0.007}$
        & $-0.262 \pm 0.001$ \\
        ModelNet
        & $0.912 \pm 0.002$
        & $0.399 \pm 0.000$
        & $0.393 \pm 0.000$
        & $-0.232 \pm 0.011$ \\
        MetaDetect
        & $0.946 \pm 0.000$
        & $0.443 \pm 0.000$
        & $0.441 \pm 0.000$
        & $-0.091 \pm 0.000$ \\
        \bottomrule
    \end{tabular}
\end{table}

ModelNet is generally weaker across both calibration and localisation alignment. It obtains Brier scores of $0.783 \pm 0.033$, $0.672 \pm 0.000$, and $0.545 \pm 0.000$ on CelebA, COCO, and TT100K, respectively, with corresponding SSC values of $0.846 \pm 0.023$, $0.693 \pm 0.000$, and $0.553 \pm 0.000$. More importantly, its Pearson uncertainty-IoU correlations are positive on CelebA, COCO, and TT100K at $0.324 \pm 0.300$, $0.032 \pm 0.061$, and $0.037 \pm 0.044$, indicating that higher assigned uncertainty does not reliably correspond to poorer localisation on these datasets. Performance improves on VisDrone, where the correlation reaches $-0.232 \pm 0.011$, but calibration remains weaker than the strongest comparative approaches.

MetaDetect provides the strongest probability calibration on several datasets. On CelebA it achieves a Brier score of $0.084 \pm 0.000$ and SSC of $0.028 \pm 0.000$, while on COCO these remain low at $0.192 \pm 0.000$ and $0.029 \pm 0.000$. It also obtains the lowest Brier score and SSC on TT100K, at $0.266 \pm 0.000$ and $0.164 \pm 0.000$. However, this calibration does not consistently translate into equally strong localisation alignment. Pearson correlations range from $-0.524 \pm 0.000$ on CelebA to only $-0.084 \pm 0.000$ on COCO and $-0.091 \pm 0.000$ on VisDrone. Its performance also degrades substantially on VisDrone, where the Brier score increases to $0.443 \pm 0.000$ and SSC to $0.441 \pm 0.000$. MetaDetect therefore performs particularly well at calibrating detection success, but is less consistent at ordering detections according to localisation quality.

\methodname\ provides the most consistent localisation-aware uncertainty across the evaluated datasets. On CelebA it achieves both the lowest FPR at 95\% TPR of $0.281 \pm 0.062$ and the strongest Pearson uncertainty-IoU correlation of $-0.773 \pm 0.031$, substantially exceeding the alignment obtained by the comparative approaches. On COCO, \methodname\ again achieves the strongest correlation of $-0.146 \pm 0.080$, while its FPR of $0.934 \pm 0.009$ remains close to the best result of $0.929 \pm 0.000$. Although \methodname\ is not the strongest method on TT100K, it maintains a negative correlation of $-0.209 \pm 0.018$, indicating meaningful alignment between uncertainty and localisation error. On VisDrone, it achieves the lowest FPR of $0.905 \pm 0.008$, the lowest Brier score of $0.281 \pm 0.028$, and the strongest Pearson correlation of $-0.494 \pm 0.043$, while its SSC of $0.192 \pm 0.067$ remains comparable to the best result. 

Overall, these results show that \methodname\ is not uniformly optimal in probability calibration, particularly where MetaDetect is strongest, but more consistently assigns greater uncertainty to poorly localised detections across datasets. This supports the intended role of \methodname\ as a localisation-aware uncertainty estimator rather than solely a calibrated detector-success predictor.

\begin{figure*}[tb]
    \centering
    \includegraphics[width=\linewidth]{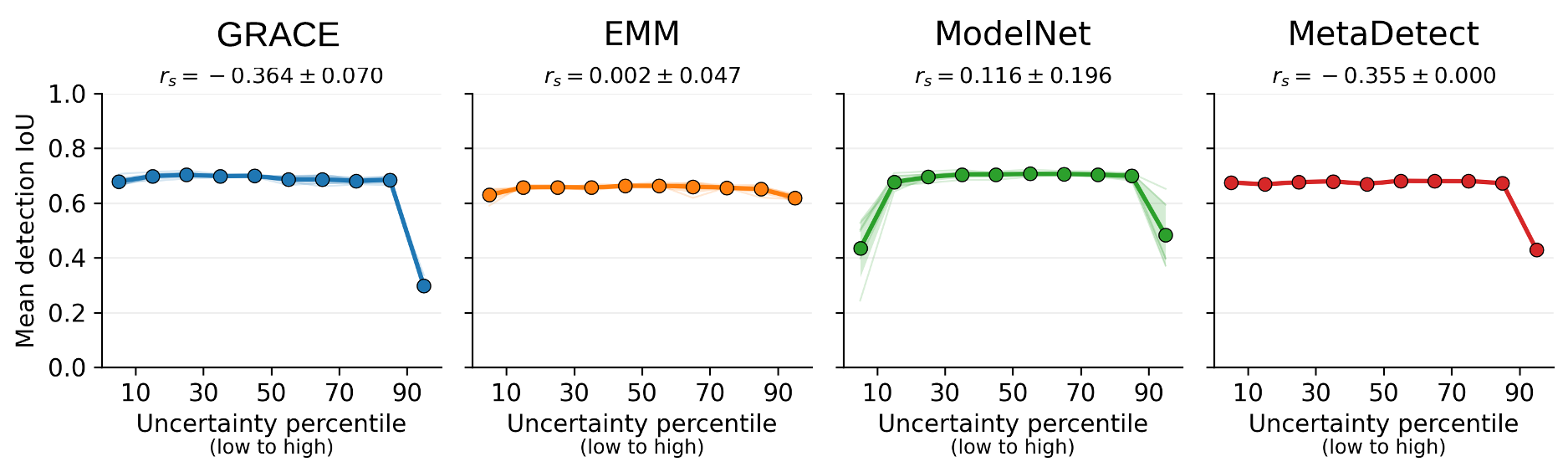}
    \caption{Mean detection IoU across uncertainty percentile bins for CelebA. Detections are ordered from lowest to highest assigned uncertainty within each run and divided into equal-count bins. $r_s$ reports the mean Spearman uncertainty–IoU correlation, where a stronger downward trend and more negative $r_s$ indicate better uncertainty–localisation alignment.}
    \label{fig:celeba-uncert-iuo-corr}
\end{figure*}

\begin{figure*}[tb]
    \centering
    \includegraphics[width=\linewidth]{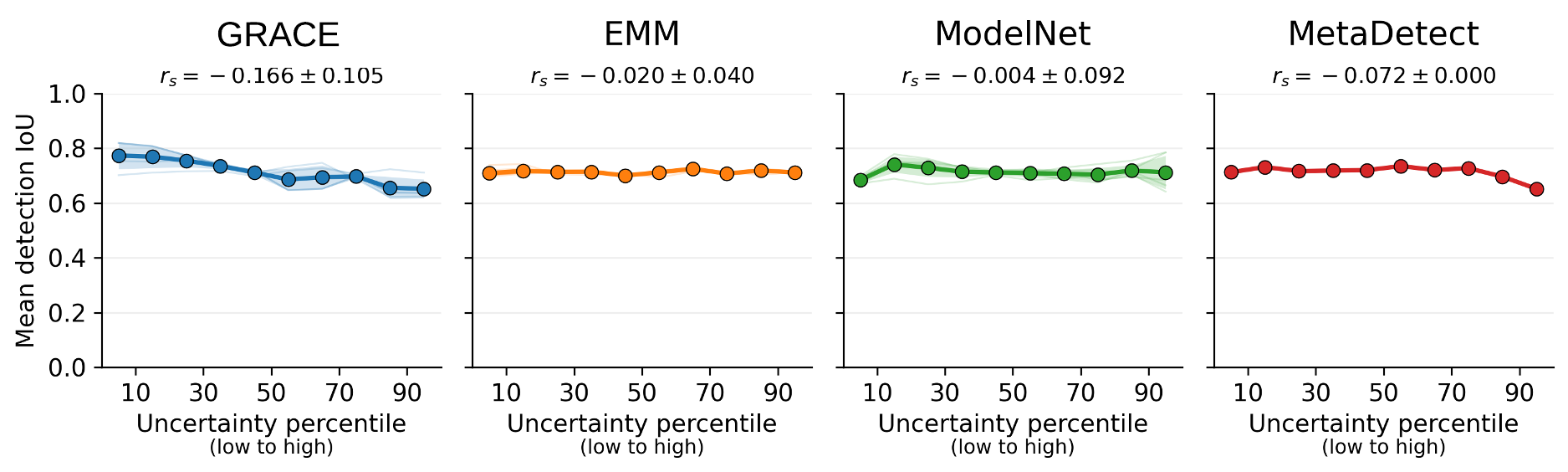}
    \caption{Mean detection IoU across uncertainty percentile bins for Coco. Detections are ordered from lowest to highest assigned uncertainty within each run and divided into equal-count bins. $r_s$ reports the mean Spearman uncertainty–IoU correlation, where a stronger downward trend and more negative $r_s$ indicate better uncertainty–localisation alignment.}
    \label{fig:coco-uncert-iuo-corr}
\end{figure*}

\begin{figure*}[tb]
    \centering
    \includegraphics[width=\linewidth]{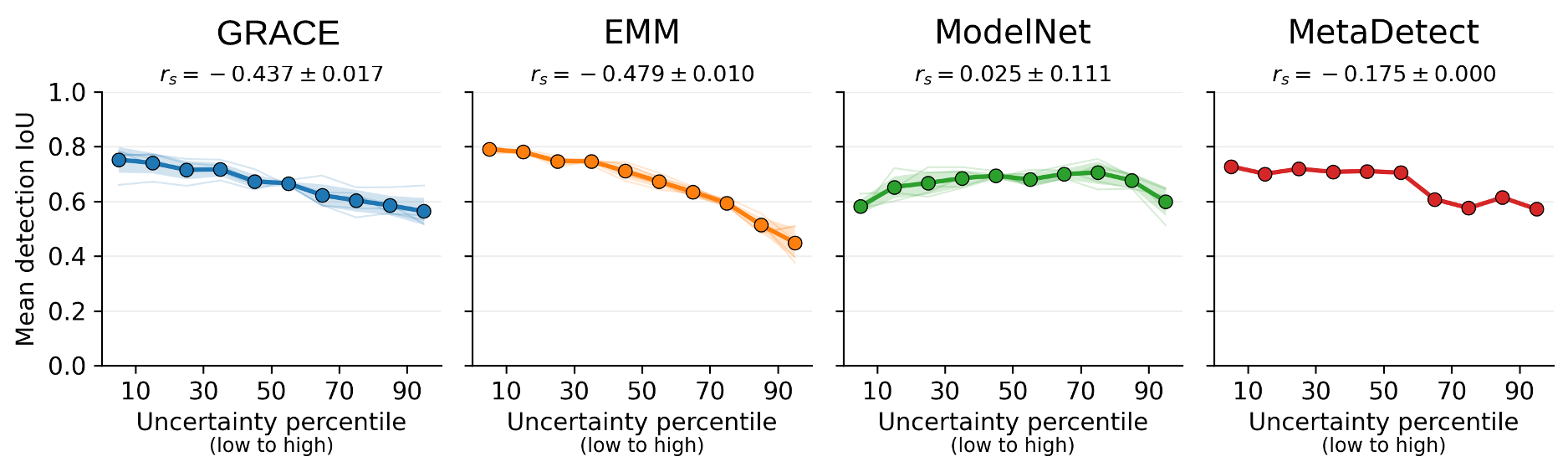}
    \caption{Mean detection IoU across uncertainty percentile bins for TT100K. Detections are ordered from lowest to highest assigned uncertainty within each run and divided into equal-count bins. $r_s$ reports the mean Spearman uncertainty–IoU correlation, where a stronger downward trend and more negative $r_s$ indicate better uncertainty–localisation alignment.}
    \label{fig:tt100k-uncert-iuo-corr}
\end{figure*}

\begin{figure*}[tb]
    \centering
    \includegraphics[width=\linewidth]{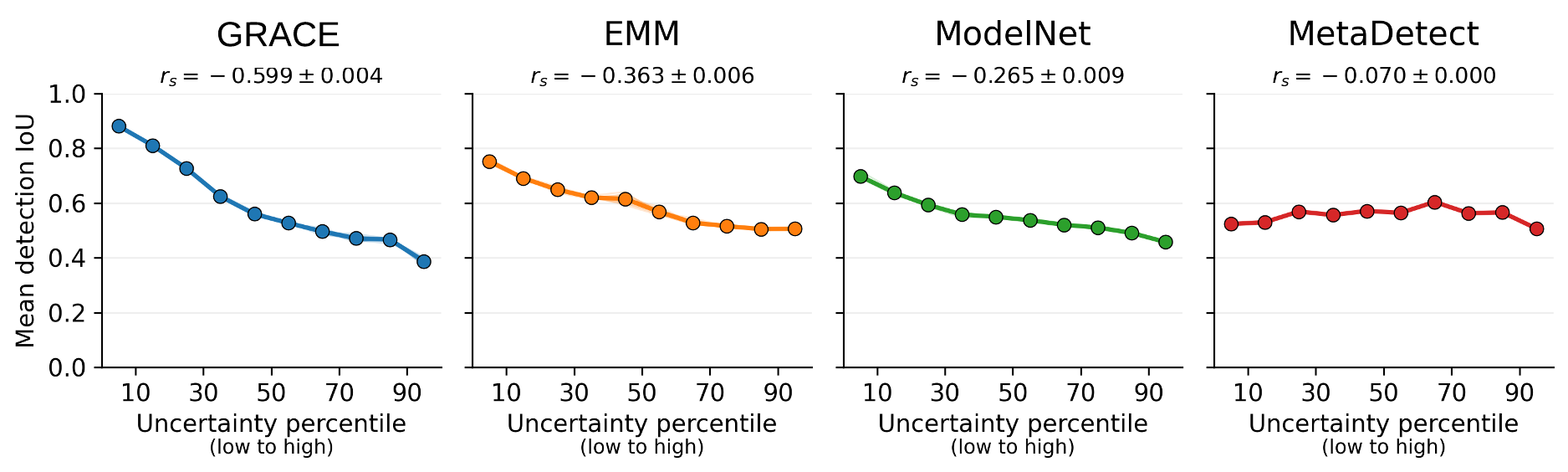}
    \caption{Mean detection IoU across uncertainty percentile bins for VisDrone. Detections are ordered from lowest to highest assigned uncertainty within each run and divided into equal-count bins. $r_s$ reports the mean Spearman uncertainty–IoU correlation, where a stronger downward trend and more negative $r_s$ indicate better uncertainty–localisation alignment.}
    \label{fig:visdrone-uncert-iuo-corr}
\end{figure*}

Figures~\ref{fig:celeba-uncert-iuo-corr}-\ref{fig:visdrone-uncert-iuo-corr} provide a more direct view of localisation-aware uncertainty by plotting mean detection IoU across uncertainty percentile bins. In general, a desirable uncertainty estimator should produce a clear monotonic downward trend, such that detections assigned higher uncertainty also exhibit lower localisation quality.

On COCO, both \methodname\ and EMM produce clear downward trends, with EMM slightly stronger at $-0.479 \pm 0.010$ compared with $-0.437 \pm 0.017$ for \methodname, while ModelNet is effectively uninformative at $0.025 \pm 0.111$ and MetaDetect is weaker at $-0.175 \pm 0.000$.

On TT100K, alignment is weaker overall, but \methodname\ produces the strongest negative trend at $-0.166 \pm 0.105$, compared with $-0.020 \pm 0.040$ for EMM, $-0.004 \pm 0.092$ for ModelNet, and $-0.072 \pm 0.000$ for MetaDetect.

On VisDrone, \methodname\ again shows a clear downward trend with $-0.364 \pm 0.070$, narrowly stronger than MetaDetect at $-0.355 \pm 0.000$, while EMM and ModelNet are effectively flat or misaligned at $0.002 \pm 0.047$ and $0.116 \pm 0.196$, respectively. Overall, these plots show that \methodname\ more consistently ranks poorly localised detections as uncertain.

Table~\ref{tab:efficiency_results} compares the inference overhead introduced by each uncertainty approach relative to the frozen detector. ModelNet consistently incurs substantial additional cost (mostly likely due to Monte-Carlo inference), adding $42.93 \pm 3.44$ms on CelebA, $49.37 \pm 1.09$ms on COCO, $45.82 \pm 2.71$ms on TT100K, and $80.18 \pm 0.94$ms on VisDrone. This produces the lowest or near-lowest end-to-end throughput across the evaluated datasets. MetaDetect similarly introduces considerable overhead, ranging from $20.39 \pm 0.95$ms on COCO to $84.58 \pm 1.66$ms on VisDrone, substantially reducing throughput relative to the lighter evidential approaches.

\begin{table}[tb]
    \centering
    \caption{Computational efficiency across four datasets. Added time reports the latency introduced by each uncertainty method relative to the frozen detector, both in milliseconds and as a percentage. Time per box measures uncertainty-processing latency per detected object, while end-to-end FPS includes both detector and uncertainty-method inference. Best results are bold and \colorbox{lightblue}{highlighted}.}
    \label{tab:efficiency_results}
    \small
    \setlength{\tabcolsep}{5pt}
    \begin{tabular}{lcccc}
        \toprule
        Method
        & \shortstack{Added time\\(ms) $\downarrow$}
        & \shortstack{Added time\\(\%) $\downarrow$}
        & \shortstack{Time per box\\(ms) $\downarrow$}
        & \shortstack{End-to-end\\FPS $\uparrow$} \\
        \midrule
        \multicolumn{5}{c}{\textbf{CelebA}} \\
        \midrule
        \methodname
        & \cellcolor{lightblue}$\mathbf{2.66 \pm 0.48}$
        & \cellcolor{lightblue}$\mathbf{11.66 \pm 0.67}$
        & \cellcolor{lightblue}$\mathbf{2.616 \pm 0.479}$
        & \cellcolor{lightblue}$\mathbf{41.84 \pm 0.65}$ \\
        EMM
        & $11.76 \pm 0.75$
        & $25.19 \pm 1.26$
        & $11.555 \pm 0.749$
        & $25.69 \pm 1.46$ \\
        ModelNet
        & $42.93 \pm 3.44$
        & $118.02 \pm 0.64$
        & $42.071 \pm 3.381$
        & $14.81 \pm 1.01$ \\
        MetaDetect
        & $27.00 \pm 15.77$
        & $78.67 \pm 20.44$
        & $26.476 \pm 15.493$
        & $23.50 \pm 5.44$ \\
        \midrule
        \addlinespace
        \multicolumn{5}{c}{\textbf{COCO}} \\
        \midrule
        \methodname
        & $8.01 \pm 4.76$
        & $19.12 \pm 7.27$
        & $3.105 \pm 1.828$
        & $29.79 \pm 7.75$ \\
        EMM
        & \cellcolor{lightblue}$\mathbf{2.79 \pm 0.61}$
        & \cellcolor{lightblue}$\mathbf{11.43 \pm 0.93}$
        & \cellcolor{lightblue}$\mathbf{1.061 \pm 0.242}$
        & \cellcolor{lightblue}$\mathbf{40.61 \pm 1.11}$ \\
        ModelNet
        & $49.37 \pm 1.09$
        & $215.49 \pm 2.82$
        & $15.923 \pm 0.396$
        & $15.94 \pm 0.38$ \\
        MetaDetect
        & $20.39 \pm 0.95$
        & $83.89 \pm 1.88$
        & $6.780 \pm 0.373$
        & $24.36 \pm 0.43$ \\
        \midrule
        \addlinespace
        \multicolumn{5}{c}{\textbf{TT100K}} \\
        \midrule
        \methodname
        & $4.30 \pm 0.21$
        & $4.27 \pm 0.16$
        & $1.781 \pm 0.086$
        & $9.60 \pm 0.15$ \\
        EMM
        & \cellcolor{lightblue}$\mathbf{4.20 \pm 0.22}$
        & \cellcolor{lightblue}$\mathbf{4.23 \pm 0.21}$
        & \cellcolor{lightblue}$\mathbf{1.757 \pm 0.093}$
        & \cellcolor{lightblue}$\mathbf{9.65 \pm 0.14}$ \\
        ModelNet
        & $45.82 \pm 2.71$
        & $45.43 \pm 1.83$
        & $18.940 \pm 1.221$
        & $6.93 \pm 0.22$ \\
        MetaDetect
        & $80.75 \pm 2.00$
        & $80.27 \pm 0.79$
        & $32.288 \pm 0.867$
        & $5.72 \pm 0.09$ \\
        \midrule
        \addlinespace
        \multicolumn{5}{c}{\textbf{VisDrone}} \\
        \midrule
        \methodname
        & $6.24 \pm 1.14$
        & \cellcolor{lightblue}$\mathbf{8.61 \pm 1.04}$
        & \cellcolor{lightblue}$\mathbf{0.145 \pm 0.024}$
        & $13.04 \pm 0.33$ \\
        EMM
        & \cellcolor{lightblue}$\mathbf{6.16 \pm 0.76}$
        & $8.61 \pm 0.72$
        & $0.151 \pm 0.026$
        & \cellcolor{lightblue}$\mathbf{13.11 \pm 0.31}$ \\
        ModelNet
        & $80.18 \pm 0.94$
        & $272.71 \pm 3.31$
        & $1.680 \pm 0.023$
        & $9.50 \pm 0.12$ \\
        MetaDetect
        & $84.58 \pm 1.66$
        & $280.15 \pm 2.90$
        & $1.345 \pm 0.029$
        & $10.41 \pm 0.24$ \\
        \bottomrule
    \end{tabular}
\end{table}

EMM is considerably more efficient, achieving the lowest added latency on COCO at $2.79 \pm 0.61$ms and narrowly outperforming \methodname\ on TT100K and VisDrone with $4.20 \pm 0.22$ms and $6.16 \pm 0.76$ms, respectively. This translates to the highest end-to-end throughput on these datasets, although the differences from \methodname\ are small on TT100K and VisDrone.

\methodname\ remains similarly lightweight while providing the strongest efficiency on CelebA, adding only $2.66 \pm 0.48$ms and retaining $41.84 \pm 0.65$ FPS. On TT100K, its $4.30 \pm 0.21$ms overhead is effectively comparable to EMM, while on VisDrone it achieves the lowest per-box processing time at approximately $0.15 \pm 0.02$ms. COCO represents the largest \methodname\ overhead at $8.01 \pm 4.76$ms, but this remains substantially below ModelNet and MetaDetect. Overall, \methodname\ introduces a small inference overhead comparable to the lightweight EMM baseline, while remaining considerably more efficient than ModelNet and MetaDetect across all four datasets.

\begin{figure*}[tb]
    \centering
    \includegraphics[width=\linewidth]{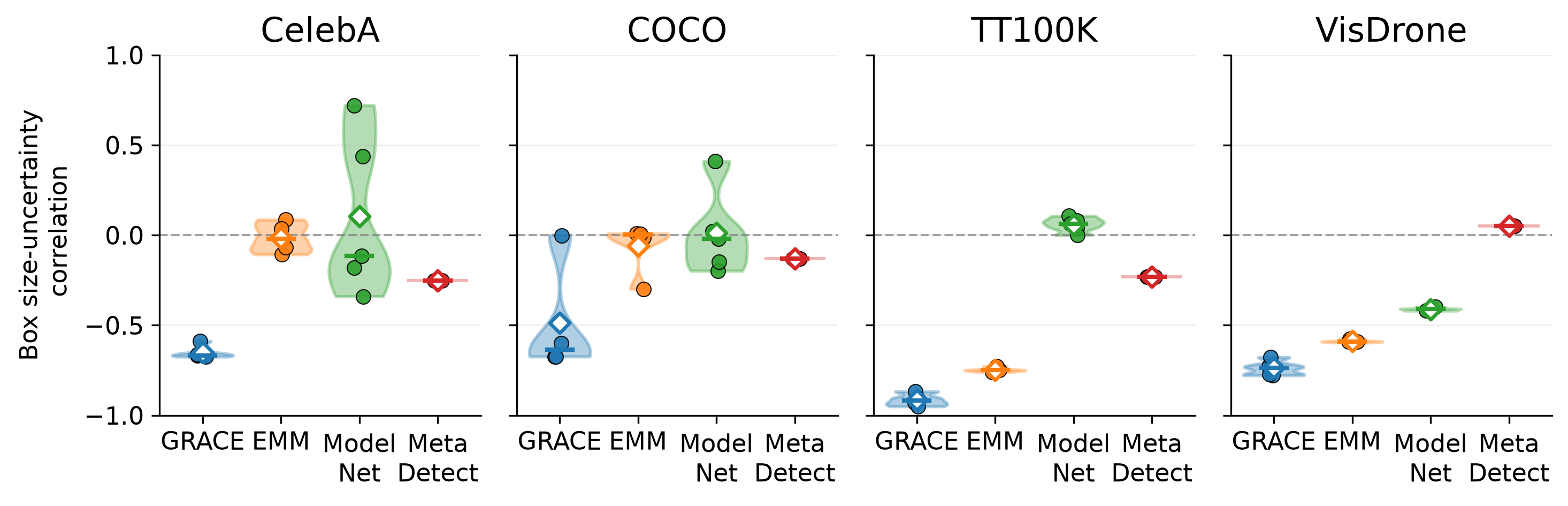}
    \caption{Pearson correlations between detection box size and assigned uncertainty across four datasets. Violins show the distribution across five runs, dots individual runs, diamonds means, solid horizontal lines medians, and the dashed line zero correlation. Values closer to zero indicate less systematic dependence on box size, while more negative correlations indicate greater uncertainty for smaller detections.}
    \label{fig:box_size_uncert_corr}
\end{figure*}

Figure~\ref{fig:box_size_uncert_corr} examines whether assigned uncertainty is systematically associated with predicted box size. \methodname\ exhibits consistently negative correlations across all four datasets, indicating that smaller detections tend to receive greater uncertainty. This effect is strongest on TT100K and VisDrone, where small and distant objects are common, and remains evident on CelebA and COCO. EMM shows comparatively weak size dependence on CelebA and COCO but stronger negative correlations on TT100K and VisDrone, while ModelNet varies substantially across datasets and runs. MetaDetect generally exhibits weaker correlations, including a near-zero relationship on VisDrone. These results show that \methodname's localisation uncertainty is sensitive to object scale; while this reflects the greater localisation difficulty of small objects, it also indicates that box size contributes systematically to the uncertainty estimates and should therefore be considered when interpreting detection-level uncertainty.

\subsection{Adversarial Attack Results}
\label{sec:appendix-Adversarial Attack Results}

We further evaluate uncertainty robustness under adversarial perturbation, where inputs are deliberately modified to degrade detector performance while remaining visually similar to the original image. We consider a mixture of white-box gradient-based and black-box query-based attacks, including PGD, TOG Vanishing, TOG Fabrication, and Square Attack, across increasing attack strengths. This allows us to assess whether each uncertainty method continues to identify unreliable detections when the detector is subjected to targeted adversarial manipulation.

\begin{figure*}[tb]
    \centering
    \includegraphics[width=\linewidth]{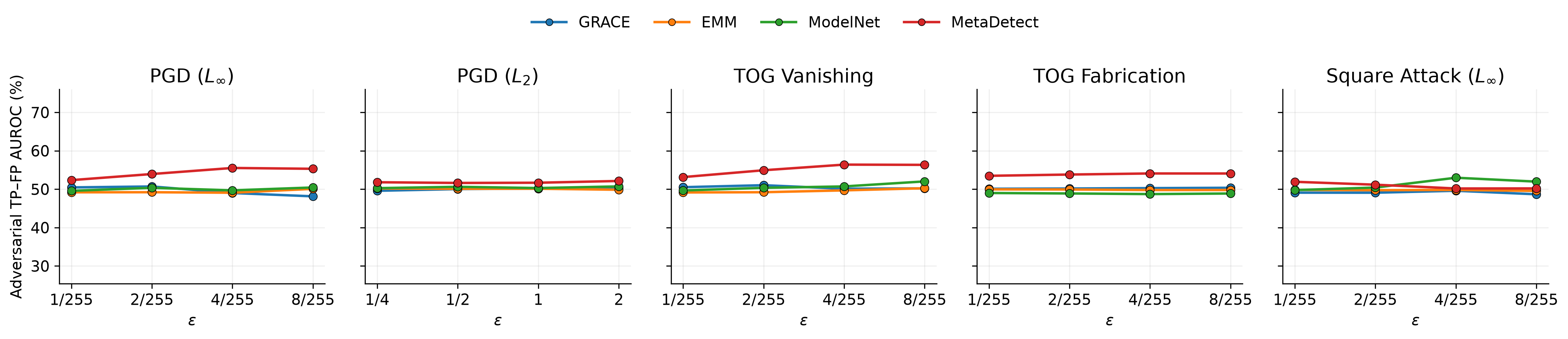}
    \caption{COCO TP–FP AUROC under five adversarial attacks across increasing attack strength $\epsilon$. Lines show mean performance across five runs for \methodname, EMM, ModelNet, and MetaDetect under PGD ($L_\infty$), PGD ($L_2$), TOG Vanishing, TOG Fabrication, and Square Attack ($L_\infty$); higher AUROC indicates better discrimination between true- and false-positive detections under adversarial perturbation.}
    \label{fig:coco-adv-attacks}
\end{figure*}

\begin{figure*}[tb]
    \centering
    \includegraphics[width=\linewidth]{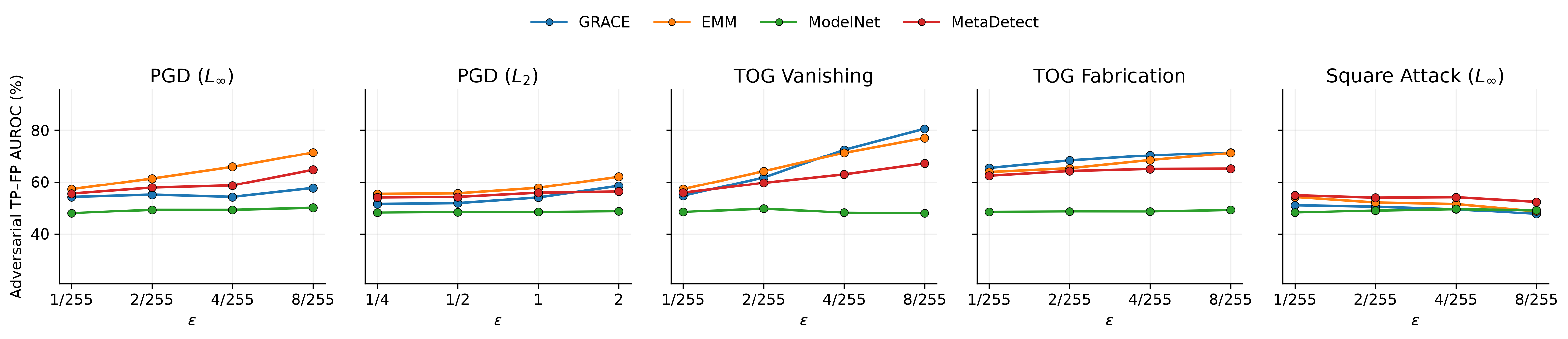}
    \caption{TT100K TP–FP AUROC under five adversarial attacks across increasing attack strength $\epsilon$. Lines show mean performance across five runs for \methodname, EMM, ModelNet, and MetaDetect under PGD ($L_\infty$), PGD ($L_2$), TOG Vanishing, TOG Fabrication, and Square Attack ($L_\infty$); higher AUROC indicates better discrimination between true- and false-positive detections under adversarial perturbation.}
    \label{fig:tt100k-adv-attacks}
\end{figure*}

\begin{figure*}[tb]
    \centering
    \includegraphics[width=\linewidth]{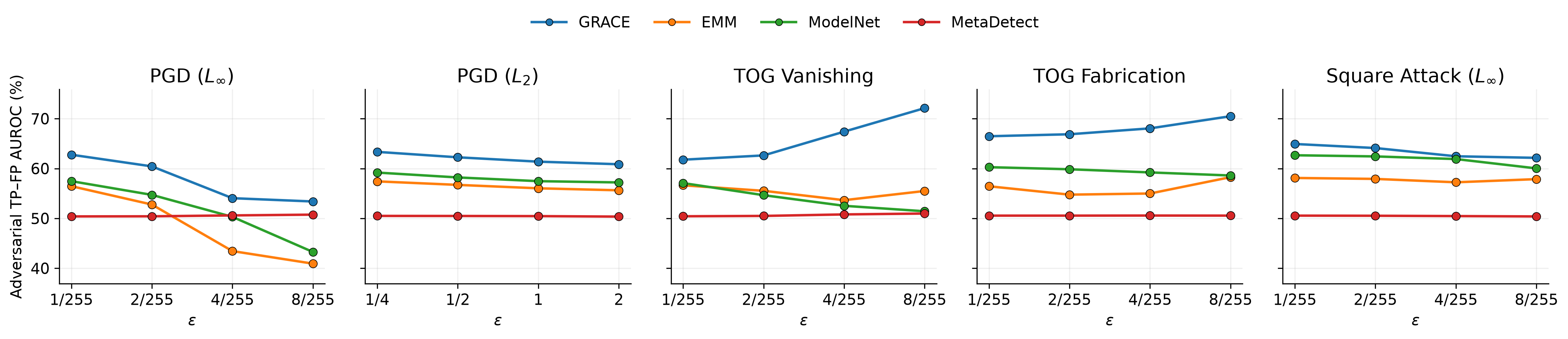}
    \caption{VisDrone TP–FP AUROC under five adversarial attacks across increasing attack strength $\epsilon$. Lines show mean performance across five runs for \methodname, EMM, ModelNet, and MetaDetect under PGD ($L_\infty$), PGD ($L_2$), TOG Vanishing, TOG Fabrication, and Square Attack ($L_\infty$); higher AUROC indicates better discrimination between true- and false-positive detections under adversarial perturbation.}
    \label{fig:visdrone-adv-attacks}
\end{figure*}

Figures~\ref{fig:coco-adv-attacks}-~\ref{fig:visdrone-adv-attacks} extend the adversarial evaluation to COCO, TT100K, and VisDrone across PGD $L_\infty$, PGD $L_2$, TOG Vanishing, TOG Fabrication, and Square Attack $L_\infty$. On COCO (Figure~\ref{fig:coco-adv-attacks}), performance is comparatively compressed across methods, with most TP-FP AUROC values remaining close to $0.50$. MetaDetect is generally strongest, particularly under PGD $L_\infty$ and TOG attacks, while \methodname\ remains comparable to EMM and ModelNet across most attack strengths.

On TT100K (Figure~\ref{fig:tt100k-adv-attacks}), \methodname\ and EMM separate more clearly from the comparative approaches. Both generally improve as PGD $L_\infty$ and TOG attack strength increase, with \methodname\ reaching approximately $0.75$ under the strongest TOG Vanishing attack. EMM is particularly competitive under PGD $L_\infty$, while ModelNet remains consistently weaker across the evaluated attacks.

The clearest behaviour is observed on VisDrone (Figure~\ref{fig:visdrone-adv-attacks}), where \methodname\ achieves the strongest TP-FP discrimination across nearly all attacks and strengths. Under TOG Vanishing and TOG Fabrication, its AUROC increases with attack strength, reaching approximately $0.75$ and $0.70$, respectively. Although performance decreases under stronger PGD $L_\infty$ perturbations, \methodname\ remains above the comparative methods for most strengths. Overall, the results demonstrate that adversarial robustness varies substantially by dataset and attack type, but \methodname\ remains particularly effective on the more challenging TT100K and VisDrone settings.

\begin{table*}[tb]
    \centering
    \caption{Uncertainty performance under adversarial attacks across four datasets. Results aggregate performance across PGD ($L_\infty$), PGD ($L_2$), TOG Vanishing, TOG Fabrication, and Square Attack ($L_\infty$) and their tested attack strengths within each run, and report mean $\pm$ standard deviation across runs. TP-FP AUROC and FP AUPRC evaluate identification of unreliable detections, while risk-coverage AUC and IoU-coverage AUC evaluate selective prediction and retained localisation quality under adversarial perturbation. Best results are bold and \colorbox{lightblue}{highlighted}.}
    \label{tab:adversarial_results}
    \small
    \setlength{\tabcolsep}{4pt}
    \resizebox{\textwidth}{!}{%
    \begin{tabular}{lcccc}
        \toprule
        Method & \shortstack{Adversarial TP-FP\\AUROC $\uparrow$} & \shortstack{Adversarial FP\\AUPRC $\uparrow$} & \shortstack{Adversarial risk-coverage\\AUC $\downarrow$} & \shortstack{Adversarial IoU-coverage\\AUC $\uparrow$} \\
        \midrule
        \multicolumn{5}{c}{\textbf{CelebA}} \\
        \midrule
        \methodname
        & \cellcolor{lightblue}$\mathbf{0.825 \pm 0.020}$
        & \cellcolor{lightblue}$\mathbf{0.850 \pm 0.025}$
        & \cellcolor{lightblue}$\mathbf{0.555 \pm 0.009}$
        & \cellcolor{lightblue}$\mathbf{0.400 \pm 0.005}$ \\
        EMM
        & $0.519 \pm 0.028$
        & $0.661 \pm 0.009$
        & $0.641 \pm 0.010$
        & $0.317 \pm 0.009$ \\
        ModelNet
        & $0.501 \pm 0.241$
        & $0.715 \pm 0.090$
        & $0.692 \pm 0.050$
        & $0.306 \pm 0.067$ \\
        MetaDetect
        & $0.676 \pm 0.000$
        & $0.757 \pm 0.000$
        & $0.589 \pm 0.000$
        & $0.340 \pm 0.000$ \\
        \midrule
        \addlinespace
        \multicolumn{5}{c}{\textbf{COCO}} \\
        \midrule
        \methodname
        & $0.499 \pm 0.064$
        & $0.548 \pm 0.031$
        & $0.547 \pm 0.032$
        & $0.482 \pm 0.047$ \\
        EMM
        & $0.497 \pm 0.007$
        & $0.544 \pm 0.003$
        & $0.549 \pm 0.005$
        & $0.485 \pm 0.002$ \\
        ModelNet
        & $0.503 \pm 0.024$
        & $0.552 \pm 0.018$
        & $0.555 \pm 0.008$
        & $0.472 \pm 0.013$ \\
        MetaDetect
        & \cellcolor{lightblue}$\mathbf{0.532 \pm 0.000}$
        & \cellcolor{lightblue}$\mathbf{0.576 \pm 0.001}$
        & \cellcolor{lightblue}$\mathbf{0.533 \pm 0.001}$
        & \cellcolor{lightblue}$\mathbf{0.499 \pm 0.000}$ \\
        \midrule
        \addlinespace
        \multicolumn{5}{c}{\textbf{TT100K}} \\
        \midrule
        \methodname
        & $0.591 \pm 0.052$
        & $0.593 \pm 0.034$
        & $0.491 \pm 0.031$
        & $0.590 \pm 0.044$ \\
        EMM
        & \cellcolor{lightblue}$\mathbf{0.616 \pm 0.006}$
        & \cellcolor{lightblue}$\mathbf{0.635 \pm 0.017}$
        & \cellcolor{lightblue}$\mathbf{0.484 \pm 0.008}$
        & \cellcolor{lightblue}$\mathbf{0.605 \pm 0.007}$ \\
        ModelNet
        & $0.488 \pm 0.043$
        & $0.559 \pm 0.017$
        & $0.564 \pm 0.013$
        & $0.490 \pm 0.021$ \\
        MetaDetect
        & $0.588 \pm 0.001$
        & $0.595 \pm 0.001$
        & $0.495 \pm 0.001$
        & $0.568 \pm 0.000$ \\
        \midrule
        \addlinespace
        \multicolumn{5}{c}{\textbf{VisDrone}} \\
        \midrule
        \methodname
        & \cellcolor{lightblue}$\mathbf{0.634 \pm 0.003}$
        & \cellcolor{lightblue}$\mathbf{0.717 \pm 0.003}$
        & \cellcolor{lightblue}$\mathbf{0.530 \pm 0.002}$
        & \cellcolor{lightblue}$\mathbf{0.567 \pm 0.002}$ \\
        EMM
        & $0.548 \pm 0.003$
        & $0.656 \pm 0.002$
        & $0.583 \pm 0.003$
        & $0.506 \pm 0.003$ \\
        ModelNet
        & $0.569 \pm 0.003$
        & $0.677 \pm 0.001$
        & $0.570 \pm 0.003$
        & $0.506 \pm 0.001$ \\
        MetaDetect
        & $0.505 \pm 0.000$
        & $0.612 \pm 0.002$
        & $0.642 \pm 0.002$
        & $0.460 \pm 0.003$ \\
        \bottomrule
    \end{tabular}%
    }
\end{table*}

Table~\ref{tab:adversarial_results} summarises uncertainty performance aggregated across all five adversarial attacks and their evaluated strengths. \methodname\ achieves the strongest overall results on CelebA and VisDrone, leading all four metrics on both datasets. On CelebA, it achieves a TP-FP AUROC of $0.825 \pm 0.020$ and FP AUPRC of $0.850 \pm 0.025$, substantially outperforming the comparative methods. Similarly, on VisDrone, \methodname\ obtains the highest TP-FP AUROC of $0.634 \pm 0.003$ and FP AUPRC of $0.717 \pm 0.003$, alongside the strongest selective-prediction performance.

Performance is more dataset-dependent on COCO and TT100K. MetaDetect performs best on COCO, achieving a TP-FP AUROC of $0.532 \pm 0.000$, while EMM is strongest on TT100K with $0.616 \pm 0.006$. Nevertheless, \methodname\ remains competitive on both datasets and achieves the best result in 8 of the 16 adversarial comparisons overall. These results show that \methodname\ maintains strong uncertainty quality under adversarial perturbation, particularly on CelebA and VisDrone, while robustness remains dependent on the underlying detection domain.

\subsection{Distributional Shift Results}
\label{sec:appendix-Distributional Shift Results}

We next evaluate uncertainty robustness under covariate shift, where the input distribution changes while the underlying detection task remains unchanged. Such shifts can arise from variations in image quality, acquisition conditions, or preprocessing, and may cause otherwise accurate detectors to become poorly calibrated or overconfident. We therefore assess whether each uncertainty method can identify localisation degradation under a range of controlled input perturbations.

\begin{figure}[htb]
    \centering
    \begin{subfigure}[t]{\linewidth}
        \centering
        \includegraphics[width=\linewidth]{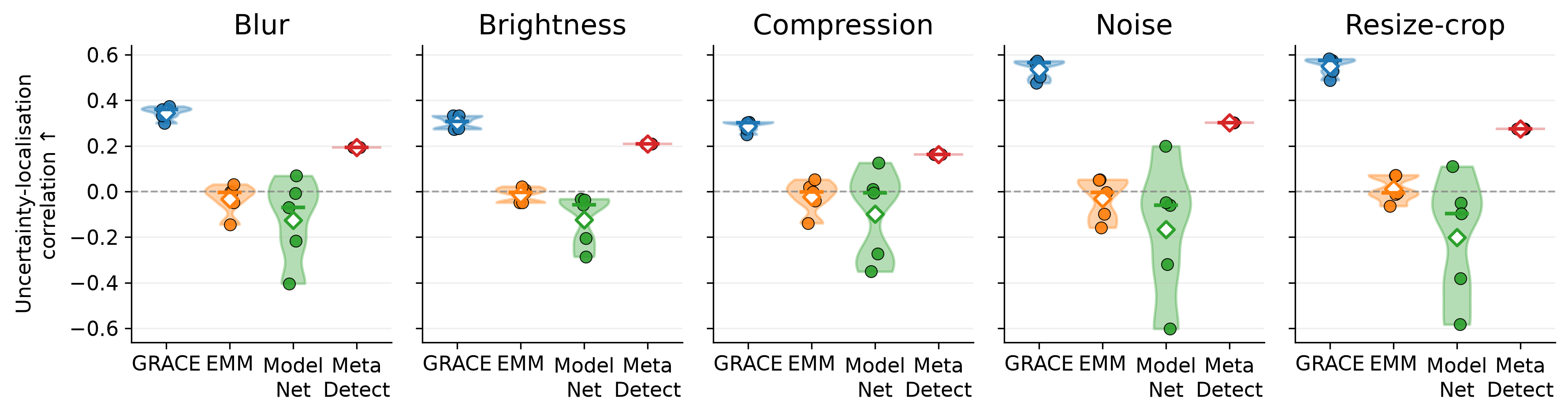}
        \caption{$t=2$}
        \label{fig:celeba_pert_t2}
    \end{subfigure}
    \begin{subfigure}[t]{\linewidth}
        \centering
        \includegraphics[width=\linewidth]{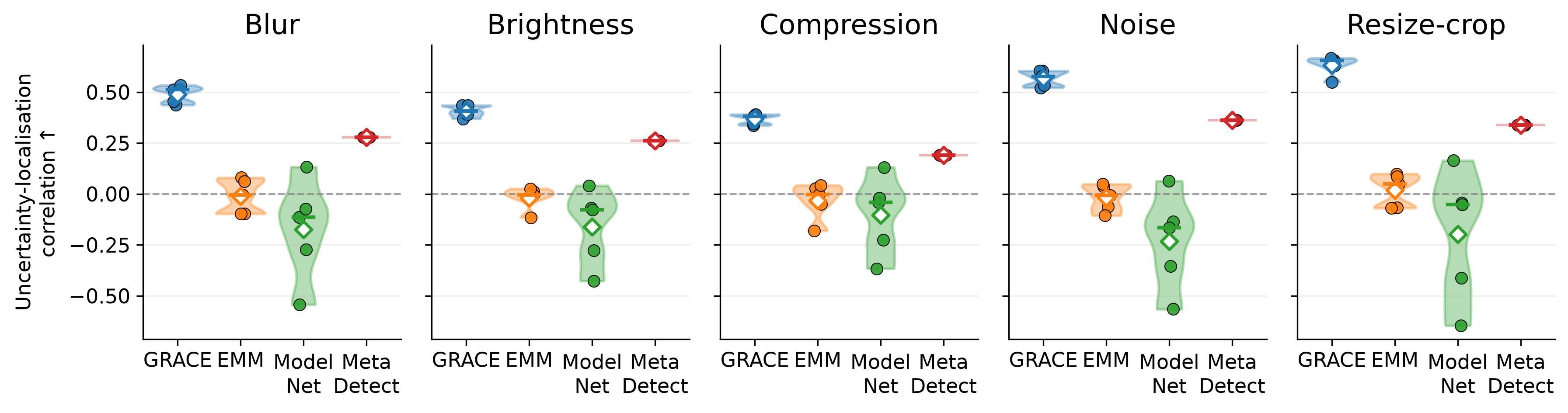}
        \caption{$t=5$}
        \label{fig:celeba_pert_t5}
    \end{subfigure}
    \begin{subfigure}[t]{\linewidth}
        \centering
        \includegraphics[width=\linewidth]{Images/celeba_t8_perturbation_violin_pearson_jitter.png}
        \caption{$t=8$}
        \label{fig:celeba_pert_t8}
    \end{subfigure}
\caption{CelebA correlations between uncertainty assigned to clean detections and their localisation error under five perturbations at different severity levels. Violins show the distribution across five runs, dots individual runs, diamonds means, solid horizontal lines medians, and the dashed line zero correlation; higher Pearson $r$ is better.}
\label{fig:celeba_pert_all}
\end{figure}

\begin{figure}[htb]
    \centering
    \begin{subfigure}[t]{\linewidth}
        \centering
        \includegraphics[width=\linewidth]{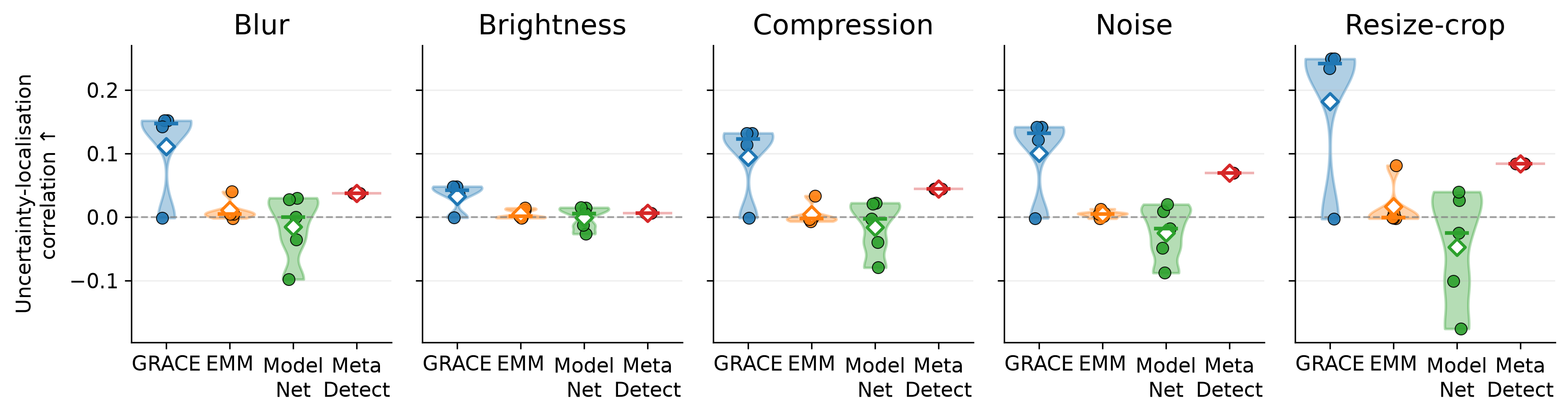}
        \caption{$t=2$}
        \label{fig:coco_pert_t2}
    \end{subfigure}
    \begin{subfigure}[t]{\linewidth}
        \centering
        \includegraphics[width=\linewidth]{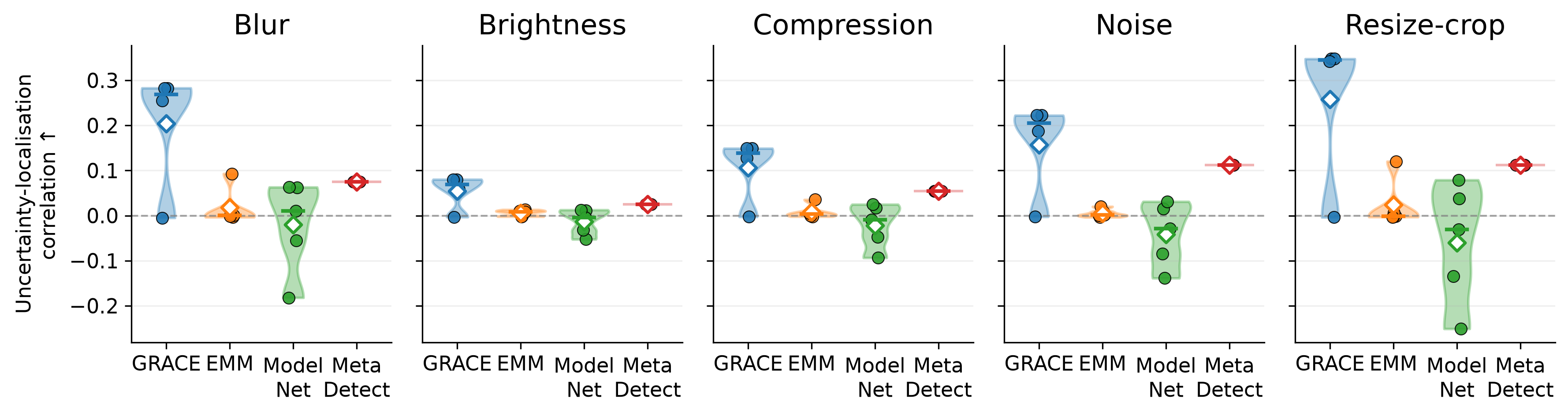}
        \caption{$t=5$}
        \label{fig:coco_pert_t5}
    \end{subfigure}
    \begin{subfigure}[t]{\linewidth}
        \centering
        \includegraphics[width=\linewidth]{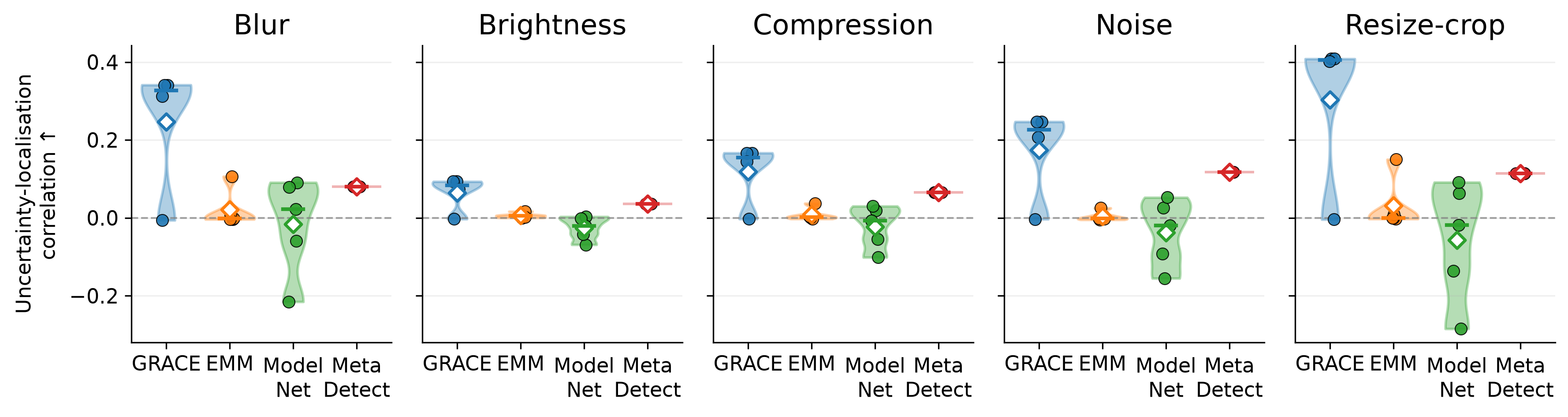}
        \caption{$t=8$}
        \label{fig:coco_pert_all}
    \end{subfigure}
\caption{Coco correlations between uncertainty assigned to clean detections and their localisation error under five perturbations at different severity levels. Violins show the distribution across five runs, dots individual runs, diamonds means, solid horizontal lines medians, and the dashed line zero correlation; higher Pearson $r$ is better.}
\label{fig:coco_pert_t8}
\end{figure}

\begin{figure}[htb]
    \centering
    \begin{subfigure}[t]{\linewidth}
        \centering
        \includegraphics[width=\linewidth]{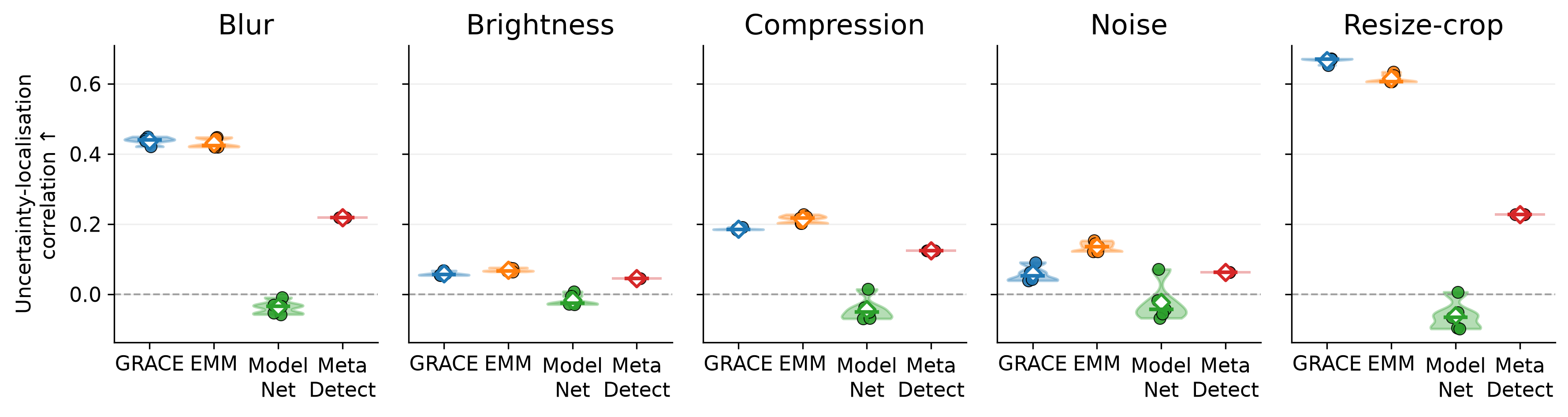}
        \caption{$t=2$}
        \label{fig:tt100k_pert_t2}
    \end{subfigure}
    \begin{subfigure}[t]{\linewidth}
        \centering
        \includegraphics[width=\linewidth]{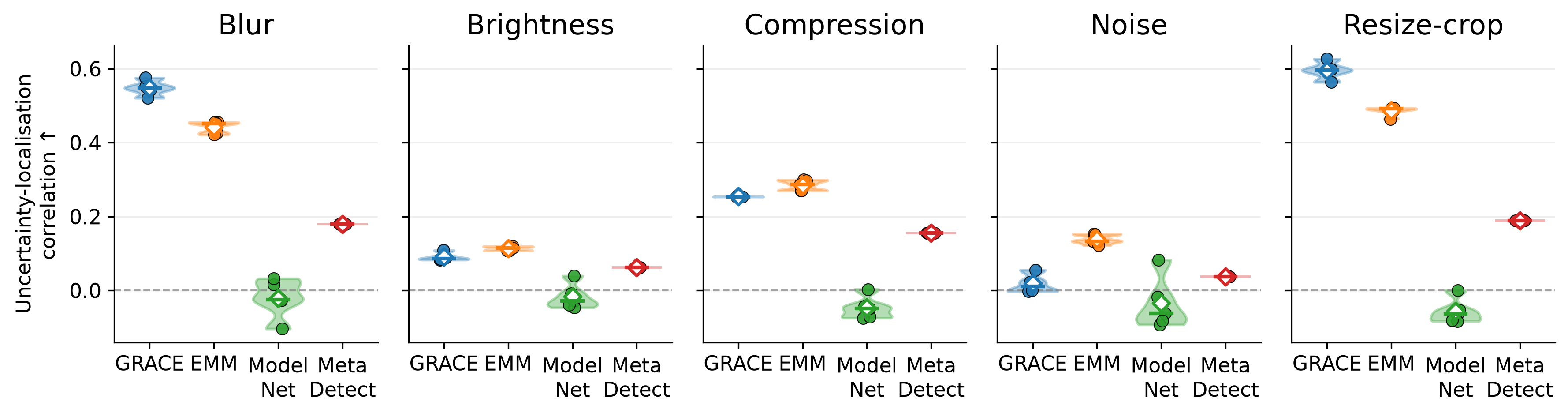}
        \caption{$t=5$}
        \label{fig:tt100k_pert_t5}
    \end{subfigure}
    \begin{subfigure}[t]{\linewidth}
        \centering
        \includegraphics[width=\linewidth]{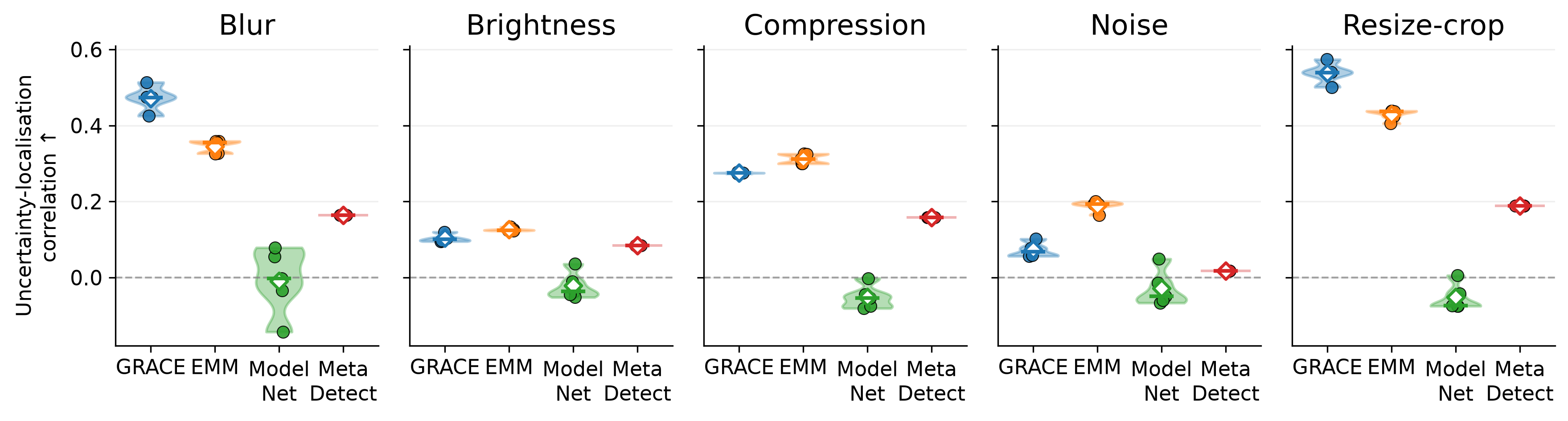}
        \caption{$t=8$}
        \label{fig:tt100k_pert_t8}
    \end{subfigure}
\caption{TT100K correlations between uncertainty assigned to clean detections and their localisation error under five perturbations at different severity levels. Violins show the distribution across five runs, dots individual runs, diamonds means, solid horizontal lines medians, and the dashed line zero correlation; higher Pearson $r$ is better.}
\label{fig:tt100k_pert_all}
\end{figure}

\begin{figure}[htb]
    \centering
    \begin{subfigure}[t]{\linewidth}
        \centering
        \includegraphics[width=\linewidth]{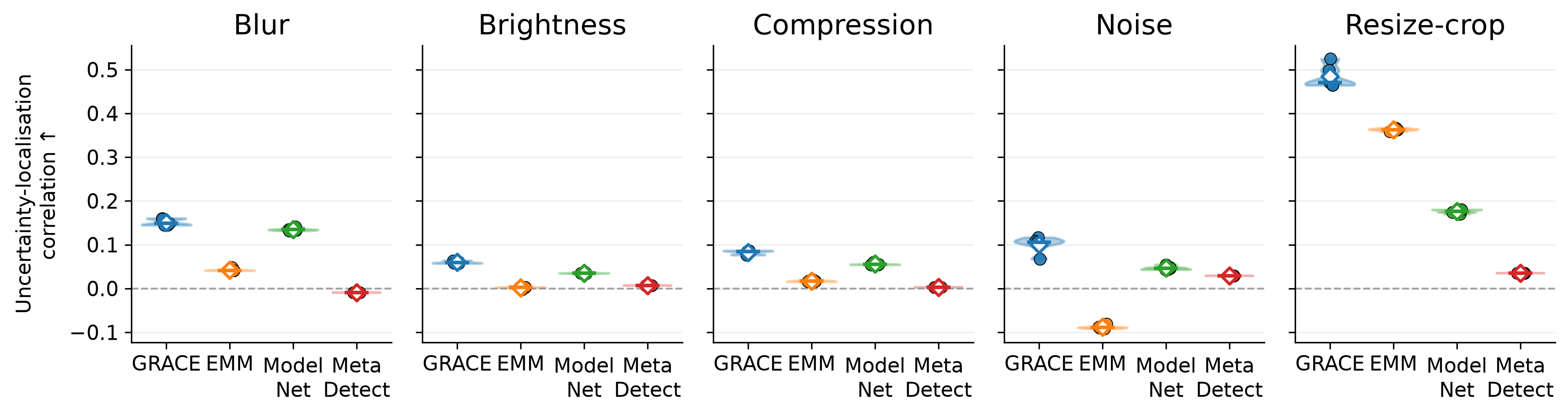}
        \caption{$t=2$}
        \label{fig:visdrone_pert_t2}
    \end{subfigure}
    \begin{subfigure}[t]{\linewidth}
        \centering
        \includegraphics[width=\linewidth]{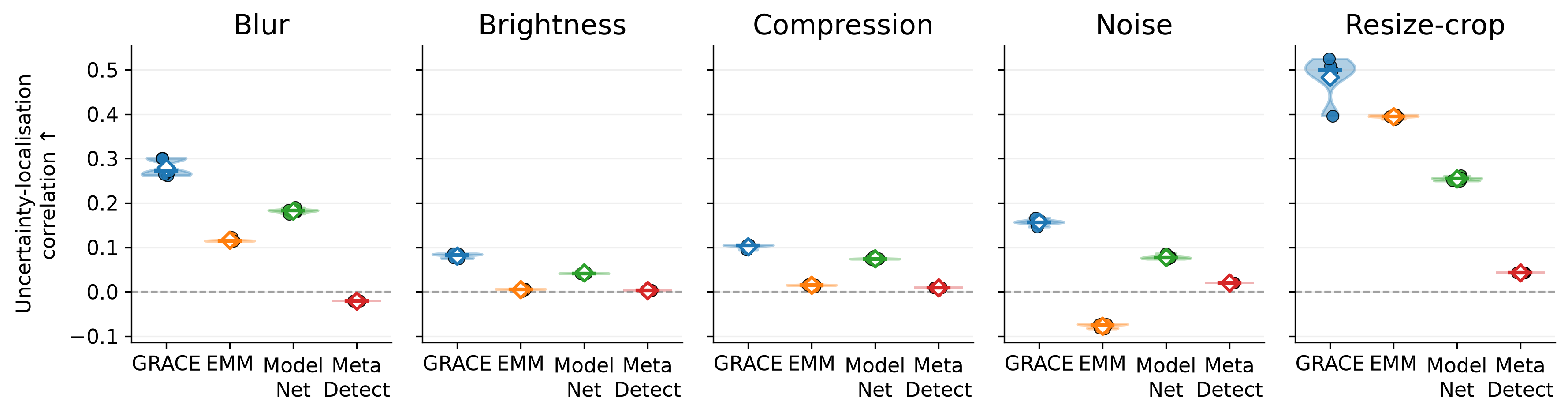}
        \caption{$t=5$}
        \label{fig:visdrone_pert_t5}
    \end{subfigure}
    \begin{subfigure}[t]{\linewidth}
        \centering
        \includegraphics[width=\linewidth]{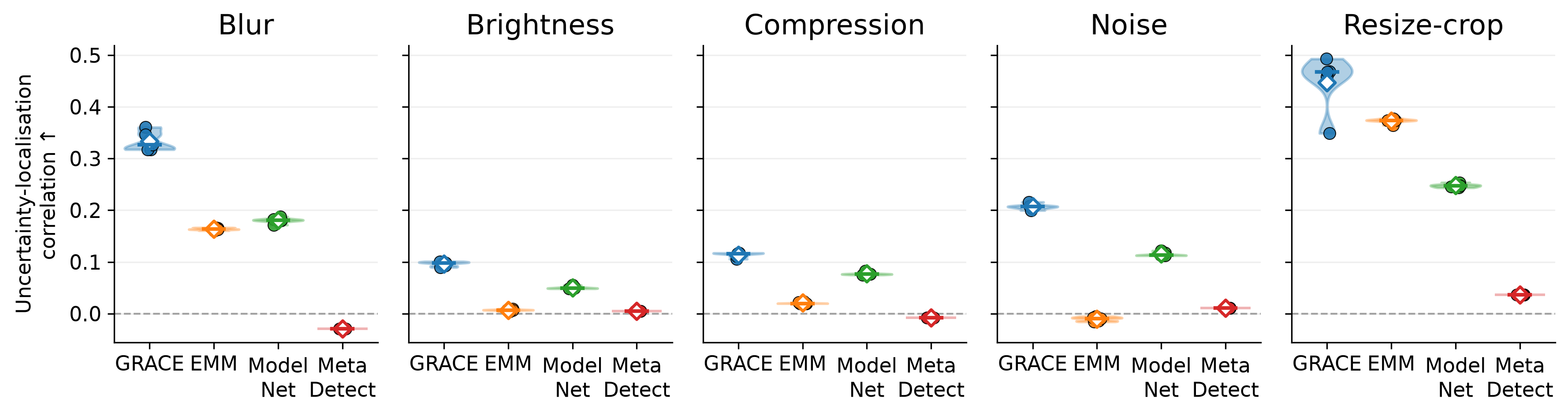}
        \caption{$t=8$}
        \label{fig:visdrone_pert_t8}
    \end{subfigure}
\caption{VisDrone correlations between uncertainty assigned to clean detections and their localisation error under five perturbations at different severity levels. Violins show the distribution across five runs, dots individual runs, diamonds means, solid horizontal lines medians, and the dashed line zero correlation; higher Pearson $r$ is better.}
\label{fig:visdrone_pert_all}
\end{figure}

Figures~\ref{fig:celeba_pert_all}-~\ref{fig:visdrone_pert_all} examine whether uncertainty assigned to the original clean detections predicts their subsequent localisation error under blur, brightness, compression, noise, and resize-crop perturbations at severities $t\in{2,5,8}$. A larger positive Pearson correlation therefore indicates that detections identified as uncertain before the shift are also those that degrade most strongly after perturbation.

On CelebA (Figure~\ref{fig:celeba_pert_all}), \methodname\ consistently exhibits the strongest positive relationship across perturbations and severities. At $t=2$, correlations are already approximately $0.35$ for blur, $0.31$ for brightness, $0.29$ for compression, and above $0.50$ for both noise and resize-crop. This relationship generally strengthens with severity: at $t=8$, \methodname\ reaches approximately $0.56$ under blur, $0.45$ under brightness, and around $0.65$ under resize-crop. MetaDetect remains positively correlated but substantially weaker, typically around $0.15$--$0.30$, while EMM remains close to zero. ModelNet frequently produces negative correlations, particularly under noise and resize-crop, indicating that its clean uncertainty does not reliably anticipate which face detections will become poorly localised.

A similar, although weaker, pattern is observed on COCO (Figure~\ref{fig:coco_pert_t8}). \methodname\ achieves positive correlations across all five shifts, increasing from approximately $0.14$ to $0.32$ for blur between $t=2$ and $t=8$, from around $0.13$ to $0.21$ for noise, and from approximately $0.23$ to $0.38$ for resize-crop. Brightness is less discriminative, remaining below approximately $0.10$ even at stronger severities. MetaDetect remains consistently positive but generally below $0.12$, while EMM stays close to zero and ModelNet is frequently negatively correlated. These results suggest that the clean \methodname\ uncertainty retains predictive information even in the considerably more heterogeneous COCO setting, although the relationship is naturally weaker than on CelebA.

TT100K (Figure~\ref{fig:tt100k_pert_all}) exhibits stronger shift predictability but also greater competition from EMM. \methodname\ achieves correlations of approximately $0.44$, $0.18$, and $0.66$ for blur, compression, and resize-crop at $t=2$, increasing to around $0.55$, $0.25$, and $0.59$ at $t=5$. At $t=8$, \methodname\ remains particularly strong for blur and resize-crop at approximately $0.47$ and $0.54$. EMM is similarly strong for several perturbations, reaching approximately $0.43$ under blur and $0.62$ under resize-crop at $t=2$, and slightly exceeding \methodname\ under compression and noise at higher severities. In contrast, ModelNet remains close to or below zero across most conditions, while MetaDetect produces moderate positive correlations, generally below approximately $0.25$. Thus, TT100K confirms that frozen-representation uncertainty can be effective for some structured domains, while \methodname\ remains particularly effective at anticipating degradation under blur and geometric resize-crop shifts.

Finally, VisDrone (Figure~\ref{fig:visdrone_pert_all}) shows a clear severity-dependent increase for \methodname. Under blur, its correlation rises from approximately $0.15$ at $t=2$ to $0.28$ at $t=5$ and $0.33$ at $t=8$, while noise increases from approximately $0.10$ to $0.21$. Resize-crop produces the strongest relationship throughout, remaining around $0.46$-$0.50$. EMM is competitive under resize-crop but substantially weaker for the remaining perturbations, while ModelNet achieves moderate positive correlations and MetaDetect remains near zero. Overall, Figures 14--17 show that \methodname\ most consistently assigns greater clean uncertainty to detections that later experience localisation degradation, with this relationship generally becoming stronger as shift severity increases.

\begin{figure}[htb]
    \centering
    \begin{subfigure}[t]{\linewidth}
        \centering
        \includegraphics[width=\linewidth]{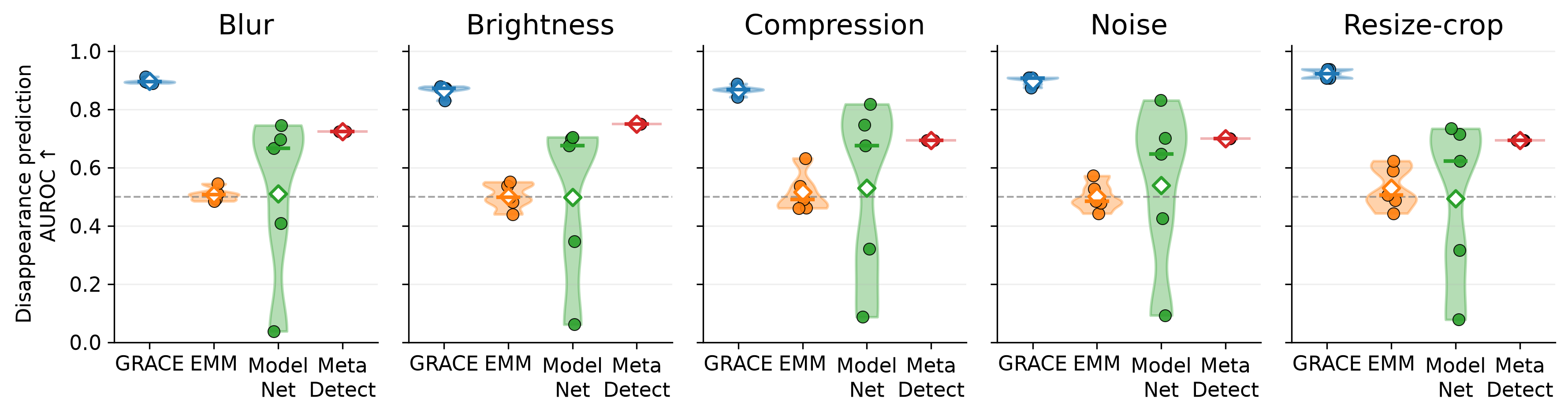}
        \caption{$t=2$}
        \label{fig:celeba_disp_t2}
    \end{subfigure}
    \begin{subfigure}[t]{\linewidth}
        \centering
        \includegraphics[width=\linewidth]{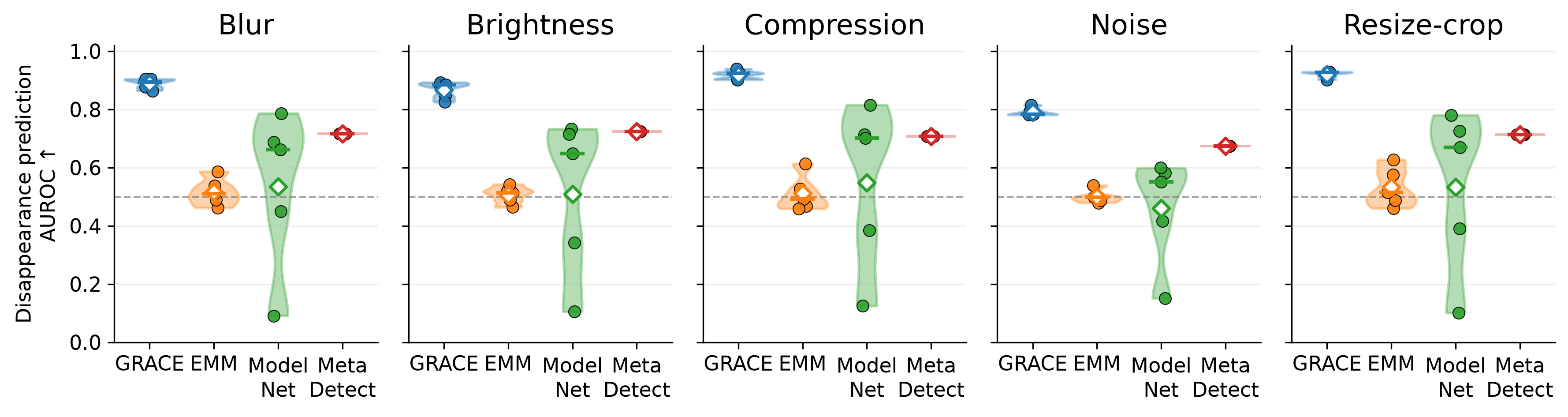}
        \caption{$t=5$}
        \label{fig:celeba_disp_t5}
    \end{subfigure}
    \begin{subfigure}[t]{\linewidth}
        \centering
        \includegraphics[width=\linewidth]{Images/celeba_t8_perturbation_violin_disappearance_auroc.png}
        \caption{$t=8$}
        \label{fig:celeba_disp_t8}
    \end{subfigure}
\caption{CelebA disappearance-prediction AUROC using uncertainty assigned to clean detections under five perturbations at different severity levels. Violins show the distribution across five runs, dots individual runs, diamonds means, solid horizontal lines medians, and the dashed line zero correlation; a higher AUROC is better.}
\label{fig:celeba_disp_all}
\end{figure}

\begin{figure}[htb]
    \centering
    \begin{subfigure}[t]{\linewidth}
        \centering
        \includegraphics[width=\linewidth]{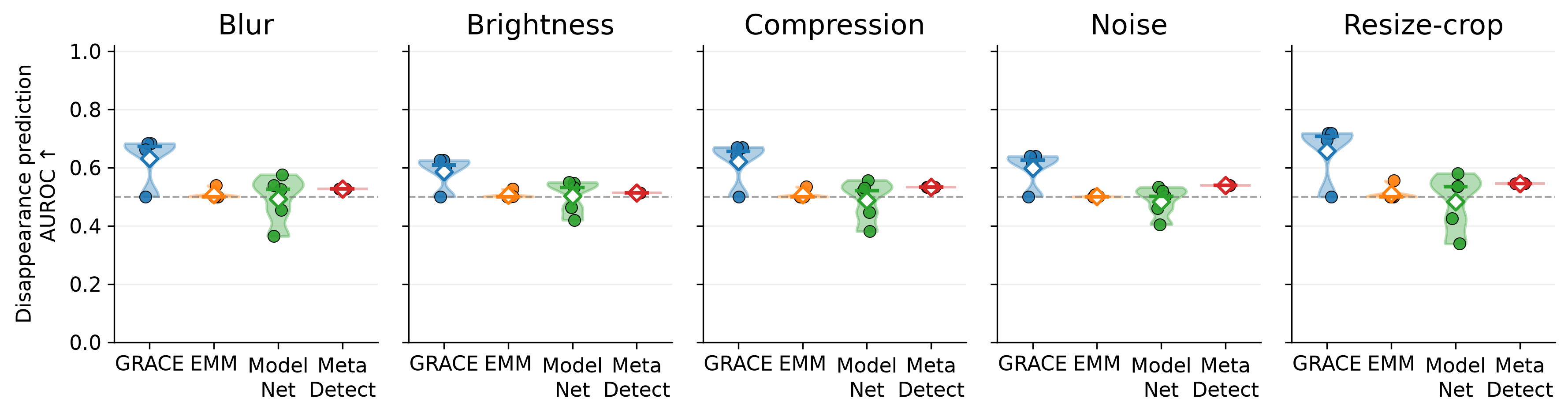}
        \caption{$t=2$}
        \label{fig:coco_disp_t2}
    \end{subfigure}
    \begin{subfigure}[t]{\linewidth}
        \centering
        \includegraphics[width=\linewidth]{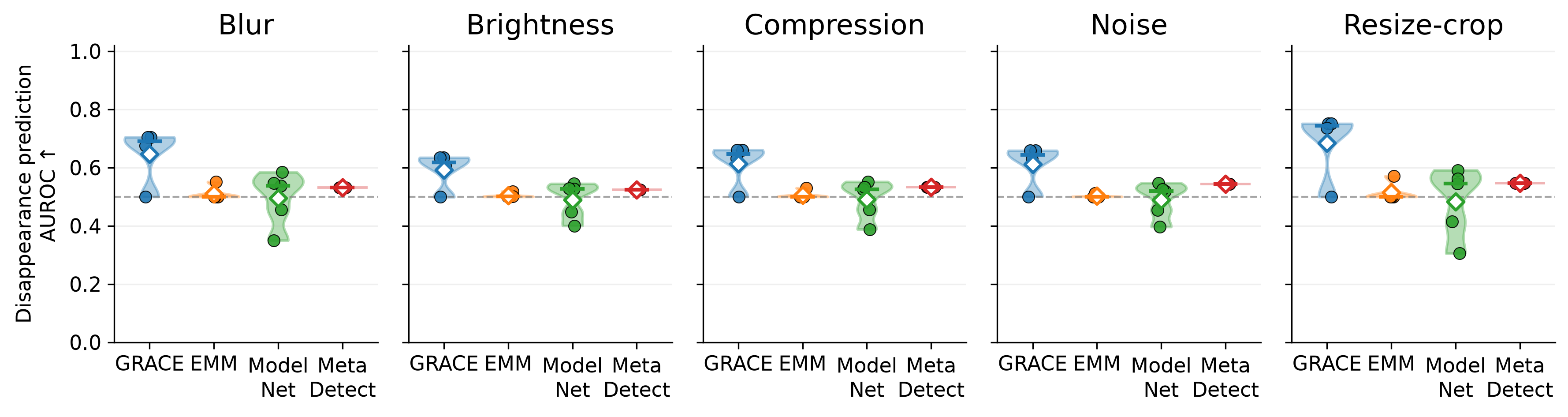}
        \caption{$t=5$}
        \label{fig:coco_disp_t5}
    \end{subfigure}
    \begin{subfigure}[t]{\linewidth}
        \centering
        \includegraphics[width=\linewidth]{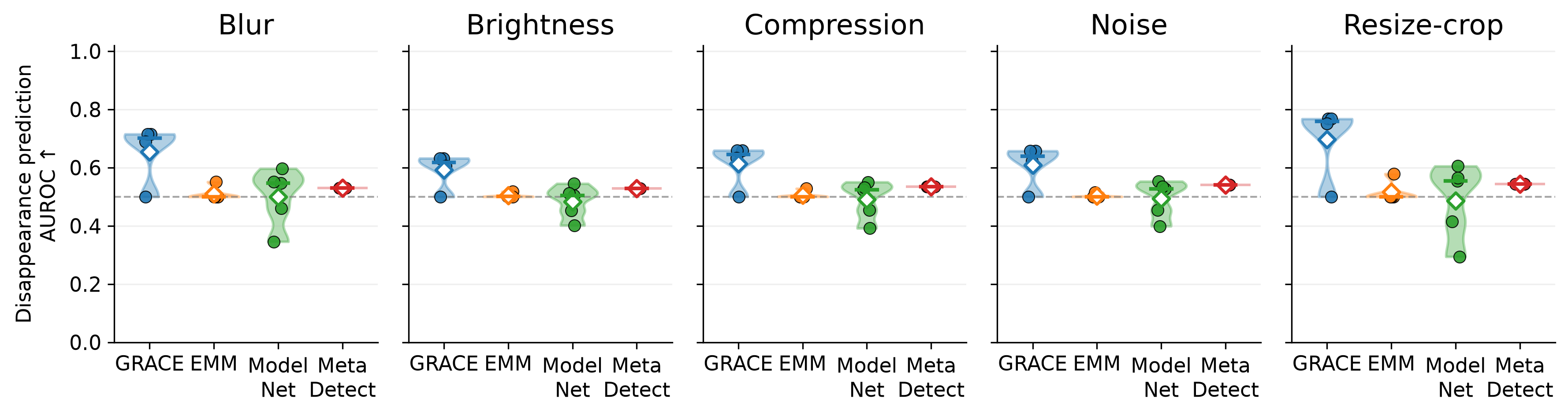}
        \caption{$t=8$}
        \label{fig:coco_disp_t8}
    \end{subfigure}
\caption{Coco disappearance-prediction AUROC using uncertainty assigned to clean detections under five perturbations at different severity levels. Violins show the distribution across five runs, dots individual runs, diamonds means, solid horizontal lines medians, and the dashed line zero correlation; a higher AUROC is better.}
\label{fig:coco_disp_all}
\end{figure}

\begin{figure}[htb]
    \centering
    \begin{subfigure}[t]{\linewidth}
        \centering
        \includegraphics[width=\linewidth]{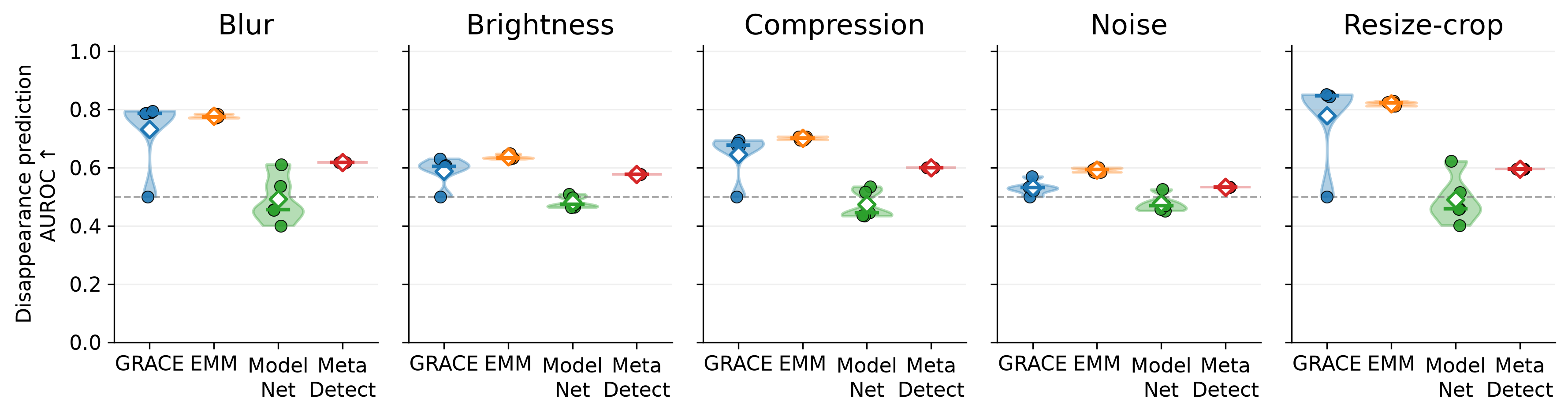}
        \caption{$t=2$}
        \label{fig:tt100k_disp_t2}
    \end{subfigure}
    \begin{subfigure}[t]{\linewidth}
        \centering
        \includegraphics[width=\linewidth]{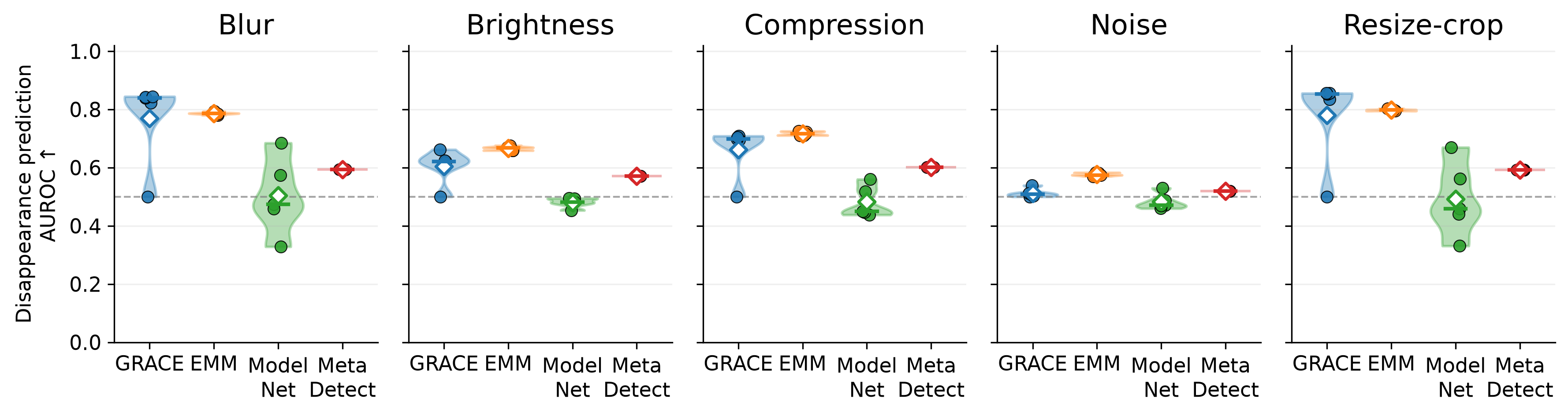}
        \caption{$t=5$}
        \label{fig:tt100k_disp_t5}
    \end{subfigure}
    \begin{subfigure}[t]{\linewidth}
        \centering
        \includegraphics[width=\linewidth]{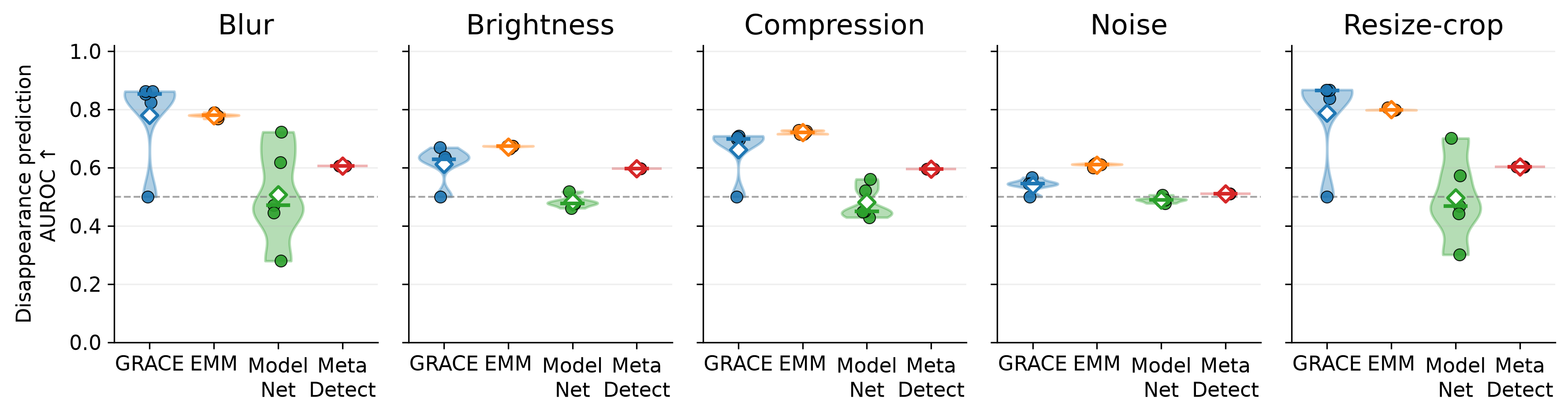}
        \caption{$t=8$}
        \label{fig:tt100k_disp_t8}
    \end{subfigure}
\caption{TT100K disappearance-prediction AUROC using uncertainty assigned to clean detections under five perturbations at different severity levels. Violins show the distribution across five runs, dots individual runs, diamonds means, solid horizontal lines medians, and the dashed line zero correlation; a higher AUROC is better.}
\label{fig:tt100k_disp_all}
\end{figure}

\begin{figure}[htb]
    \centering
    \begin{subfigure}[t]{\linewidth}
        \centering
        \includegraphics[width=\linewidth]{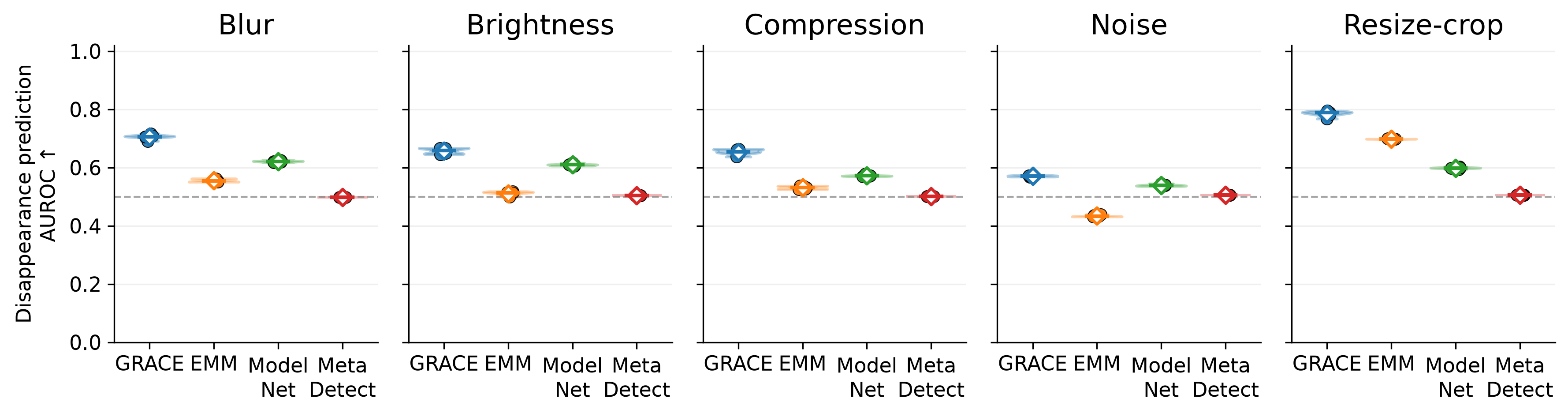}
        \caption{$t=2$}
        \label{fig:visdrone_disp_t2}
    \end{subfigure}
    \begin{subfigure}[t]{\linewidth}
        \centering
        \includegraphics[width=\linewidth]{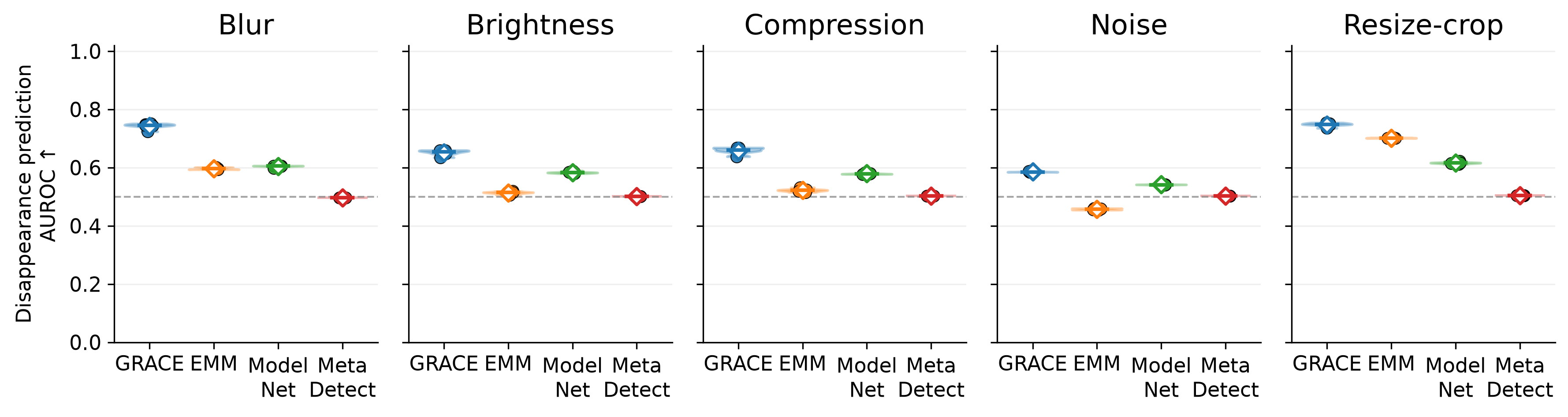}
        \caption{$t=5$}
        \label{fig:visdrone_disp_t5}
    \end{subfigure}
    \begin{subfigure}[t]{\linewidth}
        \centering
        \includegraphics[width=\linewidth]{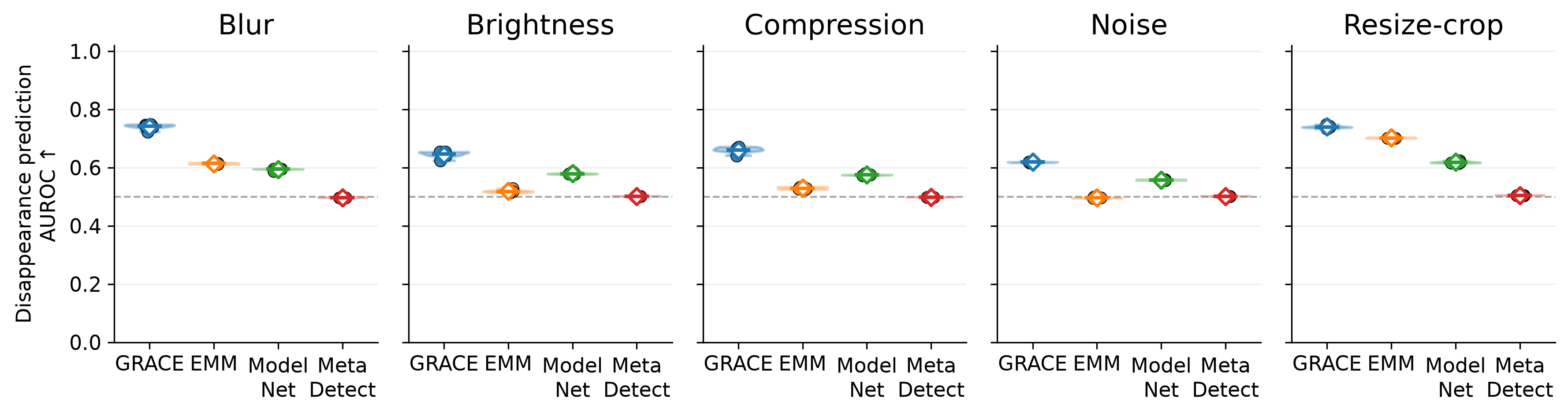}
        \caption{$t=8$}
        \label{fig:visdrone_disp_t8}
    \end{subfigure}
\caption{VisDrone disappearance-prediction AUROC using uncertainty assigned to clean detections under five perturbations at different severity levels.. Violins show the distribution across five runs, dots individual runs, diamonds means, solid horizontal lines medians, and the dashed line zero correlation; a higher AUROC is better.}
\label{fig:visdrone_disp_all}
\end{figure}

Figures~\ref{fig:celeba_disp_all}-\ref{fig:visdrone_disp_all} complement the localisation-degradation analysis by evaluating whether uncertainty assigned to a clean detection can predict whether that detection disappears entirely after covariate shift. A disappearance-prediction AUROC above $0.50$ indicates that clean uncertainty contains information about subsequent detection failure, with higher values indicating stronger predictive ability. Results are reported across blur, brightness, compression, noise, and resize-crop at severities $t\in{2,5,8}$.

On CelebA (Figure~\ref{fig:celeba_disp_all}), \methodname\ provides consistently strong disappearance prediction across both perturbation type and severity. At $t=2$, AUROC is approximately $0.87$--$0.92$ across the five shifts, substantially above EMM, which remains around $0.50$-$0.58$, and MetaDetect at approximately $0.69$-$0.75$. This behaviour is largely preserved as severity increases: at $t=8$, \methodname\ achieves approximately $0.88$ under blur, $0.86$ under brightness, $0.89$ under compression, and $0.92$ under resize-crop. Noise is the main exception, where performance decreases to approximately $0.65$ at $t=8$, although \methodname\ remains stronger than the comparative approaches. ModelNet exhibits substantially greater variation between runs, with mean performance generally around $0.50$--$0.70$. Thus, \methodname\ uncertainty identifies detections vulnerable to complete failure even before the perturbation is applied, and does so consistently across substantially different image shifts.

On COCO (Figure~\ref{fig:coco_disp_all}), absolute separation is smaller, but \methodname\ remains the strongest approach across nearly all perturbations and severity levels. At $t=2$, \methodname\ achieves approximately $0.62$-$0.68$ under blur, brightness, compression, and noise, and around $0.70$ under resize-crop, whereas EMM and MetaDetect remain close to $0.50$-$0.55$. At $t=8$, \methodname\ reaches approximately $0.69$ under blur, $0.61$ under brightness, $0.64$ under compression, $0.64$ under noise, and approximately $0.74$ under resize-crop. ModelNet is occasionally competitive but remains close to $0.50$ on average and exhibits greater variability. These results are notable given the diversity of COCO, where predicting complete detection loss from the clean input is inherently more difficult than on the single-class CelebA setting.

TT100K (Figure~\ref{fig:tt100k_disp_all}) exhibits a different pattern, with both \methodname\ and EMM providing strong disappearance prediction. At $t=2$, \methodname\ achieves approximately $0.76$ for blur, $0.59$ for brightness, $0.67$ for compression, $0.54$ for noise, and above $0.80$ for resize-crop. EMM is stronger for several perturbations, reaching approximately $0.78$ under blur, $0.64$ under brightness, $0.72$ under compression, and $0.82$ under resize-crop. As severity increases, \methodname\ becomes particularly strong for blur and resize-crop, reaching approximately $0.84$ for both at $t=8$, while EMM remains stronger under brightness, compression, and noise. ModelNet generally remains around or below $0.50$, while MetaDetect produces moderate performance of approximately $0.52$-$0.62$. This mirrors the core TT100K results, where EMM is particularly competitive, but shows that \methodname\ remains highly effective for shifts that strongly affect object scale and spatial structure.

Finally, VisDrone (Figure~\ref{fig:visdrone_disp_all}) shows strong and highly consistent behaviour for \methodname. At $t=2$, \methodname\ achieves disappearance-prediction AUROCs of approximately $0.70$ for blur, $0.65$ for brightness and compression, $0.57$ for noise, and $0.78$ for resize-crop. These values are generally maintained or improved at stronger severities, reaching approximately $0.74$, $0.64$, $0.66$, $0.61$, and $0.74$, respectively, at $t=8$. EMM is competitive under resize-crop at approximately $0.70$, but remains substantially weaker under the other perturbations, while ModelNet typically lies between \methodname\ and EMM. MetaDetect remains close to chance at approximately $0.50$ across almost all settings. Overall, Figures 18--21 demonstrate that \methodname's clean uncertainty is informative not only of subsequent localisation degradation, but also of whether a detection will disappear entirely under covariate shift, with particularly consistent performance on CelebA, COCO, and VisDrone.

\subsection{Ablation Study}
\label{sec:appendix-Ablation Study}

We next examine the sensitivity of \methodname\ to the hyperparameters it introduced. The ablation study evaluates the contribution of the noise-driven curriculum and saliency calibration stages, together with the effect of curriculum length $T$, corruption rate $\gamma$, and relevance coverage threshold $\eta$. Unless otherwise stated, all remaining training and evaluation settings are held fixed. These experiments assess both the robustness of \methodname\ to reasonable hyperparameter variation and the extent to which each component contributes to localisation-aware uncertainty quality.

\begin{figure*}[tb]
    \centering
    \includegraphics[width=\linewidth]{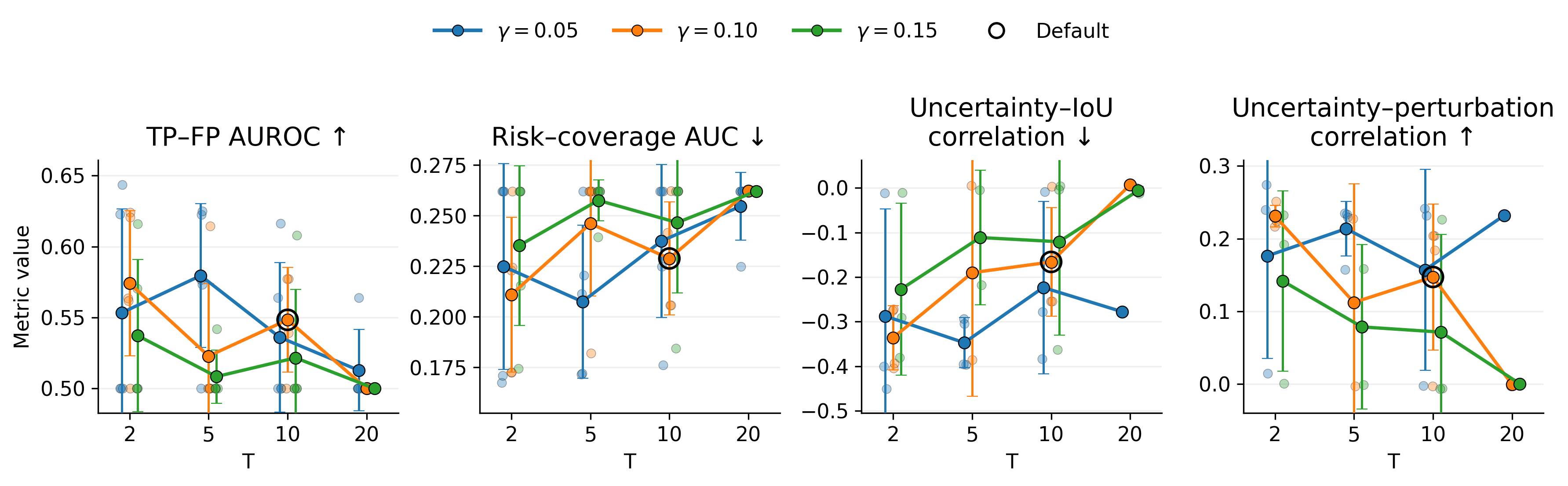}
    \caption{Ablation of curriculum length $T$ and corruption rate $\gamma$ on the COCO dataset. We report TP–FP AUROC, risk–coverage AUC, uncertainty–IoU correlation, and uncertainty–perturbation correlation for a combination of $T$ and $\gamma$.}
    \label{fig:coco_t_gamma_ablation}
\end{figure*}

\begin{figure*}[tb]
    \centering
    \includegraphics[width=\linewidth]{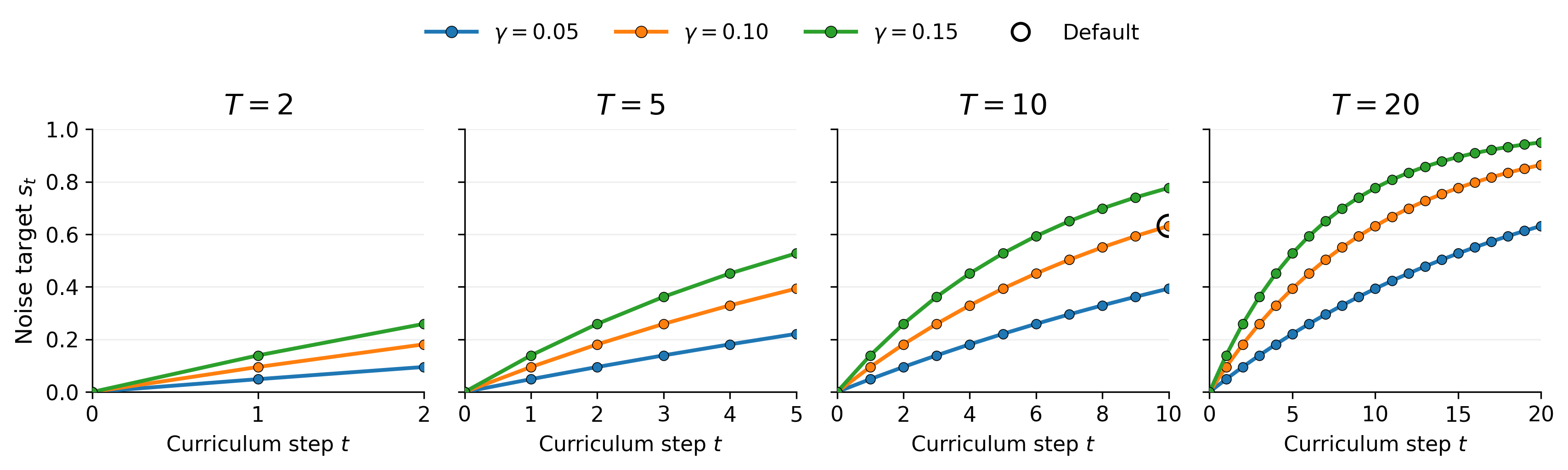}
    \caption{Noise-driven curriculum schedules for varying curriculum lengths $T$ and corruption rate $\gamma$. Each curve shows the target corruption level ($s_t = 1-e^{-\gamma t}$) across curriculum steps $t$.}
    \label{fig:guide_curriculum_schedules}
\end{figure*}

Figure~\ref{fig:coco_t_gamma_ablation} examines the interaction between curriculum length $T$ and corruption rate $\gamma$ on COCO, while Figure~\ref{fig:guide_curriculum_schedules} visualises the corresponding corruption schedules $s_t=1-e^{-\gamma t}$. Together, these figures show that the choice of $T$ and $\gamma$ controls not only how many curriculum stages are used, but also the maximum corruption level reached during training. For example, at $T=10$, the final noise target is approximately $0.39$, $0.63$, and $0.78$ for $\gamma=0.05$, $0.10$, and $0.15$, respectively, while at $T=20$ these increase to approximately $0.63$, $0.86$, and $0.95$.

Shorter or more gradual curricula generally provide stronger uncertainty quality. The strongest TP-FP AUROC is obtained at $T=5,\gamma=0.05$, reaching approximately $0.58$, while the same configuration also achieves the lowest risk-coverage AUC at approximately $0.21$ and the strongest uncertainty-IoU correlation at approximately $-0.35$. Similarly, $T=2,\gamma=0.10$ performs strongly across the evaluated metrics, with a TP-FP AUROC of approximately $0.57$ and uncertainty-perturbation correlation of approximately $0.23$. These settings correspond to comparatively mild final corruption targets of approximately $0.22$ and $0.18$, respectively.

In contrast, the $T=20$ schedules expose the meta-model to substantially stronger corruption. As shown in Figure 23, $\gamma=0.10$ and $\gamma=0.15$ reach final noise targets of approximately $0.86$ and $0.95$. Correspondingly, Figure 22 shows TP-FP AUROC approaching $0.50$ and both uncertainty-IoU and uncertainty-perturbation correlations moving towards zero for these settings. This suggests that excessively severe curricula may introduce inputs that are too degraded to provide useful localisation-reliability supervision. The smaller rate $\gamma=0.05$ is more robust at $T=20$, where the final corruption target remains substantially lower at approximately $0.63$.

The default configuration, $T=10$ and $\gamma=0.10$, reaches a final corruption target of approximately $0.63$ and provides a balanced intermediate setting. It achieves a TP-FP AUROC of approximately $0.55$, risk-coverage AUC of $0.23$, uncertainty-IoU correlation of approximately $-0.16$, and uncertainty-perturbation correlation of approximately $0.15$. Although individual configurations outperform the default on particular metrics, the combined results indicate that moderate curriculum severity provides the most reliable trade-off, whereas excessively long and aggressive schedules tend to weaken localisation-aware uncertainty.

\begin{figure*}[tb]
    \centering
    \includegraphics[width=\linewidth]{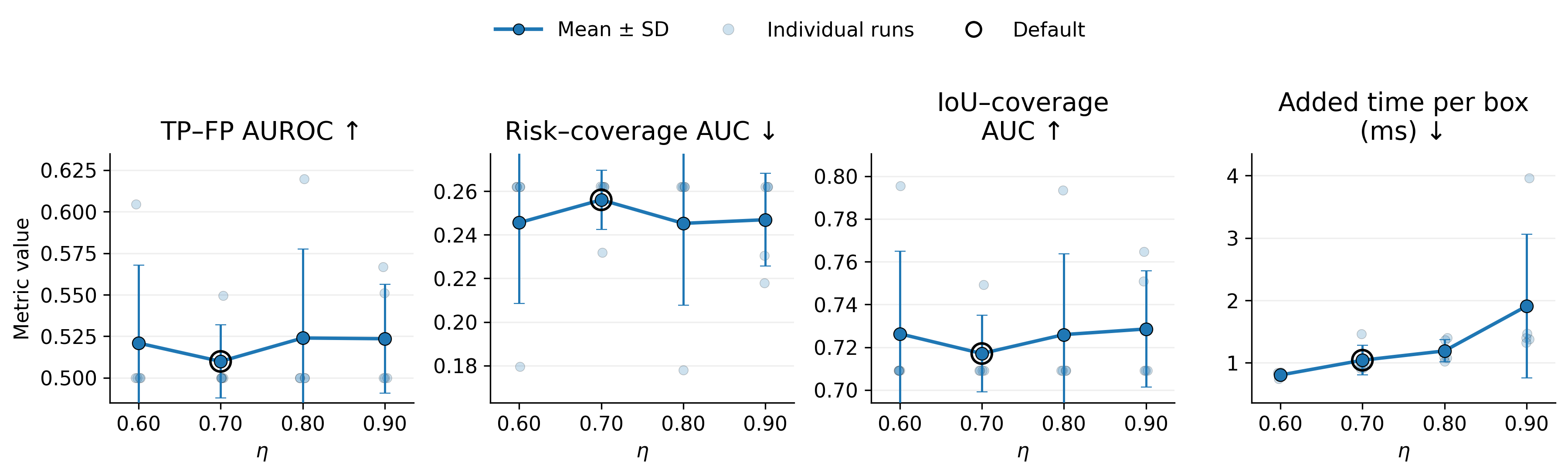}
    \caption{Ablation of the saliency coverage threshold $\eta$ on COCO. We report TP–FP AUROC, risk–coverage AUC, IoU–coverage AUC, and added inference time per detected box.}
    \label{fig:coco_eta_ablation}
\end{figure*}

Figure~\ref{fig:coco_eta_ablation} examines the effect of the saliency relevance coverage threshold $\eta$ on COCO. Overall, performance is relatively stable across $\eta\in{0.60,0.70,0.80,0.90}$, indicating that \methodname\ is not highly sensitive to the precise proportion of relevance mass retained during layer selection. TP-FP AUROC varies only modestly, from approximately $0.51$ at $\eta=0.70$ to around $0.52$--$0.53$ at $\eta=0.60$, $0.80$, and $0.90$. Similarly, risk-coverage AUC remains within approximately $0.245$--$0.257$, while IoU-coverage AUC varies only slightly between around $0.72$ and $0.73$.

The default $\eta=0.70$ provides a strong efficiency-performance trade-off. Although $\eta=0.80$ and $\eta=0.90$ achieve marginally higher mean TP-FP AUROC and IoU-coverage AUC, the differences are small relative to the variation across runs. In contrast, computational cost increases consistently as $\eta$ grows: added time per box rises from approximately $0.8$ms at $\eta=0.60$ to around $1.0$ms at the default $\eta=0.70$, $1.2$ms at $\eta=0.80$, and nearly $1.9$ms at $\eta=0.90$.

These results suggest that retaining additional lower-ranked salient layers provides limited uncertainty-quality improvement while increasing inference cost. Consequently, $\eta=0.70$ is selected as the default, preserving most localisation-relevant information while avoiding the additional computational overhead associated with larger salient-layer sets.


\section{Additional Discussion}
\label{sec:appendix-Additional Discussion}

The relative improvement provided by \methodname\ varies across datasets, with TT100K representing the clearest case in which EMM remains particularly competitive. Rather than indicating a failure of the proposed uncertainty-guidance mechanism, this result suggests that the frozen detector representations on TT100K already encode a comparatively strong localisation-reliability signal, reducing the additional benefit available from curriculum-based refinement. This interpretation is supported consistently across the clean, covariate-shift, and adversarial evaluations, where EMM also performs strongly on TT100K. However, \methodname\ still outperforms EMM on this dataset, showing the additional benefit of the saliency calibration and the curriculum training. Importantly, \methodname\ remains competitive in this setting while providing substantially clearer gains on CelebA and VisDrone, indicating that its benefit is greatest where reliability is not already well captured by the frozen detector representation.

\methodname\ is designed primarily to estimate localisation reliability rather than to optimise a calibrated probability of binary detection success. This distinction explains the different behaviour observed across the calibration metrics. MetaDetect achieves lower Brier and success-score calibration errors on several datasets, reflecting its explicit prediction-quality objective, whereas \methodname\ more consistently assigns greater uncertainty to poorly localised detections and provides stronger uncertainty-IoU alignment. These results therefore reflect complementary objectives rather than a contradiction: accurate calibration of detection success does not necessarily imply that uncertainty is well aligned with localisation quality. Accordingly, the intended role of \methodname\ is localisation-aware uncertainty estimation, for which its correlation and selective-prediction results provide the more direct evaluation.

\methodname\ assigns greater uncertainty to smaller predicted objects, particularly on TT100K and VisDrone. This relationship is expected in object detection, since smaller objects generally provide less spatial and visual evidence and are consequently more difficult to localise accurately. Detector confidence and localisation quality are therefore commonly associated with object scale~\citep{kuppers2020multivariate}, and a corresponding relationship with localisation uncertainty is not inherently undesirable. In this context, the observed size dependence is consistent with \methodname\ assigning greater uncertainty to detections that are intrinsically more difficult to localise, rather than indicating an independent source of bias in the uncertainty estimator.


\section{Experimental Setup}
\label{sec:appendix-Experimental Setup}

This section outlines our experimental setup in detail to ensure reproducibility and to make the evaluation protocol transparent. We first present the datasets employed in our evaluation, after which we describe the models used. Next, we summarise the comparative methods considered in our study and provide a thorough account of the model training process. Lastly, we report task-specific configurations and implementation details pertinent to each experimental setting.

\subsection{Datasets}
\label{sec:appendix-datasets}

\begin{figure*}[htb]
    \centering
    \begin{subfigure}{0.475\linewidth}
        \centering
        \includegraphics[width=\linewidth]{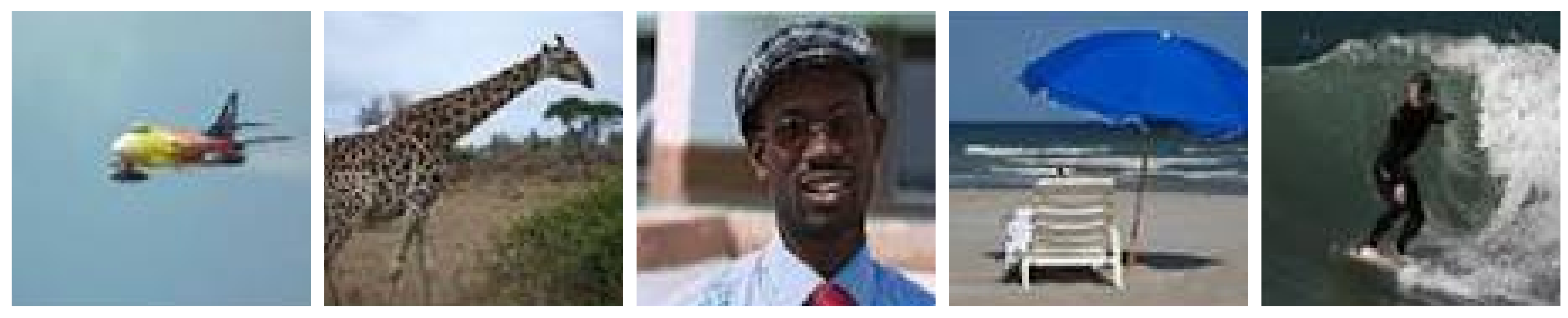}
        \caption{COCO}
    \end{subfigure}
    \hfill
    \begin{subfigure}{0.475\linewidth}
        \centering
        \includegraphics[width=\linewidth]{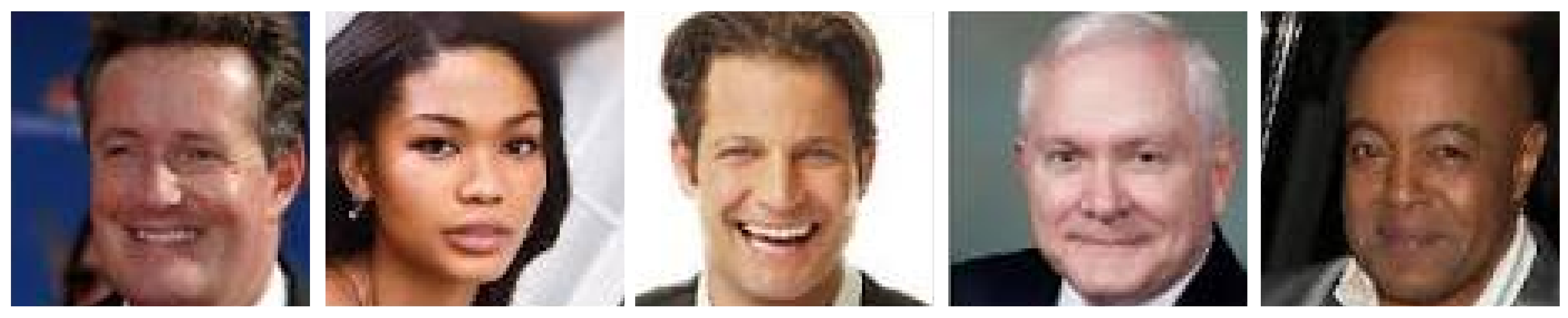}
        \caption{CelebA}
    \end{subfigure}
    \begin{subfigure}{0.475\linewidth}
        \centering
        \includegraphics[width=\linewidth]{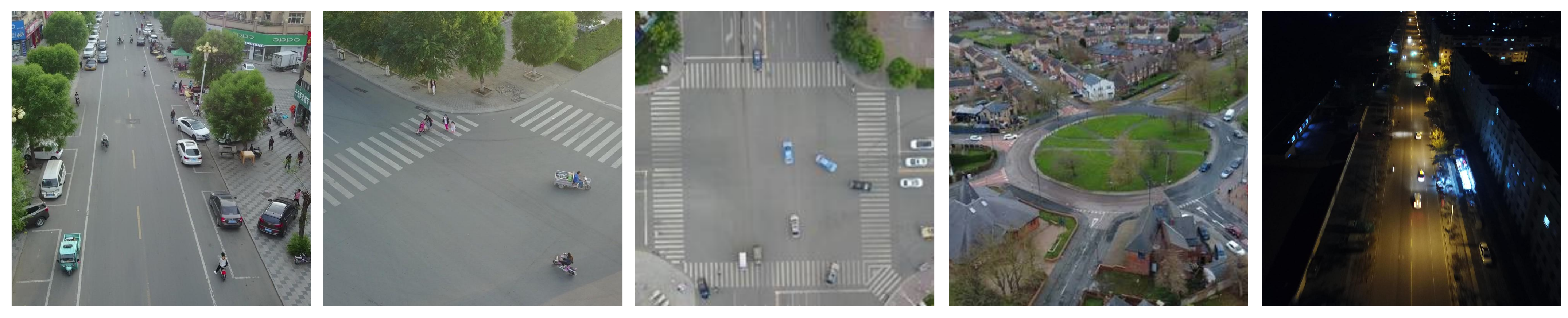}
        \caption{VisDrone}
    \end{subfigure}
    \hfill
    \begin{subfigure}{0.475\linewidth}
        \centering
        \includegraphics[width=\linewidth]{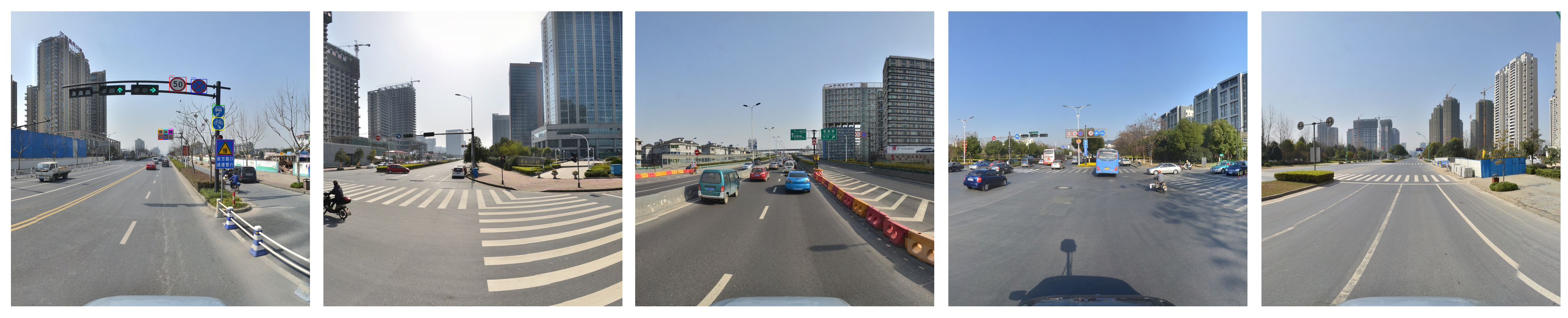}
        \caption{TT100K}
    \end{subfigure}
\caption{Example images from the datasets used in our experimental evaluation.}
\label{fig:appendix-dataset-vis}
\end{figure*}

We conduct evaluations using a broad suite of well-established object detection benchmarks that cover different domains and levels of task complexity. This design allows for a thorough assessment of repurposing across diverse conditions. Detailed information on each dataset, including sample counts, input resolution, class composition, and data splits, is provided below. Representative examples are displayed in Figure~\ref{fig:appendix-dataset-vis}.

\begin{itemize}[noitemsep, nolistsep]
    \item \textbf{COCO~\citep{lin2014microsoft}:} consists of variable-resolution RGB images depicting complex real-world scenes containing common objects (e.g., people, vehicles, animals, and household items) spanning 80 classes. We utilise the standard COCO 2017 split of 118,287 training images, 5,000 calibration images, and 40,670 test images. Images were resized to $640\times640$ pixels, with pixel normalisation bounded $[0,1]$ across all three RGB channels.
    \item \textbf{CelebA~\citep{liu2015faceattributes}:} consists of $178\times218$ aligned RGB images of celebrity faces, spanning a single object-detection class (face). We utilise the standard split of 162,770 training images, 19,867 calibration images, and 19,962 test images. Images were resized to $640\times640$ pixels, with pixel normalisation bounded $[0,1]$ across all three RGB channels.
    \item \textbf{VisDrone~\citep{zhu2021detection}:} consists of variable-resolution RGB aerial images captured by drone-mounted cameras across urban and rural environments, containing objects such as pedestrians, bicycles, cars, trucks, and buses spanning 10 classes. We utilise the standard VisDrone2019-DET split of 6,471 training images, 1,610 test images, and 548 calibration images. Images were resized to $1600\times1600$ pixels, with pixel normalisation bounded $[0,1]$ across all three RGB channels.
    \item \textbf{TT100k~\citep{Zhe_2016_CVPR}:} consists of $2048\times2048$ RGB street-view images captured under diverse illumination, weather, viewing-distance, and occlusion conditions, containing traffic signs (e.g., speed limits, prohibitions, warnings, and information signs) spanning 221 classes. We utilise the standard split of 6,105 training images, 7,641 calibration images, and 3,071 test images. Images were resized to $1280\times1280$ pixels, with pixel normalisation bounded $[0,1]$ across all three RGB channels.
\end{itemize}

\subsection{Models}
\label{sec:appendix-models}

To assess the performance and generalisability of \methodname, we assess its performance against multiple models and architectures across our experimental evaluation. Below, we provide a concise description of each model together with the implementation details and hyperparameter settings adopted in our experiments:

\begin{itemize}[noitemsep, nolistsep]
    \item \textbf{YOLO11n~\citep{khanam2024yolov11}:} is used as the base object detector for all datasets in the core experiments. We initialise the lightweight nano variant from COCO-pretrained weights. For COCO, the original 80-class detection head is retained, while for CelebA, VisDrone, and TT100K, the detection head is adapted to the corresponding dataset classes and the model is fine-tuned on each dataset. Following fine-tuning, all detector parameters are frozen during \methodname\ training.
\end{itemize}

\subsection{Comparative Approaches}
\label{sec:appendix-Comparative Methods}

To assess the performance of \methodname, we benchmark it against a set of recent post-hoc object detection uncertainty quantification methods that capture the state of the art in this area. Below, we provide a concise description of each method together with the implementation details and hyperparameter settings adopted in our experiments:

\begin{itemize}[noitemsep, nolistsep]
    \item \textbf{Evidential Meta Model (EMM)~\citep{shen2023post}:} is a post-hoc uncertainty approach that combines representations from multiple frozen intermediate layers using a lightweight evidential meta-model. We adapt EMM to object detection by applying ROIAlign to predefined detector features for each predicted bounding box and parameterising a Normal-Inverse-Gamma distribution over its four localisation coordinates. Unlike \methodname, EMM uses only clean training detections and does not employ saliency-based layer selection, weight maps, or the noise-driven curriculum, making it a direct ablation of \methodname's uncertainty-guidance components. The meta-model is trained for 20 epochs using Adam with an initial learning rate of $10^{-2}$, cosine annealing to $10^{-6}$, an IoU matching threshold of $0.5$, evidential regularisation weight $\lambda_{\mathrm{reg}}=0.1$, and gradient clipping at $1.0$.
    \item \textbf{ModelNet~\citep{lee2022lightweight}:} is a deterministic uncertainty-distillation approach that learns the spatial and semantic uncertainty produced by an MC Dropout object detector. We adapt ModelNet as an uncertainty assistant to the frozen YOLO11n detector, using Dirichlet outputs for semantic uncertainty and Normal-Gamma outputs for bounding-box uncertainty without altering the detector's predictions. Teacher targets are generated using 40 MC Dropout passes with a dropout probability of $0.1$ and an IoU matching threshold of $0.5$. ModelNet is trained for 20 epochs using Adam with a learning rate of $10^{-4}$, cosine annealing to $10^{-6}$, gradient clipping at $1.0$, and loss weights of $100$, $10$, $1$, and $100$ for the classification, localisation, Dirichlet, and Normal-Gamma objectives, respectively.
    \item \textbf{MetaDetect~\citep{schubert2021metadetect}:} is a post-hoc prediction-quality approach that uses detector-output and bounding-box geometry features to train a meta-classifier for distinguishing true-positive and false-positive detections and a meta-regressor for estimating detection IoU. We implement a reduced geometry-only variant for YOLO11n (as a fair comparison to \methodname's experiments, where classification is not considered). True positives are defined using an IoU threshold of $0.5$. The classifier and regressor use histogram-based gradient boosting with one boosting iteration, a learning rate of $10^{-4}$, and $\ell_2$ regularisation of $100$.
\end{itemize}

\subsection{Training Details}
\label{sec:appendix-Training Details}

All experiments use YOLO11n as the pretrained base detector. For COCO, we use the pretrained Ultralytics weights. For CelebA, VisDrone, and TT100K, the detector is initialised from COCO-pretrained weights and fine-tuned independently on each dataset for 100 epochs using Adam with an initial learning rate of 0.01 and batch size 8. Following training, all detector parameters are frozen and remain unchanged during uncertainty-model training.

\methodname\ is trained independently for each frozen detector and dataset combination. Saliency calibration uses a coverage threshold of $\eta=0.7$. The uncertainty meta-model is subsequently trained for 20 epochs using Adam with an initial learning rate of 0.01, annealed to $10^{-6}$, a batch size of 8, and gradient clipping at 1.0. Unless otherwise stated, the noise-driven curriculum uses $T=10$, $\gamma=0.1$, an IoU matching threshold of 0.1, and evidential regularisation weight $\lambda_{\mathrm{reg}}=0.1$ (chosen following the recommended practice~\citep{amini2020deep}). The base detector remains frozen throughout both saliency calibration and uncertainty-guided training.

We report results over five independent runs using distinct random seeds. All models are implemented, trained, and evaluated using PyTorch. Experiments are conducted on a high-performance computing cluster using up to three NVIDIA A40 GPUs, 32 CPU cores, and 167\,GB of system memory.

\subsection{Metrics and Attacks}
\label{sec:appendix-Metrics and Attacks}
In addition to standard detection metrics, we consider and describe some additional uncertainty-specific metrics as well as some detection-specific adversarial attacks:

\begin{itemize}[noitemsep, nolistsep]
    \item \textbf{Success-Score Calibration Error (SSCE):}
    measures agreement between the evaluated detection-success score $s_i$ and
    the empirical frequency of true-positive detections. For $B$ score bins:
    \begin{equation}
        \mathrm{SSCE}
        =
        \sum_{b=1}^{B}
        \frac{|B_b|}{N}
        \left|
        \frac{1}{|B_b|}\sum_{i\in B_b}y_i
        -
        \frac{1}{|B_b|}\sum_{i\in B_b}s_i
        \right|
    \end{equation}
    where $y_i=1$ for a TP and $y_i=0$ for an FP.
    \item \textbf{PGD ($\ell_\infty$)~\citep{madry2018towards}:}
    is a white-box projected-gradient attack constrained within an
    $\ell_\infty$ perturbation budget. The detection-specific objective
    jointly reduces the scores of clean reference detections and increases
    their localisation displacement.

    \item \textbf{PGD ($\ell_2$)~\citep{madry2018towards}:}
    uses the same detection-specific objective as PGD ($\ell_\infty$), but
    constrains the complete image perturbation within an $\ell_2$ norm ball.

    \item \textbf{TOG-Vanishing~\citep{chow2020adversarial}:}
    is a white-box Targeted Objectness Gradient attack that suppresses the
    confidence of detections present in the clean image, encouraging existing
    objects to disappear from the detector output.

    \item \textbf{TOG-Fabrication~\citep{chow2020adversarial}:}
    instead increases the scores of low-confidence background candidates,
    encouraging the detector to produce spurious object detections.

    \item \textbf{Square Attack ($\ell_\infty$)~\citep{andriushchenko2020square}:}
    is a query-based black-box attack that performs random square-shaped
    perturbation updates without access to model gradients. Our
    detection-adapted objective accepts updates that reduce the
    confidence-weighted IoU retention of the clean detections.
\end{itemize}

\subsection{Runtime and Complexity}
\label{sec:Runtime and Complexity}

All experiments were implemented in PyTorch and executed on a large performance GPU cluster using a maximum of three Nvidia A40 GPUs, 32 CPU cores, 167GB of memory (per GPU). All models were trained from scratch. All runs used random seeds.

Let $N$ denote the number of calibration samples, $E$ the number of meta-model training epochs, $\bar{J}$ the mean number of retained detections, $F$ the cost of a detector forward pass, and $A$ the cost of a relevance-attribution pass. \methodname\ incurs a one-off saliency-calibration cost dominated by detector inference and relevance propagation. Layer selection requires one attribution pass per calibration sample, while construction of localisation weight maps requires four attribution passes per retained detection, giving a dominant calibration cost of approximately $O(N\bar{J}A)$. Meta-model training is dominated by a constant number of frozen-detector forward passes per sample, giving $O(ENF)$, alongside lower-order costs for corruption, detection matching, and the lightweight evidential head. Although the curriculum defines $T+1$ corruption levels, only one is sampled per example, avoiding an $O(ENTF)$ cost. At inference, intermediate features are reused from the detector's existing forward pass, yielding $O(F+\bar{J}D)$ total complexity, or $O(\bar{J})$ additional overhead for fixed feature dimensions $D$.


\end{document}